\documentclass{article}
\usepackage[preprint]{colm2026_conference}

\usepackage{microtype}
\usepackage[dvipsnames]{xcolor}
\usepackage{amsmath,amssymb}
\usepackage{graphicx}
\usepackage{subcaption}
\graphicspath{{images/}}
\usepackage{float}
\usepackage{wrapfig}
\usepackage{needspace}
\usepackage{booktabs,tabularx,makecell,array}
\usepackage{colortbl}
\usepackage{pifont}
\usepackage{fontawesome5}
\usepackage{enumitem}
\usepackage{url}
\usepackage{lineno}
\usepackage{hyperref}
\usepackage[capitalize]{cleveref}

\definecolor{darkblue}{rgb}{0,0,0.5}
\hypersetup{colorlinks=true,citecolor=darkblue,linkcolor=darkblue,urlcolor=darkblue}
\newcommand{\yes}{\textcolor{green!50!black}{\ding{51}}}
\newcommand{\no}{\textcolor{red!70!black}{\ding{55}}}

\title{\centering PatentAgentBench: Benchmarking LLM Agents with  Professional Patent Task Suites%
  \texorpdfstring{\\[0.6em]
    {\normalfont\normalsize
      \makebox[\textwidth][c]{%
        \href{https://github.com/patsnap/PatentAgentBench}{\faGithub\hspace{0.4em}\texttt{PatentAgentBench}}%
        \hspace{1.5em}%
        \href{https://huggingface.co/collections/PatSnap/patentagentbench}{%
          \raisebox{-0.15em}{\includegraphics[height=1.1em]{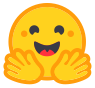}}\hspace{0.4em}\texttt{PatentAgentBench}}%
      }}%
  }{}%
}

\author{%
\makebox[\dimexpr\textwidth-2\tabcolsep\relax][c]{%
  {\small\bfseries Yijie Hu, Wei Li, Junjie Zhou, Zhiqi Jiang, Yi Jin}%
}\\[4pt]
\makebox[\dimexpr\textwidth-2\tabcolsep\relax][c]{%
  {\small\bfseries Licong Xu, Xinkai Sun, Kai Duan$^{\ddagger}$, Lisha Zhang$^{\dagger,\ddagger}$}%
}\\[7pt]
\makebox[\dimexpr\textwidth-2\tabcolsep\relax][c]{%
  {\normalfont AI Eval \& Alignment, PatSnap Ltd. Co.}%
}\\[6pt]
\makebox[\dimexpr\textwidth-2\tabcolsep\relax][c]{%
  {\normalfont\small $^{\ddagger}$project lead\qquad $^{\dagger}$corresponding author}%
}%
}

\begin{document}
\vspace*{-12pt}
\maketitle
\fancyhead{}
\fancyhead[R]{\normalfont\footnotesize2026-10-26}
\renewcommand{\headrulewidth}{1.5pt}
\AddToShipoutPictureFG*{%
  \AtTextUpperLeft{%
    \raisebox{17mm}[0pt][0pt]{%
      \includegraphics[trim=0 0 696bp 0,clip,height=6mm]{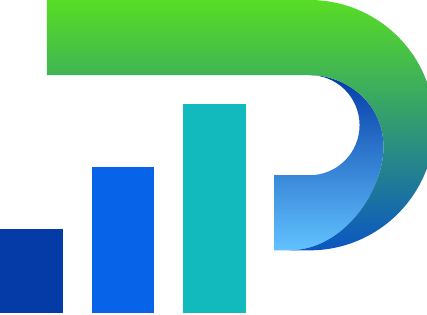}%
      \hspace{4mm}%
      \includegraphics[trim=0 0 166bp 0,clip,height=6mm]{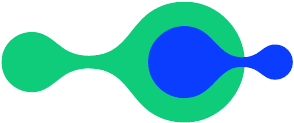}%
    }%
  }%
}

\vspace{-8pt}
\raggedbottom
\renewenvironment{abstract}{\vskip.025in\centerline{\large\bfseries Abstract}%
\vspace{0.5ex}\begin{list}{}{\leftmargin=10pt\rightmargin=10pt}\item\relax}%
{\par\end{list}\vskip 0.5ex}
\begin{abstract}
Patent practice lies at the intersection of technical, legal, and linguistic work, making it challenging yet crucial to technological innovation. Recently, rapid advances in large language models (LLMs) and agentic systems have demonstrated the potential to complete realistic business workflows and even patent works. This raises a question: \emph{can current AI systems perform patent work reliably?} Existing benchmarks cover a limited range of patent tasks and simplify evidence acquisition by providing reference materials or candidate documents. In this paper,  we introduce \textbf{PatentAgentBench}, a benchmark comprising 510 instances across eight task suites: novelty search, freedom-to-operate (FTO) search, design FTO, invalidity search, claim charting, patent drafting, office-action response, and patent translation. Curated from real patent records and professional materials, the benchmark combines task-specific metrics with expert-reviewed annotations and rubrics. Retrieval tasks such novelty search, FTO, etc., require agents to search and compare patent collections without a supplied candidate pool. The drafting and OAR tasks assess the quality of agents' work given technical disclosure and office action notice. We evaluate seven LLMs across four agent harnesses on PatentAgentBench,  finding that current systems still struggle to identify novelty, assess risk, and understand practitioner intent throughout patent workflows. The strongest model--harness pair achieves an average score of 52.8, leaving substantial room for improvement across professional patent tasks. Our analysis demonstrates that current agents struggle to retrieve relevant evidence and produce substantively sound patent documents. Although patent drafts generally follow the expected structure, weaknesses persist in claim organization, case understanding, and legal argumentation. These findings highlight the gap between generating plausible documents and delivering reliable patent work.
\end{abstract}

\begingroup
\setlength{\intextsep}{4pt}
\begin{figure}[H]
  \centering
  \captionsetup{font=small,skip=3pt}
  \includegraphics[width=0.96\linewidth,trim=0 10 0 45,clip]{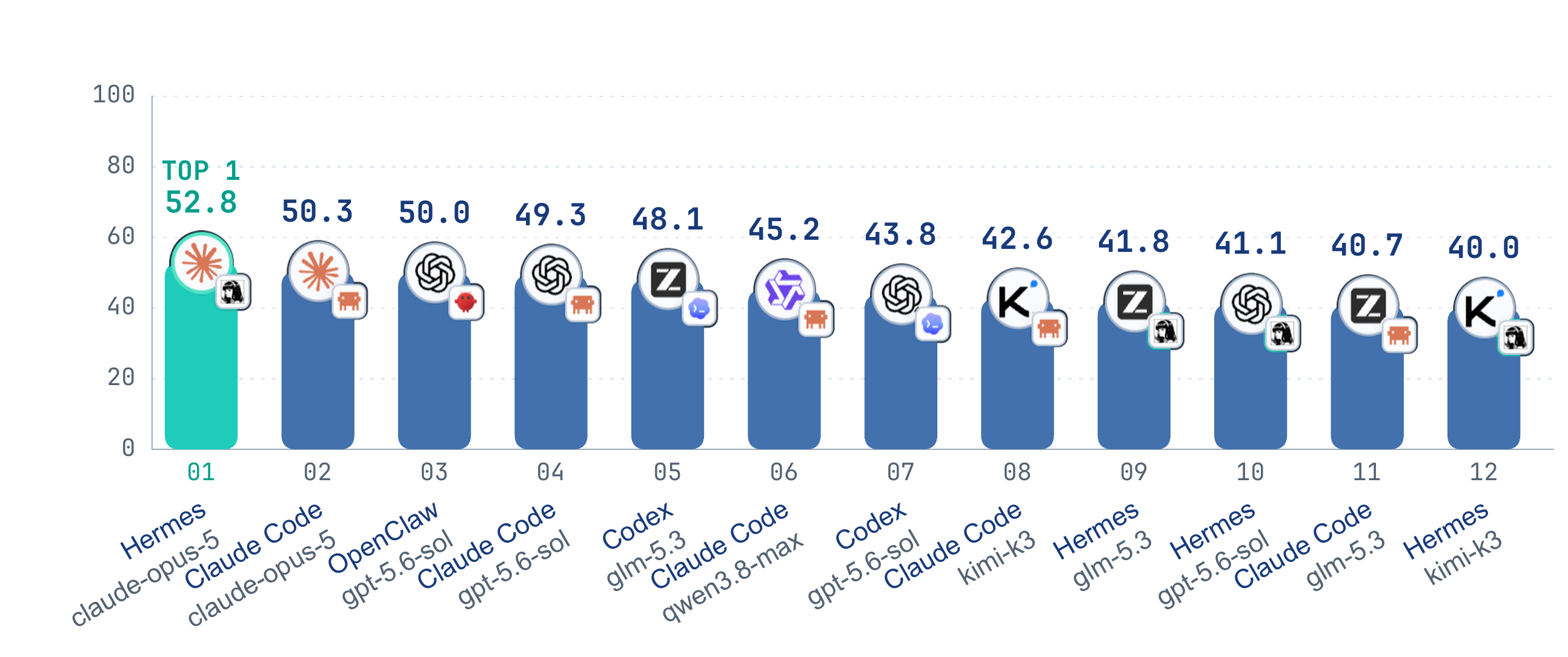}
  \caption{Top twelve model--harness pairs ranked by mean score across the eight task suites.}
  \label{fig:overall_score_overview}
\end{figure}
\endgroup

\clearpage
\setcounter{tocdepth}{2}
{\fontsize{9}{10}\selectfont\setlength{\parskip}{0pt}\tableofcontents}
\clearpage
\raggedbottom

\section{Introduction}

Patents play a vital role in protecting inventions, facilitating technology transfer, and providing legal certainty. Patent practice consists of a series of important tasks. For example, a technical idea must be first assessed through novelty search, evaluated in freedom-to-operate analyses or invalidity proceedings~\citep{krestel2021survey}, then developed into a patent draft, followed by examination and rebuttal before final publication~\citep{srinivas2024towards}. Patent practice is therefore both knowledge and labor-intensive, requiring coordinated technical, legal, and linguistic expertise, while remaining essential to protecting innovation and managing technology-related risks.

Recently, the advancements in Large Language Models (LLM) and agentic systems~\citep{liu2024deepseek,singh2025openai, comanici2025gemini, zeng2026glm, team2026kimi, xu2026deepseek} have demonstrated that agentic systems~\citep{luo2025large} possess long-horizon working abilities, delivering high-value tasks such as coding and legal tasks~\citep{yang2024swe,merrill2026terminal,harveylab2026}. These promising advancements demonstrate the potential to accelerate or even automate patent work. This raises a question: \textit{Can prevalent LLM agents reliably perform patent-related tasks?} 

Previous efforts that evaluate patent-related capabilities are mainly through examination questions, classification, and output evaluation~\citep{wang2024ipeval,lee2024patentedits,knappich2025pedantic,jiang2025towards,wang2025ipbench,jang2025pilot}, as summarized in \cref{tab1}. These benchmarks provide valuable measures of patent knowledge, reasoning, and selected generation capabilities for LLMs, yet leave two gaps in the evaluation of professional patent work. \textbf{First, coverage of professional tasks remains limited}. For example, IPBench~\citep{wang2025ipbench} covers diverse IP topics but primarily uses multiple-choice and classification tasks, with generation limited to abstracts and dependent claims. PANORAMA~\citep{lim2026panorama} focuses on prior-art selection, evidence localization, and patentability assessment within examination. Broader tasks, including FTO search, invalidity search, and patent translation, remain unexplored. \textbf{Second, the evidence acquisition procedure is often over-simplified by supplying case materials or pre-constructed candidate references}. This alleviates the need for systems to formulate queries, retrieve documents, and refine their search as new evidence emerges. Consequently, the performance under these settings does not establish whether a system can independently obtain and use the evidence needed to complete a professional patent task.
\begin{figure*}
    \centering
    \includegraphics[width=\linewidth]{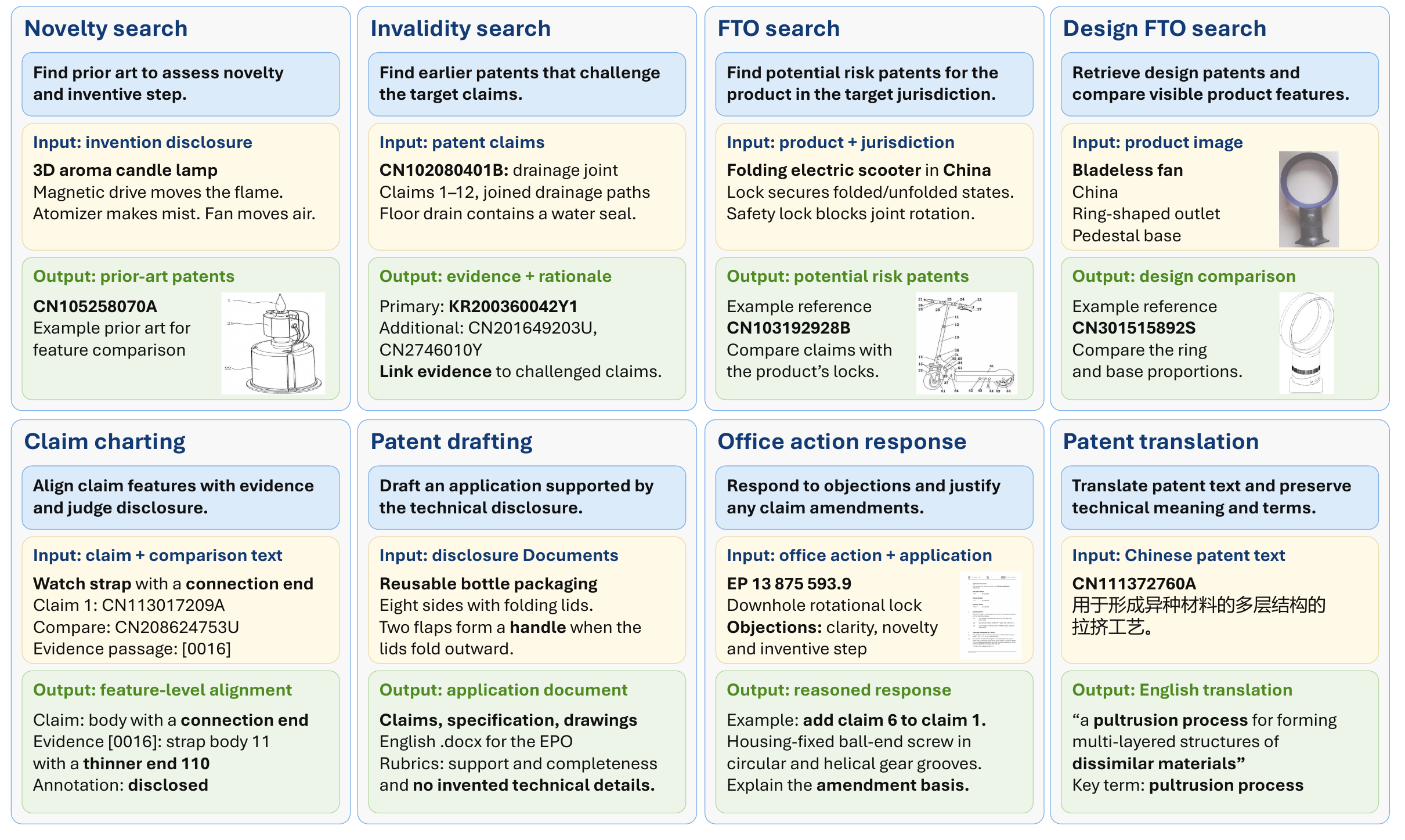}
    \caption{Representative tasks of PatentAgentBench. The search-related tasks, such as novelty search, ask the agent to retrieve evidence and results without the given references. Drafting and OAR tasks ask the agent to do research based on the given material, then deliver reliable patent drafting or rebuttal.}
    \label{fig_banner}
\end{figure*}

In this paper, we introduce \textbf{PatentAgentBench}, a benchmark
for evaluating LLM agents on eight professional patent tasks:
novelty search, freedom-to-operate (FTO) search, design FTO,
invalidity search, claim charting, patent drafting,
office-action response (OAR), and patent translation.
We illustrate some of the representative tasks in \cref{fig_banner}, covering evidence discovery, claim-level analysis, and professional document production across patent practice. The retrieval-related task suites require agents to formulate queries, search patent collections, and identify relevant documents \textit{without} a supplied candidate pool. We construct the benchmark through a four-stage process that combines expert guidance with automated data collection, case reconstruction, and final validation. 
Cases are constructed from actual patent records, official proceedings, and professional work materials. We identify the task inputs and reference outcomes within each source record, then reconstruct the inputs available to an agent while withholding reference answers. Domain experts are involved to validate the completeness of the materials and annotate task-specific reference evidence or scoring criteria. Each instance specifies the permitted evidence, required output, and evaluation protocol, with reference evidence linked to its source. Evaluation combines retrieval and extraction metrics with expert-reviewed rubrics to assess evidence coverage, factual support, and the quality of the final work product. 

We evaluate seven prevalent LLMs across four agent harnesses, comparing models under a fixed harness and harnesses with a fixed model. The strongest pairing achieves an average score of 52.8, leaving substantial room for improvement across professional patent tasks. Detailed analysis further reveal gaps beyond the aggregate scores: under a fixed harness such as Claude Code, all seven models recover less than one-third of annotated references in task such as invalidity search, and majority of them perform poor on evidence alignment than on feature extraction. Document-level analysis reveals a gap between completing a document and ensuring its substantive quality. Patent drafts follow the expected structure but lack meaningful layers of dependent-claim protection. In office-action response, high completion and compliance scores coexist with weaknesses in understanding the case, choosing a response strategy, and developing well-supported arguments. These results reveal that current LLM agents still cannot perform these patent-related tasks reliably, especially in evidence acquisition, evidence-grounded reasoning, and deliver high quality patent documents.
We summarize our contributions as follows:

\begin{itemize}
    \item We introduce PatentAgentBench, a benchmark covering eight professional patent tasks, with cases constructed from actual patent records, official proceedings, and professional work materials.

    \item We propose task-specific evaluation metrics that combine reference-based metrics with expert-reviewed rubrics, assessing evidence retrieval, factual support, and the substantive quality of patent work products.

    \item We evaluate seven LLMs across four agent harnesses and analyze performance from both perspectives. Our results reveal task-dependent model--harness interactions and persistent gaps in evidence coverage, evidence alignment, and substantive document quality.
\end{itemize}

\section{Related Works}
\subsection{LLM Agents}
Research on LLM-based agents has progressed from general formulations of reasoning and action to autonomous cooperation, multi-model coordination, multi-agent frameworks, and generalist orchestration~\citep{yao2022react,tran2025multi}. Early work characterizes agents as systems that perceive their environment, make decisions, and take actions~\citep{shinn2023reflexion}. CAMEL~\citep{li2023camel} enables autonomous cooperation through role-playing, while HuggingGPT~\citep{shen2023hugginggpt} uses an LLM to plan tasks and coordinate external AI models across multiple modalities. AutoGen~\citep{wu2024autogen} extends this direction by providing a general framework for interactions among LLMs, humans, and tools. More recent systems move toward generalist orchestration: Magentic-One~\citep{fourney2024magenticone} coordinates specialized agents operating over browsers, files, and code, while SuperAgent~\citep{yao2025superagent} routes user requests to tool-equipped agents or generated workflows and selects models according to task complexity. Together, these studies trace a shift from individual LLM agents to coordinated systems that complete complex workflows through a unified interface. PatentAgentBench evaluates the final professional work products of agentic systems across eight representative patent-practice task suites.

\subsection{Patent Practice Benchmarks}
Prior research has studied patent analysis through classification, retrieval, quality assessment, and text generation~\citep{shomee2025survey}. IPEval~\citep{wang2024ipeval} evaluates IP-related knowledge and legal reasoning using bilingual questions drawn from patent-agent examinations. IPBench further expands the scope to 20 tasks~\citep{wang2025ipbench}. However, most IPBench instances remain closed-set questions or classification problems, patent generation tasks are limited to patent abstracts and dependent claims. PatentEdits~\citep{lee2024patentedits} asks model to predict whether claim will be kept, edited, or deleted during prosecution without revisions. PEDANTIC~\citep{knappich2025pedantic} evaluates claim definiteness and the identification of rejection reasons. Patent-CE~\citep{jiang2025towards} evaluates the quality of candidate claims against reference claims. PILOT-Bench~\citep{jang2025pilot} formulates PTAB appeal reasoning as independent classification tasks for issues, authorities, and outcomes, examining the legal reasoning capabilities. More recent work moves closer to prosecution practice. PANORAMA~\citep{lim2026panorama} evaluates prior-art selection, evidence localization, and patentability classification as separate closed-pool tasks. PatRe~\citep{wang2026patre} generates office actions and applicant responses from prosecution records, although its retrieval-simulated setting supplies a small candidate pool containing oracle references. In contrast, PatentAgentBench evaluates complete, evidence-grounded work products across nine representative patent-practice task suites, covering invention intake, patentability analysis, clearance, validity analysis, evidence mapping, drafting, prosecution, and multilingual practice under a unified evaluation framework.

\begin{table*}[t]
\centering
\caption{Coverage of professional patent tasks in existing benchmarks.}
\label{tab1}

\footnotesize
\renewcommand{\arraystretch}{1.15}
\setlength{\tabcolsep}{3.5pt}

\resizebox{\textwidth}{!}{%
\begin{tabular}{
    @{}
    >{\raggedright\arraybackslash}m{2.9cm}
    >{\raggedright\arraybackslash}m{3.65cm}
    *{8}{>{\centering\arraybackslash}m{1.45cm}}
    @{}
}
\toprule
\textbf{Benchmark}
& \textbf{Evaluated Tasks}
& \makecell{\textbf{Novelty}\\\textbf{Search}}
& \makecell{\textbf{Patent}\\\textbf{FTO}}
& \makecell{\textbf{Design}\\\textbf{FTO}}
& \makecell{\textbf{Invalidity}\\\textbf{Search}}
& \makecell{\textbf{Claim}\\\textbf{Charting}}
& \makecell{\textbf{Patent}\\\textbf{Drafting}}
& \makecell{\textbf{OA}\\\textbf{Response}}
& \makecell{\textbf{Patent}\\\textbf{Translation}} \\
\midrule

IPEval~\citeyearpar{wang2024ipeval}
& \makecell[l]{IP, legal knowledge\\QA}
& \no & \no & \no & \no
& \no & \no & \no & \no \\

\addlinespace[2pt]

PatentEdits~\citeyearpar{lee2024patentedits}
& \makecell[l]{Claim-sentence\\edit prediction}
& \no & \no & \no & \no
& \no & \no & \no & \no \\

\addlinespace[2pt]

PEDANTIC~\citeyearpar{knappich2025pedantic}
& \makecell[l]{Claim\\identification}
& \no & \no & \no & \no
& \no & \no & \no & \no \\

\addlinespace[2pt]

Patent-CE~\citeyearpar{jiang2025towards}
& \makecell[l]{Patent claim\\quality assessment}
& \no & \no & \no & \no
& \no & \no & \no & \no \\

\addlinespace[2pt]

IPBench~\citeyearpar{wang2025ipbench}
& \makecell[l]{Patent search\\and drafting}
& \yes & \no & \no & \no
& \no & \yes & \no & \no \\

\addlinespace[2pt]

PANORAMA~\citeyearpar{lim2026panorama}
& \makecell[l]{Novelty search\\and claim charting}
& \yes & \no & \no & \no
& \yes & \no & \no & \no \\

\addlinespace[2pt]

PILOT-Bench~\citeyearpar{jang2025pilot}
& \makecell[l]{Legal reasoning\\and outcome prediction}
& \no & \no & \no & \no
& \no & \no & \no & \no \\

\addlinespace[2pt]

PatRe~\citeyearpar{wang2026patre}
& \makecell[l]{Office action\\and response generation}
& \no & \no & \no & \no
& \no & \no & \yes & \no \\

\midrule

\textbf{PatentAgentBench}&
& \yes & \yes & \yes & \yes
& \yes & \yes & \yes & \yes \\

\bottomrule
\end{tabular}%
}
\end{table*}

\section{PatentAgentBench}
We first demonstrate the general data curation pipeline in~\cref{data_curation}, each sub-task follows the similar data curation process with small variations, we will introduce them separately in the section in \cref{bench_feature}.

\subsection{Data Curation}
\label{data_curation}

\begin{figure*}
    \centering
    \includegraphics[width=\linewidth]{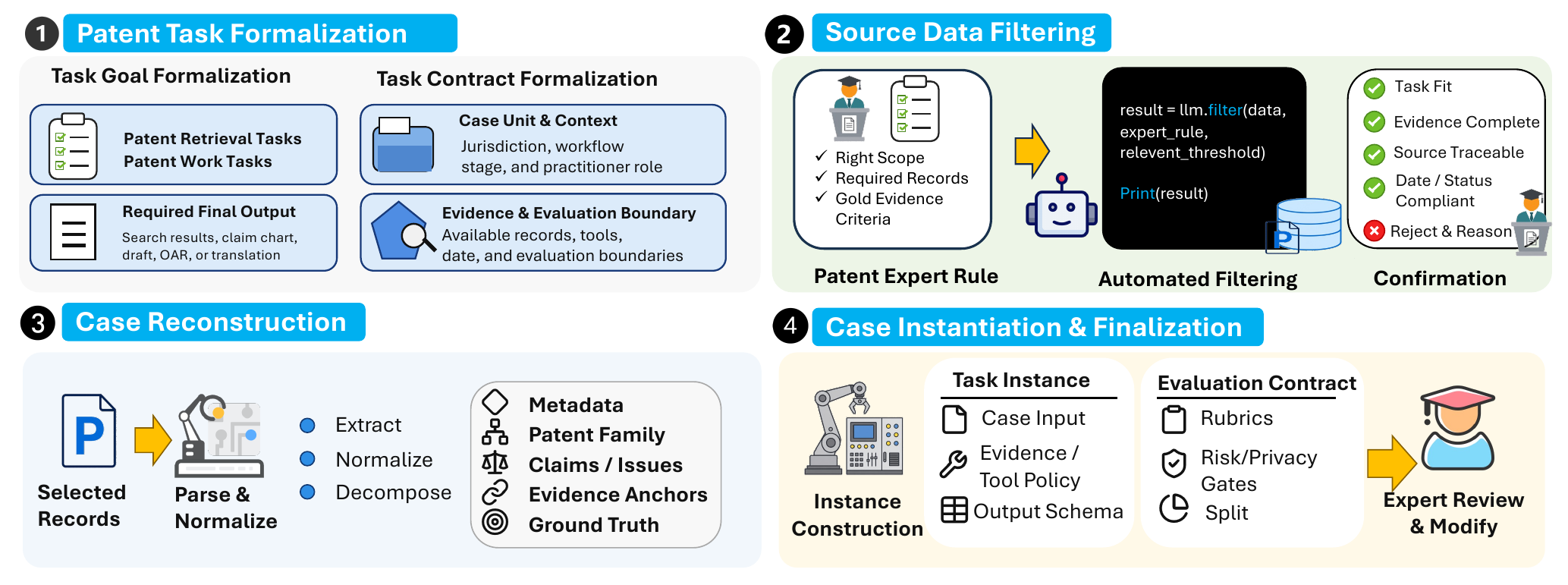}
    \caption{Four-stage data curation pipeline.}
    \label{fig2}
\end{figure*}

PatentAgentBench follows a four-stage curation pipeline that combines automated processing with expert review, as illustrated in \cref{fig2}. First, we define each task's input, expected deliverable item, task evidence, and evaluation criteria based on professional patent practice. Second, experts establish inclusion and exclusion criteria to guide rule-based and LLM-assisted screening of patent records, official proceedings, and professional work materials. Our expert team comprises 31 professionals: five senior patent examiners, fifteen senior patent agents, three dual-qualified lawyers, and eight senior IPR specialists. Their detailed backgrounds are reported in Appendix~\ref{app:expert_profile}. Experts review candidate cases for relevance, completeness, and suitability. Third, we parse and normalize the selected materials, reconstruct task inputs, and derive reference evidence or scoring criteria from the source records. Automated pre-annotation supports this process, while experts correct labels and review rubrics for factual support and professional relevance. Each instance is annotated by one expert and reviewed by a second expert. Finally, each case is packaged with its input, evidence conditions, output requirements, and evaluation anchors. Before final validation, we check that reference evidence is traceable to its source and that task inputs do not expose the answers.

\subsection{Task Taxonomy}
\label{bench_feature}
PatentAgentBench covers eight task suites spanning evidence discovery, feature-level comparison, and document preparation. Novelty search, patent FTO, design FTO, and invalidity search require agents to retrieve relevant evidence. Claim charting requires them to map technical features to supplied evidence.Patent drafting, office action response, and translation require them to produce professional documents. Table~\ref{tab:PatentAgentBench_statistics} summarizes the dataset statistics, see more details in Appendix~\ref{tab:dataset_details}. We describe the purpose, task specification, and data construction of each suite below. 

\begin{wraptable}{r}{0.64\textwidth}
\vspace{-\intextsep}
\setlength{\tabcolsep}{3pt}
\footnotesize
\centering
\caption{Statistics of PatentAgentBench across eight task suites.}
\label{tab:PatentAgentBench_statistics}

\resizebox{\linewidth}{!}{%
\begin{tabular}{@{}lccc@{}}
\toprule
\textbf{Subset} & \textbf{Size} & \textbf{Jurisdiction} & \textbf{Metrics} \\
\midrule
Novelty Search    & 80  & CN/US/EP/WO & Hit Rate/Recall \\
FTO               & 60  & CN/US/EP    & Recall \\
Design FTO        & 80  & CN/EU       & Recall \\
Invalidity Search & 80  & CN/US       & Recall \\
Claim Charting    & 30  & CN/US/WO    & F1 \\
OAR               & 40  & CN/US/EP    & 57 Rubrics \\
Patent Drafting   & 60  & CN/US/EP    & 977 Rubrics \\
Translation       & 80  & CN/US/EP/WO & \makecell{BLEU/METEOR/ROUGE} \\
\midrule
\textbf{Total } & \textbf{510} &  &  \\
\bottomrule
\end{tabular}%
}

\end{wraptable}

\noindent\textbf{Novelty Search.}
Novelty search identifies prior art relevant to assessing whether a technical solution is new. The agent receives a technical specification and returns a ranked list of prior patents. We construct the ground truth from patent families examined by multiple offices. For each family, we collect the X references cited by examiners for the target patent and its family members,then merge them into a deduplicated reference set. The input contains only the technical disclosure, with citation information excluded to prevent leakage. The 80 cases cover Chinese and English and are balanced across IPC main classes and patent offices.

\noindent\textbf{Patent FTO.}
Freedom-to-operate (FTO) search identifies claims that may block the implementation of a technical solution or product by systematically searching granted, valid patents in a target jurisdiction. The agent receives a technical description and the target jurisdiction, and returns a screened list of potentially blocking patents. We construct cases from court records with established infringement findings: the accused product's technical features form the input, and the patents held to be infringed form the reference set. The cases cover CN, US, and EP jurisdictions. This suite requires agents to consider claim scope and legal status when identifying infringement risks, rather than relying on technical similarity alone.

\noindent\textbf{Design FTO.}
Design FTO identifies design patents that may pose infringement risks to a product's appearance. The agent receives a target image and jurisdiction, and returns a ranked list of design patents with similarity scores and risk levels. We construct 80 cases from final invalidation decisions issued by the relevant offices, using the invalidated design patents and evidence images from these cases. The cases span 26 first-level Locarno classes. The task requires agents to extract the target design features, compare them with retrieved design images, and assess infringement risks under the rules of the target jurisdiction. We evaluate retrieval coverage against the annotated reference patents.

\noindent\textbf{Invalidity Search.}
Invalidity search retrieves prior art that can support a challenge to granted patent claims. The agent receives the target patent's application and publication numbers, filing date, and earliest priority date. It retrieves and analyzes the patent, then returns prior-art references with the relevant novelty or inventive-step grounds, affected claims, and brief explanations. We construct cases from official invalidation decisions and final written decisions in inter partes review. We retain one representative granted patent per case and deduplicate cases at the application level. Alongside the cited references, we preserve the statutory provisions, reasoning, evidence combinations, and affected claims, so that retrieved evidence can be checked against the grounds established in the decision.

\noindent\textbf{Claim Charting.}
Claim charting maps individual technical features to supporting evidence for novelty assessment, FTO analysis, and invalidity proceedings. Given a target text and comparison documents, the agent must identify the target's technical features, determine whether each is disclosed, and cite the corresponding evidence with an explanation. We construct cases from Chinese and English patents and technical texts covering IPC sections A, B, C, G, and H. External patent experts annotate feature-level disclosure judgments, word-level evidence mappings, and explanations. We cross-check and standardize these annotations so that each judgment is traceable to specific evidence.

\noindent\textbf{Patent Drafting.}
Patent drafting turns an invention disclosure into an initial patent application. The agent receives disclosure materials and produces a complete draft, including claims and a specification. We construct cases from real patents, anonymize confidential information, and ask patent experts to review the disclosure materials and develop evaluation rubrics. The cases cover CN, US, and EP jurisdictions and IPC sections A--H. General rubrics assess language quality, document structure, claim coverage, and consistency between the claims and specification. Case-specific rubrics assess the technical features, relationships, and constraints in each disclosure. Together, they evaluate whether the draft clearly describes the invention, preserves its essential features, and supports the claimed scope without introducing unsupported content.

\noindent\textbf{Office Action Response.}
An office action sets out an examiner's objections to a patent application. Responding requires addressing these objections through technical and legal arguments and, where appropriate, amendments. The agent receives the office action, jurisdiction, and available application materials and cited references, and produces a formal response. We collect 40 cases from CN, US, and EP patent examinations. Patent experts review the case materials and response requirements, including the objections raised, evidence needed to support an argument, and disclosure supporting an amendment. Evaluation uses 57 rubric criteria covering case understanding, response strategy, amendment quality, substantive argumentation, completeness, and factual and citation accuracy. Each criterion specifies its applicability and scoring levels.

\noindent\textbf{Patent Translation.}
Patent translation supports filing and examination across languages while preserving technical meaning and claim scope. We include 80 document-level translation cases with 881 word, character, sentence and paragraph-level items, supporting the category-specific evaluation. The agent translates Chinese patent text into English or English text into Chinese without adding, omitting, or altering technical content. We organize translation pairs at character, word, sentence, paragraph, and document levels, covering titles, abstracts, claims, and descriptions. Each instance is labeled by text granularity and evaluation dimension, with reference translations or target expressions for terminology, special characters, and patent writing conventions.

\section{Experiments}
\subsection{Experimental Settings and Metrics}
We evaluate seven LLMs through four agent harnesses, comparing models within each harness and harnesses for each model. The models are GPT-5.6-Sol, Qwen3.8-Max, Kimi K3, DeepSeek-V4-Pro, MiniMax-M3, GLM-5.3, and Claude Opus 5. The harnesses are Codex, Claude Code, Openclaw, and Hermes. Agents can call a commercial patent search MCP service to access to the patent database. We use GPT 6 sol as the LLM judge for rubric-based evaluation. More detailed Harness descriptions and versions, model names, and the execution protocol are in Appendix~\ref{experimental_details}. Claude Opus 5 was not evaluated with Codex because the model was incompatible with the protocol supported by Codex configuration.

Novelty search uses Hit@100, the proportion of cases with at least one annotated reference retrieved among the top 100 results. Patent FTO, design FTO, and invalidity search use recall to measure coverage of annotated references. Claim charting uses F1 score. Patent drafting and OAR use weighted rubric-based scores. OAR is assessed using 57 criteria covering basic quality, amendment quality, and substantive argumentation. Translation uses category-specific metrics, averaged across the two translation directions. All primary task scores in \cref{tab:main_results} are expressed on a 0--100 scale, and AVG is their  mean score. Appendix~\ref{metric_details} defines the matching rules, formulas, rubric weights, and aggregation for each task.

\subsection{Experimental Results}

\begin{table}[!t]
\centering
\caption{Main results on PatentAgentBench. Scores are on a 0--100 scale. Higher is better. AVG is the mean task score, with translation averaged over both directions. Within each harness, darker and lighter blue cells mark the highest and second-highest positive scores, respectively. Ties at the displayed precision share the same color.}
\label{tab:main_results}
\scriptsize
\setlength{\tabcolsep}{3pt}
\renewcommand{\arraystretch}{1.15}
\resizebox{\textwidth}{!}{%
\begin{tabular}{ll*{9}{c}}
\toprule
\textbf{Harness} & \textbf{Model} & \makecell[c]{\textbf{Novelty}\\\textbf{Search}} & \makecell[c]{\textbf{Patent}\\\textbf{FTO}} & \makecell[c]{\textbf{Design}\\\textbf{FTO}} & \makecell[c]{\textbf{Invalidity}\\\textbf{Search}} & \makecell[c]{\textbf{Claim}\\\textbf{Charting}} & \makecell[c]{\textbf{Patent}\\\textbf{Drafting}} & \makecell[c]{\textbf{Office Action}\\\textbf{Response}} & \makecell[c]{\textbf{Patent}\\\textbf{Translation}} & \textbf{AVG} \\
\midrule
Codex & GPT-5.6-Sol & 27.5 & 54.2 & \cellcolor[HTML]{BDD3EB}\textbf{0.4} & 26.4 & \cellcolor[HTML]{DEE9F5}53.3 & \cellcolor[HTML]{BDD3EB}\textbf{84.7} & \cellcolor[HTML]{BDD3EB}\textbf{34.8} & 69.0 & \cellcolor[HTML]{DEE9F5}43.8 \\
 & Qwen3.8-Max & \cellcolor[HTML]{DEE9F5}51.3 & \cellcolor[HTML]{DEE9F5}55.5 & 0.0 & \cellcolor[HTML]{DEE9F5}31.2 & \cellcolor[HTML]{BDD3EB}\textbf{54.9} & 31.7 & 12.4 & 69.1 & 38.3 \\
 & Kimi K3 & 13.8 & 5.7 & 0.0 & 4.6 & 48.2 & 67.7 & 6.5 & \cellcolor[HTML]{BDD3EB}\textbf{70.5} & 27.1 \\
 & DeepSeek-V4-Pro & 10.0 & 43.2 & 0.0 & 10.3 & 9.3 & 58.2 & 6.4 & \cellcolor[HTML]{BDD3EB}\textbf{70.5} & 26.0 \\
 & MiniMax-M3 & 15.0 & 8.3 & 0.0 & 0.8 & 30.3 & 17.9 & 0.0 & 67.9 & 17.5 \\
 & GLM-5.3 & \cellcolor[HTML]{BDD3EB}\textbf{86.3} & \cellcolor[HTML]{BDD3EB}\textbf{74.0} & 0.0 & \cellcolor[HTML]{BDD3EB}\textbf{33.8} & 24.8 & \cellcolor[HTML]{DEE9F5}74.5 & \cellcolor[HTML]{DEE9F5}21.0 & \cellcolor[HTML]{DEE9F5}70.1 & \cellcolor[HTML]{BDD3EB}\textbf{48.1} \\
\midrule
Claude Code & GPT-5.6-Sol & 38.8 & 79.4 & 0.0 & \cellcolor[HTML]{DEE9F5}31.8 & \cellcolor[HTML]{BDD3EB}\textbf{54.1} & \cellcolor[HTML]{BDD3EB}\textbf{84.0} & \cellcolor[HTML]{BDD3EB}\textbf{38.0} & 68.5 & \cellcolor[HTML]{DEE9F5}49.3 \\
 & Claude Opus 5 & \cellcolor[HTML]{BDD3EB}\textbf{52.5} & \cellcolor[HTML]{BDD3EB}\textbf{86.5} & \cellcolor[HTML]{DEE9F5}2.7 & 28.7 & \cellcolor[HTML]{DEE9F5}54.0 & 82.8 & \cellcolor[HTML]{DEE9F5}25.1 & 70.2 & \cellcolor[HTML]{BDD3EB}\textbf{50.3} \\
 & Qwen3.8-Max & 32.5 & \cellcolor[HTML]{DEE9F5}86.2 & \cellcolor[HTML]{BDD3EB}\textbf{3.8} & 21.0 & 50.7 & 80.8 & 16.4 & 70.2 & 45.2 \\
 & Kimi K3 & 33.8 & 83.2 & 0.0 & 25.8 & 31.8 & \cellcolor[HTML]{DEE9F5}82.9 & 12.2 & \cellcolor[HTML]{DEE9F5}71.3 & 42.6 \\
 & DeepSeek-V4-Pro & 37.5 & 78.2 & 0.0 & 24.9 & 40.8 & 43.0 & 22.3 & 70.7 & 39.7 \\
 & MiniMax-M3 & 27.5 & 72.5 & 0.0 & 12.4 & 21.1 & 72.4 & 0.5 & 67.3 & 34.2 \\
 & GLM-5.3 & \cellcolor[HTML]{DEE9F5}50.0 & 83.9 & 0.0 & \cellcolor[HTML]{BDD3EB}\textbf{32.4} & 5.5 & 82.3 & 0.0 & \cellcolor[HTML]{BDD3EB}\textbf{71.4} & 40.7 \\
\midrule
Hermes & GPT-5.6-Sol & \cellcolor[HTML]{BDD3EB}\textbf{62.5} & 62.7 & 0.0 & 10.3 & \cellcolor[HTML]{DEE9F5}12.7 & 56.7 & \cellcolor[HTML]{BDD3EB}\textbf{54.8} & 69.1 & 41.1 \\
 & Claude Opus 5 & \cellcolor[HTML]{DEE9F5}52.5 & \cellcolor[HTML]{BDD3EB}\textbf{86.4} & \cellcolor[HTML]{BDD3EB}\textbf{1.9} & \cellcolor[HTML]{BDD3EB}\textbf{55.0} & \cellcolor[HTML]{BDD3EB}\textbf{55.1} & \cellcolor[HTML]{BDD3EB}\textbf{81.0} & \cellcolor[HTML]{DEE9F5}20.4 & \cellcolor[HTML]{BDD3EB}\textbf{70.3} & \cellcolor[HTML]{BDD3EB}\textbf{52.8} \\
 & Qwen3.8-Max & 17.5 & 80.5 & 0.0 & 20.5 & 0.0 & 42.7 & 0.4 & 69.3 & 28.9 \\
 & Kimi K3 & 36.3 & 82.7 & 0.0 & 34.5 & 3.2 & 73.2 & \cellcolor[HTML]{DEE9F5}20.4 & \cellcolor[HTML]{DEE9F5}69.7 & 40.0 \\
 & DeepSeek-V4-Pro & 22.5 & 61.2 & 0.0 & 29.9 & 8.3 & 67.2 & 4.7 & \cellcolor[HTML]{BDD3EB}\textbf{70.3} & 33.0 \\
 & MiniMax-M3 & 11.3 & 40.0 & 0.0 & 6.0 & 8.4 & 58.1 & 10.2 & 67.7 & 25.2 \\
 & GLM-5.3 & 43.8 & \cellcolor[HTML]{DEE9F5}86.2 & 0.0 & \cellcolor[HTML]{DEE9F5}40.1 & 0.0 & \cellcolor[HTML]{DEE9F5}77.4 & 17.0 & \cellcolor[HTML]{DEE9F5}69.7 & \cellcolor[HTML]{DEE9F5}41.8 \\
\midrule
OpenClaw & GPT-5.6-Sol & \cellcolor[HTML]{DEE9F5}46.3 & 82.2 & 0.0 & \cellcolor[HTML]{BDD3EB}\textbf{24.8} & \cellcolor[HTML]{BDD3EB}\textbf{54.7} & \cellcolor[HTML]{BDD3EB}\textbf{80.5} & \cellcolor[HTML]{BDD3EB}\textbf{42.7} & 68.6 & \cellcolor[HTML]{BDD3EB}\textbf{50.0} \\
 & Claude Opus 5 & 5.0 & 76.8 & \cellcolor[HTML]{BDD3EB}\textbf{1.3} & 7.7 & \cellcolor[HTML]{DEE9F5}16.3 & 71.7 & 1.1 & 70.0 & 31.2 \\
 & Qwen3.8-Max & 36.3 & \cellcolor[HTML]{BDD3EB}\textbf{84.2} & \cellcolor[HTML]{DEE9F5}0.4 & 17.3 & 3.7 & \cellcolor[HTML]{DEE9F5}78.9 & 2.9 & 68.9 & \cellcolor[HTML]{DEE9F5}36.6 \\
 & Kimi K3 & 40.0 & \cellcolor[HTML]{DEE9F5}83.0 & \cellcolor[HTML]{DEE9F5}0.4 & 17.7 & 5.8 & 58.9 & 13.6 & 70.3 & 36.2 \\
 & DeepSeek-V4-Pro & 23.8 & 69.9 & 0.0 & 6.5 & 0.0 & 53.2 & \cellcolor[HTML]{DEE9F5}24.6 & \cellcolor[HTML]{DEE9F5}70.6 & 31.1 \\
 & MiniMax-M3 & 28.8 & 77.5 & 0.0 & 10.5 & 0.0 & 63.1 & 15.3 & 67.0 & 32.8 \\
 & GLM-5.3 & \cellcolor[HTML]{BDD3EB}\textbf{57.5} & 82.5 & 0.0 & \cellcolor[HTML]{DEE9F5}21.3 & 0.0 & 1.4 & 1.5 & \cellcolor[HTML]{BDD3EB}\textbf{71.5} & 29.5 \\
\bottomrule
\end{tabular}%
}
\end{table}

We demonstrate the results in Table~\ref{tab:main_results}.  Claude Opus 5 with Hermes achieves the highest average score of 52.8, followed by the same model with Claude Code at 50.3. If the harness is held constant, such as Hermes, Claude Opus 5 outperforms GPT-5.6-Sol on invalidity search, claim charting, and drafting, while GPT performs better on novelty search and OAR. 
A shared weakness is Design FTO, where recall remains below 4 across every evaluated configuration. We find that the ability to extract visual features and search for similar designs is not well supported by the current LLMs and harnesses. High recall on textual FTO does not extend to this visual search setting, making design-related evidence discovery a persistent gap.

The results show that neither model rankings nor harness advantages transfer consistently across configurations and tasks. GLM-5.3 outperforms GPT-5.6-Sol under Codex, with average scores of 48.1 and 43.8, but the ordering reverses under OpenClaw, where GPT scores 50.0 and GLM scores 29.5. Model comparisons therefore depend on the harness used to run them. Harness advantages are also task-specific: switching Opus from Claude Code to Hermes raises invalidity-search recall from 28.7\% to 55.0\%, a gain of 26.3 percentage points, while leaving FTO, claim-charting, and drafting scores comparatively similar. This gain establishes a strong pairing for invalidity search, rather than a general Hermes advantage across patent tasks. GPT-5.6-Sol performs comparably with OpenClaw and Claude Code, with average scores of 50.0 and 49.3, respectively, while Hermes yields its highest OAR score. Consequently, the highest-ranked configuration overall is not necessarily the strongest choice for an individual workflow. Task-level comparisons are necessary to identify the appropriate model–harness configuration for a given patent task. More detailed experimental results and analysis are provided in Appendix~\ref{app:detailed_results}.

\subsection{In-depth Analysis}

\label{sec:in_depth_analysis}

\subsubsection{Model--Harness Interaction}
\label{sec:model_harness}
\begin{figure}[!htb]
    \centering
    \captionsetup[subfigure]{justification=centering,singlelinecheck=false}
    \begin{subfigure}[b]{0.49\textwidth}
        \centering
        \includegraphics[width=\linewidth]{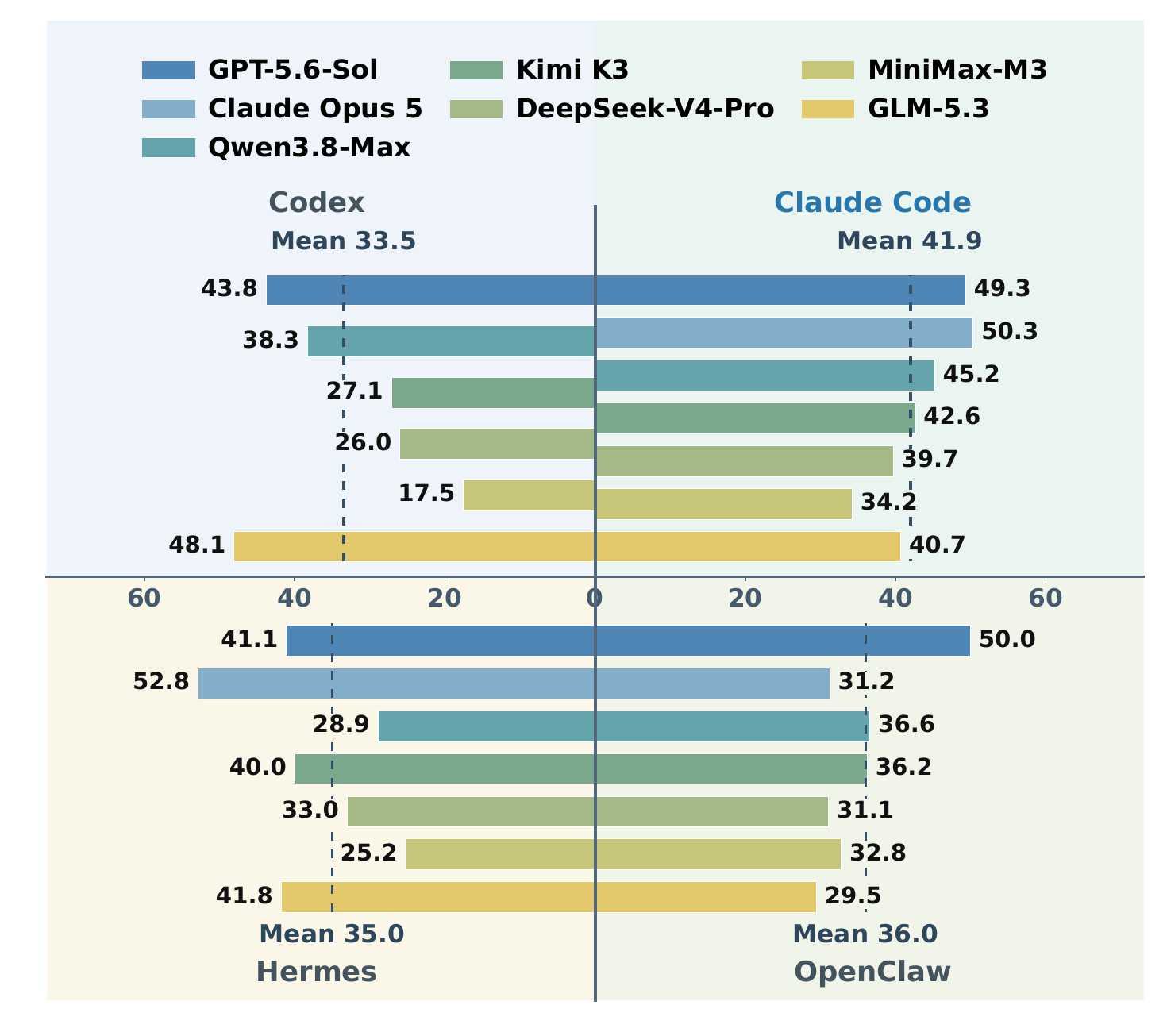}
        \caption{Model-specific harness performance}
        \label{fig:model_harness_performance}
    \end{subfigure}\hfill
    \begin{subfigure}[b]{0.49\textwidth}
        \centering
        \includegraphics[width=\linewidth]{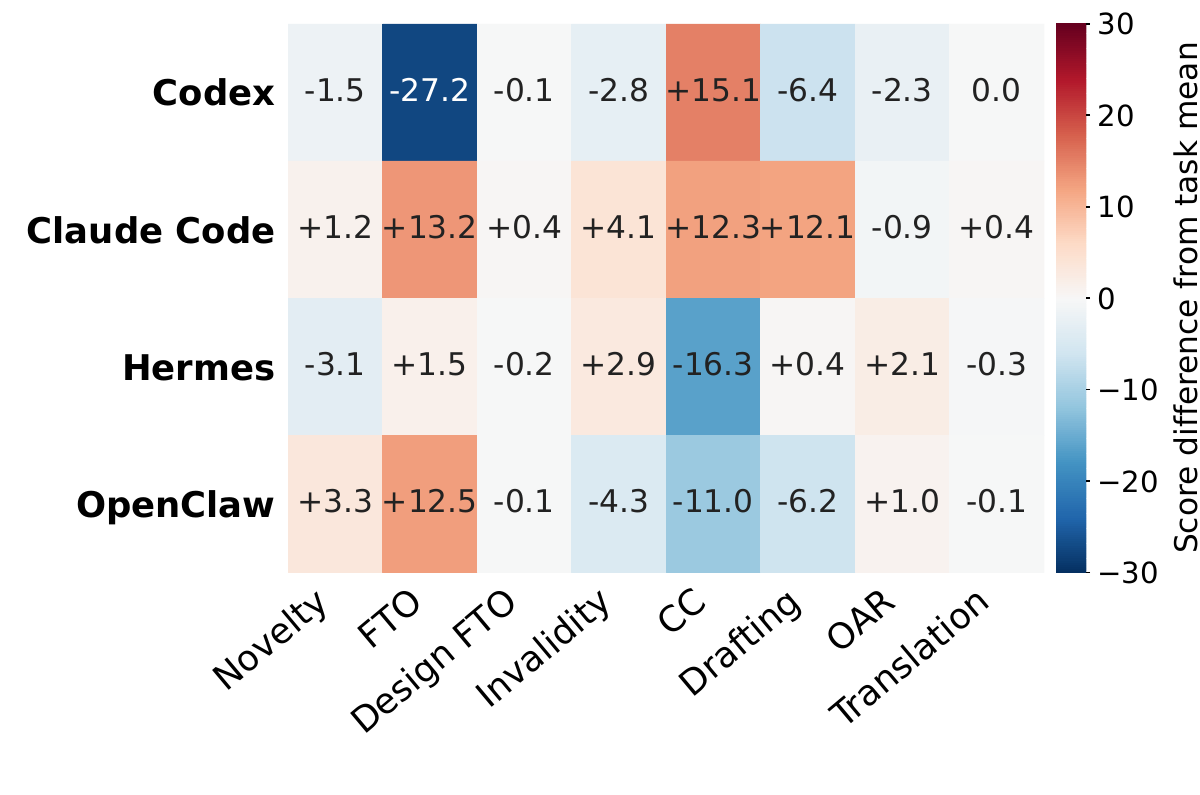}
        \caption{Task-specific advantage: six complete models}
        \label{fig:model_harness_task_advantage}
    \end{subfigure}
  \caption{(a) Eight-task average score by model and harness. Dashed lines show harness means over the same six models evaluated under all four harnesses, excluding Opus. Individual bars show all evaluated pairs. (b) Task scores averaged over these six models, centered by subtracting the four-harness mean within each task.}
    \label{fig:model_harness}
\end{figure}

\noindent\textbf{Finding: Under the same models, Claude Code achieves the strongest overall performance and leads on most tasks.} Figure~\ref{fig:model_harness_performance} compares model--harness pairs, with harness means calculated over the six models evaluated under all four harnesses. Claude Code achieves the highest mean score of 41.9, followed by OpenClaw at 36.0, Hermes at 35.0, and Codex at 33.5. It also achieves the highest average score for four of the seven evaluated models: Qwen3.8-Max, Kimi K3, DeepSeek-V4-Pro, and MiniMax-M3. GPT-5.6-Sol performs comparably with Claude Code and OpenClaw, while Opus performs best with Hermes and GLM-5.3 performs best with Codex. Claude Code therefore provides the strongest aggregate performance across the common models.

Figure~\ref{fig:model_harness_task_advantage} shows that this lead is achieved by five of the eight tasks. OpenClaw has the second-highest average score across the six common models. Claude Code outperforms it by 23.3 points on claim charting and 18.2 points on patent drafting. Together, these two tasks account for 5.2 points of its 5.9-point average-score advantage. Its FTO recall also reaches 80.6, compared with 40.1 for Codex, showing a substantial advantage over that harness in textual risk retrieval. However, Codex leads on claim charting, OpenClaw leads on novelty search, and Hermes leads on OAR. Translation scores remain close across harnesses, and all remain weak on design FTO. Claude Code's overall lead is thus supported by stronger performance across several workflows, while task-level comparisons remain necessary when selecting a harness.

\Needspace{0.30\textheight}
\subsubsection{Prior-Art Search and Evidence Alignment}
\label{sec:evidence_analysis}

\begin{figure}[H]
    \captionsetup{font=small}
    \centering
    \captionsetup[subfigure]{justification=centering,singlelinecheck=false}
    \begin{subfigure}[t]{0.49\linewidth}
        \centering
        \includegraphics[width=\linewidth]{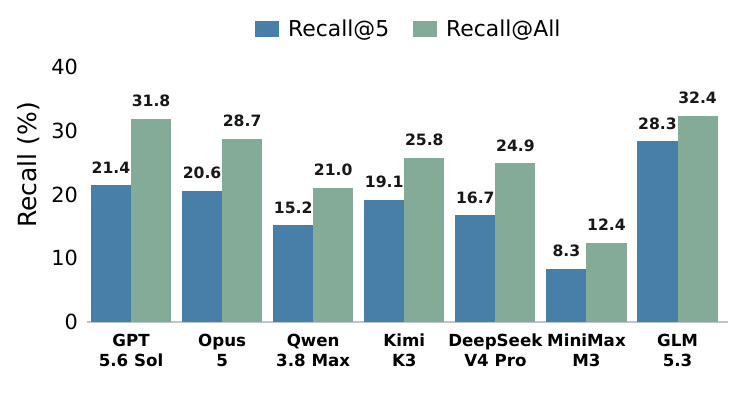}
        \caption{Invalidity-search recall}
        \label{fig:evidence_retrieval}
    \end{subfigure}
    \hfill
    \begin{subfigure}[t]{0.49\linewidth}
        \centering
        \includegraphics[width=\linewidth]{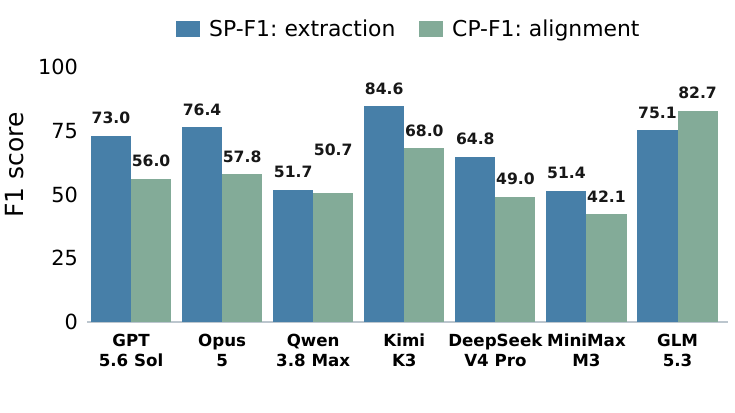}
        \caption{Feature extraction and evidence alignment}
        \label{fig:evidence_alignment}
    \end{subfigure}
   \caption{ (a) Recall@5 and Recall@All measure coverage of annotated references in the top five results and the full returned list. (b) SP-F1 measures technical feature extraction, and CP-F1 measures feature-to-evidence alignment in claim charting.}
    \label{fig:evidence_analysis}
\end{figure}

\noindent\textbf{Finding: Evidence retrieval is one dominant weakness, and evidence alignment is also weaker than feature extraction in majority of the models.} Using Claude Code as a fixed harness, invalidity search is limited mainly by missing references rather than their ranking. Figure~\ref{fig:evidence_retrieval} compares Recall@5, which measures reference coverage within the top five results, with Recall@All, which considers the full returned list. Recall@All remains below one-third for every model. Extending evaluation beyond the top five adds 4.0--10.4 points, leaving at least two-thirds of the annotated references unrecovered. For GPT-5.6-Sol, recall rises from 21.4\% to 31.8\%, but 68.2\% of references are still missing.

With comparison evidence supplied, six of seven models score lower on evidence alignment than on feature extraction in Figure~\ref{fig:evidence_alignment}. The gap is 17.0 points for GPT-5.6-Sol and 18.5 for Opus, using the same scored cases for each model's two metrics. These models identify technical features more accurately than they establish the corresponding evidence matches. These two analysis therefore locate concrete weaknesses in separate tasks: invalidity search fails mainly to recover the reference set, while claim charting scores lower on feature-to-evidence correspondence than on feature identification. 

\subsubsection{Patent Document Quality}
\label{sec:document_quality}

\begin{figure}[H]
    \centering
    \captionsetup[subfigure]{justification=centering,singlelinecheck=false}
    \begin{subfigure}[b]{0.49\linewidth}
        \centering
        \includegraphics[width=\linewidth]{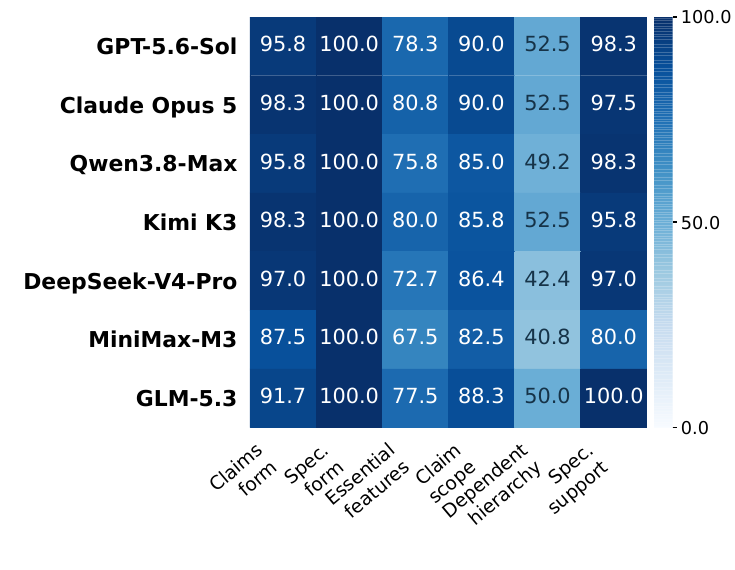}
        \caption{Drafting quality criteria}
        \label{fig:drafting_quality}
    \end{subfigure}
    \hfill
    \begin{subfigure}[b]{0.49\linewidth}
        \centering
        \includegraphics[width=\linewidth]{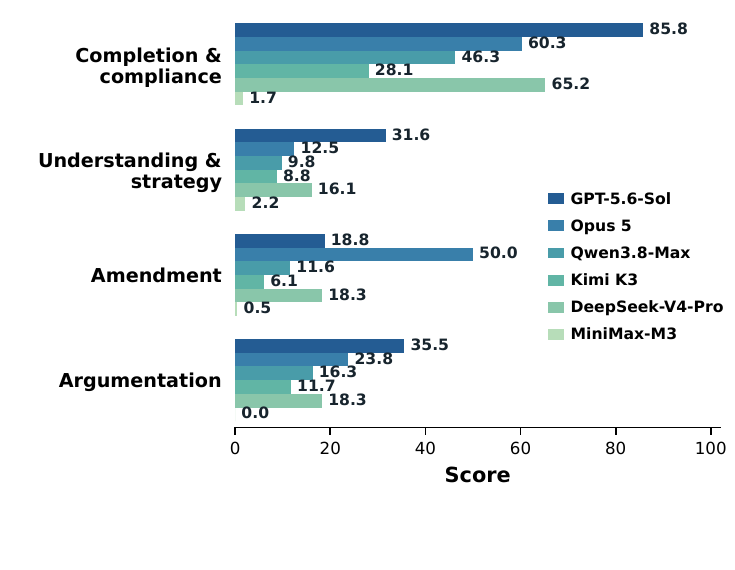}
        \caption{OAR quality dimensions}
        \label{fig:oar_quality}
    \end{subfigure}
\caption{Document quality under Claude Code. (a) Scores on six patent-drafting criteria. (b) Scores on four OAR criterion groups, computed within each case using the original weights of applicable criteria and then averaged over applicable scored cases. Amendment scores cover different case sets across models, all scores are on a 0--100 scale.}
    \label{fig:document_quality}
\end{figure}

\noindent\textbf{Finding: In patent document tasks, most of the models produce well-structured drafts and address response requirements, but struggle to organize dependent claims, understand case facts, and develop well-supported arguments.}
Under Claude Code, Figure~\ref{fig:drafting_quality} shows that all seven models score 100.0 on specification form and 87.5--98.3 on claim form. They produce the expected document sections and recognizable independent and dependent claims. However, dependent-claim hierarchy scores range from 40.8 to 52.5, the lowest displayed criterion for every model. This criterion assesses whether dependent claims introduce meaningful additional limitations and establish useful layers of narrower protection. GPT-5.6-Sol and Opus both score 52.5 on hierarchy despite claim-form scores of 95.8 and 98.3. The weakness lies in organizing the scope of protection: a formally complete claim set does not necessarily provide useful fallback positions for examination or amendment.

Figure~\ref{fig:oar_quality} examines four groups of OAR criteria: completion and compliance, case understanding and strategy, claim amendment, and substantive argumentation. Group scores retain the original criterion weights, exclude inapplicable criteria, and are averaged over applicable scored cases. Completion and compliance is the highest-scoring group for five of the six models with scored responses. GPT-5.6-Sol reaches 85.8 in this group, but scores only 31.6 on understanding and strategy and 35.5 on argumentation; DeepSeek scores 65.2, 16.1, and 18.3, respectively. These results show that completing the response requirements does not ensure an accurate reading of the case or a well-supported argument addressing the examiner's objections. Amendment quality also remains limited: five models score below 20.0, while Opus scores 50.0 over only four applicable cases, preventing a direct comparison across models. Together, the two tasks reveal a common limitation: models can assemble the expected document, but struggle with the professional judgments that make its claims and arguments effective.

\section{Case Study}
\label{sec:case_study}
\begin{figure}[!htb]
    \centering
    \captionsetup[subfigure]{justification=centering,singlelinecheck=false,skip=3pt}
    \begin{subfigure}{0.88\linewidth}
        \centering
        \includegraphics[width=\linewidth]{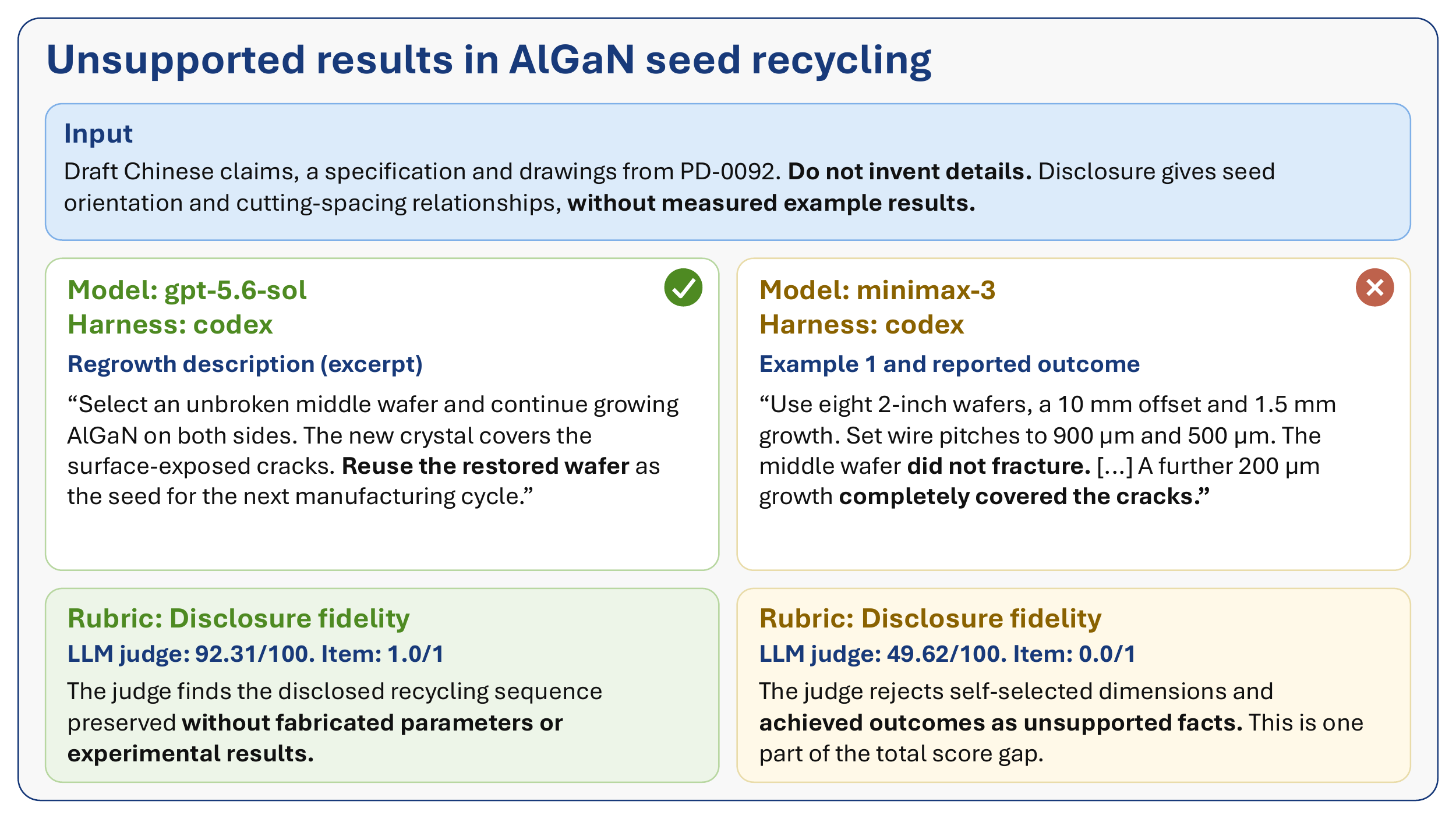}
        \caption{Fixed Codex harness: GPT-5.6-Sol versus MiniMax-M3.}
        \label{fig:main_case_fixed_harness}
    \end{subfigure}\par\vspace{0.6em}
    \begin{subfigure}{0.88\linewidth}
        \centering
        \includegraphics[width=\linewidth]{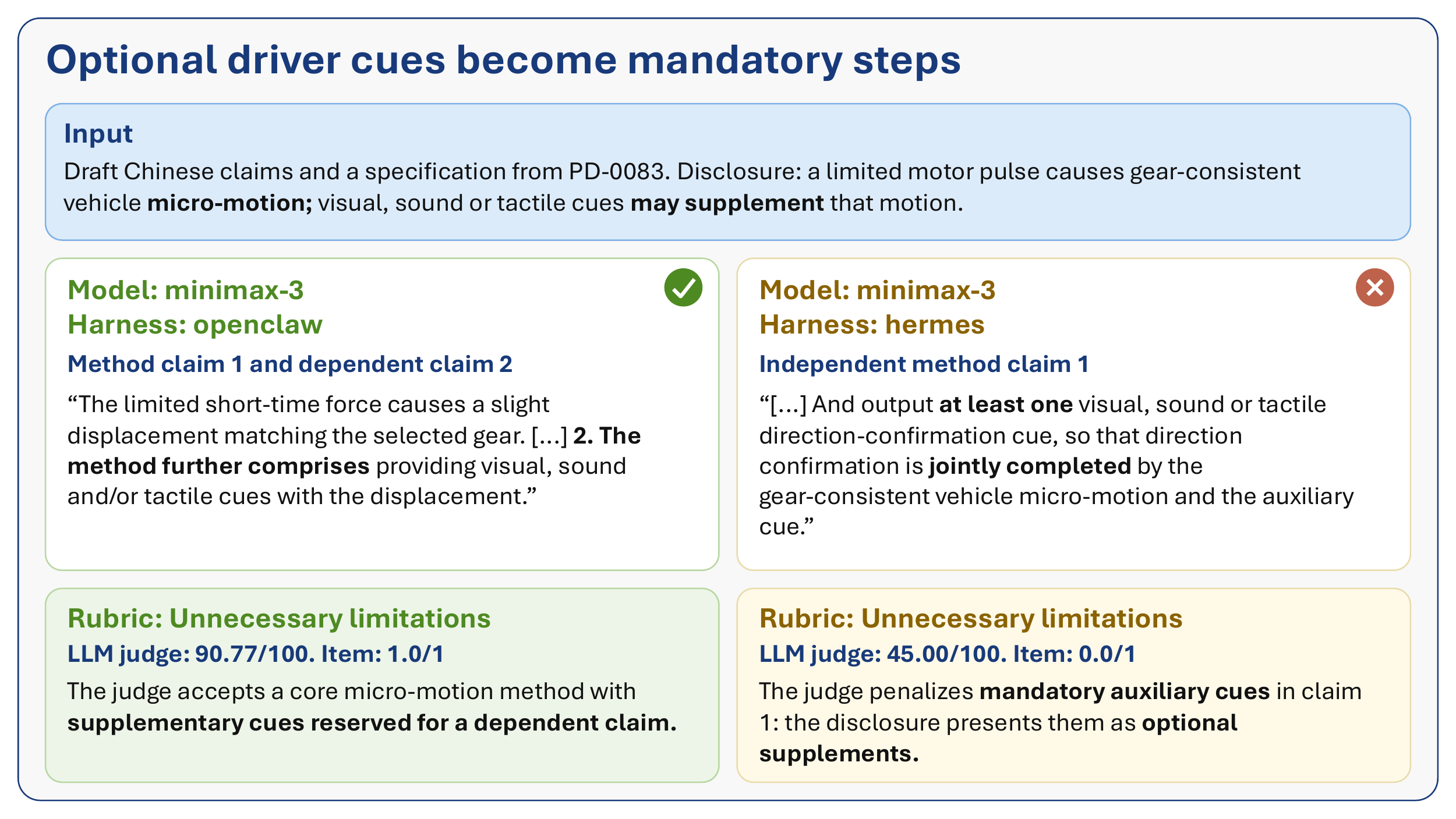}
        \caption{Fixed MiniMax-M3 model: OpenClaw versus Hermes.}
        \label{fig:main_case_fixed_model}
    \end{subfigure}
    \caption{Case studies of different model--harness pairs.}
    \label{fig:main_drafting_cases}
\end{figure}

We use two patent-drafting cases to illustrate how model and harness choices affect draft quality. Figure~\ref{fig:main_drafting_cases} compares models using the same harness and harnesses using the same model, highlighting differences in technical interpretation and claim drafting. Appendix~\ref{app:case_comparisons} provides further comparisons across seven task suites.

\noindent\textbf{With Codex as a fixed harness, MiniMax-M3 fabricates experimental results, whileGPT-5.6-Sol follows the given technical disclosure.}
In Figure~\ref{fig:main_case_fixed_harness}, the task is to draft an application for recycling AlGaN seed wafers. The disclosure describes cutting a thicker middle wafer, regrowing crystal on both sides to cover exposed cracks, and reusing the wafer in the next cycle. It provides process relationships, but no measured example results. GPT-5.6-Sol retains this sequence without claiming that an experiment has been performed. MiniMax-M3 instead supplies cutting pitches of 900 and 500~$\mu$m, reports that the wafer did not fracture, and states that a further 200~$\mu$m of growth completely covered the cracks. These statements turn the disclosed process into a report of successful manufacturing without supporting measurements. The fidelity rubric scores the two outputs at 1.0 and 0.0, respectively. The error is therefore an unsupported claim of demonstrated performance, not simply additional descriptive detail. A usable draft must preserve the distinction between how a process is intended to work and what the supplied evidence establishes.

\noindent\textbf{With MiniMax-M3 as a fixed model, Hermes makes an optional requirement mandatory, while OpenClaw preserves it as optional.}
In Figure~\ref{fig:main_case_fixed_model}, the invention confirms a vehicle's selected driving direction through a brief motor pulse that produces slight forward or backward motion. Visual, sound or tactile cues may supplement this physical feedback, but the disclosure does not require them. Under OpenClaw, MiniMax-M3 places motion-based confirmation in independent method claim 1 and adds the supplementary cues in dependent claim 2. The core method therefore remains applicable without an auxiliary cue. Under Hermes, the same model requires at least one such cue in independent method claim 1, excluding the disclosed implementation that uses vehicle motion alone. The unnecessary-limitation rubric assigns 1.0 to OpenClaw and 0.0 to Hermes. The harness comparison thus exposes a concrete change in claim scope: a supplementary feature becomes a condition of the core method. Preserving the disclosed components is insufficient if the draft changes which components are required.

\section{Conclusion}
We introduce PatentAgentBench, a benchmark for evaluating LLM agents on professional patent tasks spanning evidence retrieval, claim-level analysis, and document preparation. Built from actual patent records and professional materials, it combines task-specific metrics with expert-reviewed annotations and rubrics to assess the evidence and work products required in patent practice. Our evaluation shows that model and harness choices must be considered together, as their advantages vary across tasks. Current agents remain limited in retrieving relevant references and aligning technical features with supporting evidence. Under Claude Code, patent drafts exhibit strong formal organization but weak dependent-claim hierarchies. Office-action responses show weaknesses in case understanding, response strategy, and substantive argumentation, even when completion and compliance scores are high. The case studies further illustrate how unsupported factual additions and changes to disclosed requirements undermine draft quality. These findings show that reliable patent agents require both stronger evidence use and better control over substantive content. PatentAgentBench provides a common framework for evaluating progress toward these requirements.

\section{Limitations}
PatentAgentBench covers 8 professional patent tasks. However, tasks are evaluated separately rather than as a complete patent lifecycle.  The evaluation process focus on the final output or delivery of agents rather than the agent working process. In the future, we plan to extend the benchmark to include more tasks and evaluate the complete patent lifecycle.

{
    \small
    \bibliographystyle{colm2026_conference}
    \bibliography{references}
}

\clearpage
\section*{Contribution List}
\subsection*{Project Lead}
\noindent\begin{tabular}{@{}ll@{}}
\textbf{Lisha Zhang} & \href{mailto:zhanglisha@patsnap.com}{\texttt{zhanglisha@patsnap.com}}\\
\textbf{Kai Duan} & \href{mailto:duankai@patsnap.com}{\texttt{duankai@patsnap.com}}
\end{tabular}\par\vspace{4pt}
Responsible for overall project direction, resource coordination, technical guidance, and final review of the technical report.

\subsection*{Core Contributors}
\noindent\begin{tabular}{@{}ll@{}}
\textbf{Yijie Hu} & \href{mailto:huyijie@patsnap.com}{\texttt{huyijie@patsnap.com}}\\
\textbf{Wei Li} & \href{mailto:liwei@patsnap.com}{\texttt{liwei@patsnap.com}}\\
\textbf{Junjie Zhou} & \href{mailto:zhoujunjie@patsnap.com}{\texttt{zhoujunjie@patsnap.com}}\\
\textbf{Zhiqi Jiang} & \href{mailto:jiangzhiqi@patsnap.com}{\texttt{jiangzhiqi@patsnap.com}}\\
\textbf{Yi Jin} & \href{mailto:jinyi@patsnap.com}{\texttt{jinyi@patsnap.com}}\\
\textbf{Licong Xu} & \href{mailto:xulicong@patsnap.com}{\texttt{xulicong@patsnap.com}}\\
\textbf{Xinkai Sun} & \href{mailto:sunxinkai@patsnap.com}{\texttt{sunxinkai@patsnap.com}}
\end{tabular}\par\vspace{4pt}
Responsible for task and experimental design, data collection and annotation, data quality assurance, experimental infrastructure and execution, paper writing, and open-source development.

\clearpage
\appendix

\section{Experimental Setting Details}
\label{experimental_details}

\subsection{Agent Harnesses}
An agent harness manages the interaction between a model, its task context, and tool calls during execution. We evaluate four general-purpose harnesses with the following versions:
\begin{itemize}
    \item \href{https://github.com/openai/codex}{\textbf{Codex CLI}}, version \texttt{0.151.0}, is OpenAI's terminal-based coding agent. It provides a local interface for carrying out tasks through model interactions and tool calls.
    \item \href{https://github.com/anthropics/claude-code}{\textbf{Claude Code}}, version \texttt{2.1.251}, is Anthropic's agentic coding environment. It accepts natural-language instructions and coordinates actions within a working context.
    \item \href{https://github.com/openclaw/openclaw}{\textbf{OpenClaw}}, version \texttt{2026.8.1}, is an open-source personal assistant framework that connects configurable models with an agent runtime and external integrations.
    \item \href{https://github.com/NousResearch/hermes-agent}{\textbf{Hermes Agent}}, version \texttt{0.21.1}, is Nous Research's agent framework with a terminal interface, support for multiple model providers, and tool-calling workflows.
\end{itemize}
The main table abbreviates Codex CLI as Codex and Hermes Agent as Hermes.

\subsection{Models and Execution Protocol}
We evaluate GPT-5.6-Sol, Claude Opus 5, Kimi K3, Qwen3.8-Max, MiniMax-M3, GLM-5.3, and DeepSeek-V4-Pro. A configuration is a model--harness pair. Claude Opus 5 was not evaluated with Codex because the model interface was incompatible with the protocol supported by our Codex configuration. This untested configuration is omitted from the main table rather than scored as zero. Comparisons across all four harnesses use the six models with complete results; comparisons including Opus use only the harnesses for which its results are available.

Each configuration receives the task input and evidence conditions defined in Section~\ref{bench_feature}. An initial execution failure permits up to three retries, giving a maximum of four attempts per instance. The task score is computed over the full assigned case set, including cases that remain unsuccessful after retries.

\subsection{Tools and Judge Model}
We provide a commercial patent novelty-search service through the Model Context Protocol (MCP), which agents can call to retrieve patent evidence. This MCP service is the only tool supplied by the benchmark; we provide no additional tools. The harness controls how the model invokes the service during task execution.

For evaluation components that require an LLM judge, including rubric-based patent drafting and OAR assessment, we use \textbf{GPT 6 sol}. The judge scores the submitted work against the case materials and evaluation rubrics. Reference matching and the automatic metrics defined below are computed by their corresponding scorers.

\clearpage
\section{Expert Team Profile}
\label{app:expert_profile}

Table~\ref{tab:expert_profile} summarizes the professional
backgrounds of the 31 experts involved in
data curation and evaluation design.

\begin{table}[t]
\centering
\caption{Professional backgrounds of the expert team.}
\label{tab:expert_profile}
\begin{tabular}{@{}lc@{}}
\toprule
\textbf{Professional background} & \textbf{Count} \\
\midrule
Senior patent examiner & 5 \\
Senior patent agent & 15 \\
Dual-qualified lawyer & 3 \\
Senior IPR specialist & 8 \\
\midrule
\textbf{Total} & \textbf{31} \\
\bottomrule
\end{tabular}
\end{table}

\section{Dataset Details}
\label{app:dataset_details}
Table~\ref{tab:dataset_details} summarizes the language, jurisdiction, evaluation metrics, and rubric coverage of each task suite. Language denotes the task input or required output; design FTO uses image inputs. The benchmark contains 510 cases across eight suites. Translation contains 80 cases with 881 evaluation items. Metric definitions and aggregation rules are provided in Appendix~\ref{metric_details}.

\begin{table}[H]
\centering
\caption{Detailed dataset coverage and evaluation. ZH and EN denote Chinese and English. Jurisdiction codes identify the target jurisdiction or source patent office: CN (China), US (United States), EP (European Patent Office), EU (European Union), and WO (PCT international publication). Translation coverage follows the source patent publication numbers. Rubric counts are criterion entries summed across cases, including repeated common criteria.}
\label{tab:dataset_details}
\small
\renewcommand{\arraystretch}{1.18}
\setlength{\tabcolsep}{5pt}
\resizebox{\textwidth}{!}{%
\begin{tabular}{@{}lccclc@{}}
\toprule
\textbf{Task suite} & \textbf{Cases} & \textbf{Language} & \textbf{Jurisdiction} & \textbf{Evaluation metrics} & \makecell{\textbf{Rubric}\\\textbf{entries}} \\
\midrule
Novelty Search & 80 & ZH/EN & CN/US/EP/WO & Hit@100 & 0 \\
Patent FTO & 60 & ZH/EN & CN/US/EP & Recall@All & 0 \\
Design FTO & 80 & Image input & CN/EU & Recall@200 & 0 \\
Invalidity Search & 80 & ZH/EN & CN/US & Recall@All & 0 \\
Claim Charting & 30 & ZH/EN & CN/US/EP & CP-F1; supplementary SP-F1 & 0 \\
Patent Drafting & 60 & ZH/EN & CN/US/EP & Weighted rubrics; pass rate & 977 \\
OAR & 40 & ZH/EN & CN/US/EP & \makecell[l]{Weighted rubrics;\\pass rate} & 57 \\
Translation & 80 & ZH$\leftrightarrow$EN & CN/US/EP/WO & \makecell[l]{BLEU-1, METEOR, ROUGE-1, SacreBLEU;\\terminology, character, and writing checks} & 0 \\
\midrule
\textbf{Total} & \textbf{510} & & & & \\
\bottomrule
\end{tabular}%
}
\end{table}

Table~\ref{tab:rubric_inventory} reports the drafting rubric inventory: 15 common criteria per case and 77 case-specific entries across the suite. OAR uses 57 rubric entries; its scored dimensions, applicability rules, and aggregation are defined below. The drafting inventory counts criterion instances across cases, whereas the OAR count describes the rubric inventory, so these counts are not summed. Pass rates use the expert-defined thresholds reported with each task's detailed results.

\begin{table}[H]
\centering
\caption{Patent-drafting rubric inventory across cases. Common criteria are counted once per case in which they are provided.}
\label{tab:rubric_inventory}
\small
\begin{tabular}{@{}lrrr@{}}
\toprule
\textbf{Task suite} & \textbf{Common} & \textbf{Case-specific} & \textbf{Total} \\
\midrule
Patent Drafting & 900 & 77 & 977 \\
\bottomrule
\end{tabular}
\end{table}

\section{Evaluation Metric Definitions}
\label{metric_details}

\subsection{Notation and Aggregation}
Let $N_t$ be the number of assigned cases in task $t$, and let $s_{ti}\in[0,1]$ be the primary score for case $i$. Let $c$ be the reporting scale factor, with $c=100$. Every task score is reported on a 0--100 scale:
\begin{equation}
    S_t=\frac{c}{N_t}\sum_{i=1}^{N_t}s_{ti}.
    \label{eq:task_macro_score}
\end{equation}
For the primary task scores in Table~\ref{tab:main_results}, we include all assigned cases, with failed cases contributing zero. For drafting and OAR, a case score originally expressed on a 0--100 scale is divided by 100 before applying Eq.~\ref{eq:task_macro_score}. Let $T$ be the number of task suites, with $T=8$. The overall mean score $\bar{S}$, reported as AVG in the result tables, is
\begin{equation}
    \bar{S}=\frac{1}{T}\sum_{t=1}^{T}S_t.
    \label{eq:overall_average}
\end{equation}
Translation contributes one task score after averaging its two directions. An unavailable task result is distinct from an observed zero: AVG is not computed when any task result, or either translation direction, is missing.

We distinguish overall task performance from the quality of scored documents. The main results use all assigned cases. Quality means and document-level analyses use only scored outputs. Let $A_t$ be this scored set, $n_t=|A_t|$, and $q_{ti}$ the document quality score on a 0--100 scale. For $n_t>0$,
\begin{equation}
 Q_t=\frac{1}{n_t}\sum_{i\in A_t}q_{ti},\qquad S_t=\frac{n_t}{N_t}Q_t.
\end{equation}
Here $Q_t$ describes scored documents, whereas $S_t$ includes unscored cases as zero. The equality relates means of the same quality score; individual rubric dimensions are not task totals. Figure~\ref{fig:document_quality} and the appendix quality analyses use the scored-output denominator to compare document properties. These analyses describe the observed documents, not the frequency of completing a task. Pass rates use all $N_t$ assigned cases, with unscored cases counted as not passing. When $n_t=0$, conditional quality is undefined; the appendix displays 0.0 as a placeholder.

For retrieval, let $G_i$ be the deduplicated reference set and $R_i^{(K)}$ the set of distinct references in the first $K$ returned results. Let $R_i^{(\mathrm{all})}$ denote the full returned list after matching and deduplication. Matching uses the task-specific identifiers described below.

\subsection{Novelty Search}
The primary metric is Hit@$K$, with cutoff $K=100$. A case receives credit if at least one reference patent family is retrieved:
\begin{equation}
    s_{\mathrm{nov},i}=\mathbf{1}\!\left[
    G_i\cap R_i^{(K)}\ne\varnothing\right].
    \label{eq:novelty_hit}
\end{equation}
Publication identifiers are mapped to patent families before matching. The task score is the percentage of cases with a hit, rather than the mean fraction of all reference families recovered. Supplementary Hit@$K$ uses the same definition with a different cutoff; supplementary Recall@$K$ uses $|G_i\cap R_i^{(K)}|/|G_i|$.

\subsection{Patent FTO}
Patent FTO uses macro recall over the full returned list:
\begin{equation}
    s_{\mathrm{fto},i}=
    \frac{|G_i\cap R_i^{(\mathrm{all})}|}{|G_i|}.
    \label{eq:fto_recall}
\end{equation}
Matching is performed at the patent-application level. Equation~\ref{eq:task_macro_score} gives every case equal weight, regardless of how many reference patents it contains.

\subsection{Design FTO}
The primary metric is macro Recall@$K$, with cutoff $K=200$:
\begin{equation}
    s_{\mathrm{design},i}=
    \frac{|G_i\cap R_i^{(K)}|}{|G_i|}.
    \label{eq:design_recall}
\end{equation}
Only references found within the first 200 results contribute to the numerator. We retain the design-patent matching rules of the reference annotations.

\subsection{Invalidity Search}
Invalidity search uses Recall@All:
\begin{equation}
    s_{\mathrm{inv},i}=
    \frac{|G_i\cap R_i^{(\mathrm{all})}|}{|G_i|}.
    \label{eq:invalidity_recall}
\end{equation}
References are matched at the patent-identifier level. The score measures recovery of the reference evidence set; the associated statutory grounds and affected claims remain available for case-level analysis.

\subsection{Claim Charting}
The primary metric is CP-F1 for feature-level comparisons. Let $a_i$, $b_i$, and $c_i$ be the numbers of correct, unmatched predicted, and missed reference comparison items under the annotated matching rules. Precision and recall are
\begin{equation}
    P_i=\frac{a_i}{a_i+b_i},
    \qquad
    R_i=\frac{a_i}{a_i+c_i}.
\end{equation}
The case score is
\begin{equation}
    s_{\mathrm{cc},i}=\begin{cases}
        \dfrac{2P_iR_i}{P_i+R_i}, & P_i+R_i>0,\\[4pt]
        0, & \text{otherwise}.
    \end{cases}
    \label{eq:cc_fone}
\end{equation}
A precision or recall term with a zero denominator is set to zero. We average case-level F1 scores, rather than pooling comparison items across cases. Supplementary SP-F1 applies the same precision--recall calculation to feature-decomposition items and does not enter AVG separately.

\subsection{Patent Drafting}
Let $\mathcal{J}_i$ contain the applicable general and case-specific quality criteria for case $i$. Criterion $j$ has its prescribed weight $w_{ij}$ and a judge score $u_{ij}$ on the rubric's normalized scale. The case quality score is
\begin{equation}
    s_{\mathrm{draft},i}=
    \frac{\sum_{j\in\mathcal{J}_i}w_{ij}u_{ij}}
         {\sum_{j\in\mathcal{J}_i}w_{ij}}.
    \label{eq:drafting_quality}
\end{equation}
General and case-specific criteria share the same weighted sum; their relative contribution depends on the case's configured weights. Zero-weight eligibility and unity checks do not contribute to this quality score. The task score averages case quality over the full drafting set.

\subsection{Office Action Response}
\begingroup
OAR uses rubric-based assessment of the submitted response against the case materials. Applicable criteria receive scores $u_{ij}\in\{0,0.5,1\}$. The base weight pools allocate 25\% to basic quality, 20\% to amendment quality, and 55\% to substantive argumentation. 

Let $\mathcal{A}_i$ denote the applicable scored criteria for case $i$ and $w_{ij}$ their final allocated weights. The normalized case score is
\begin{equation}
 s_{\mathrm{oar},i}=\frac{\sum_{j\in\mathcal{A}_i}w_{ij}u_{ij}}{\sum_{j\in\mathcal{A}_i}w_{ij}}.
 \label{eq:oar_combined}
\end{equation}
Let $q_i\in[0,1]$ denote the recorded case quality after normalization. The recorded score is obtained by expressing $s_{\mathrm{oar},i}$ on the 0--100 scale, rounding to one decimal place, and converting back to the unit interval. The primary task score retains all assigned cases, with unscored cases contributing zero. In the detailed results, quality is the mean of $q_i$ over the scored set $\mathcal{D}$, and pass rate is
\begin{equation}
 Q=\frac{1}{|\mathcal{D}|}\sum_{i\in\mathcal{D}}q_i,
 \qquad P=\frac{1}{N}\sum_{i\in\mathcal{D}}\mathbf{1}[q_i\geq\tau],
 \label{eq:oar_quality_pass}
\end{equation}
where $N=40$ is the number of assigned cases and $\tau=0.8$ is the expert-defined usability threshold. Unscored cases do not pass.

For group-level analysis, we organize the criteria into four groups. Completion and compliance covers objection coverage, deliverable completeness and text quality, and jurisdictional compliance. Case understanding and strategy covers understanding the case, choosing a response strategy, logical consistency, and factual and citation accuracy. Amendment covers whether changes address the objections, preserve appropriate claim scope, and satisfy legal and formal requirements. Argumentation covers the substantive response to inventive step, novelty, eligibility, disclosure, clarity and support, and essential-feature objections. Let $\mathcal{B}_k$ contain the criteria in group $k$. Define $\mathcal{A}_{ik}=\mathcal{A}_i\cap\mathcal{B}_k$ and let $\mathcal{D}_k$ contain scored cases with positive total weight in that group. The group score is
\begin{equation}
 R_k=\frac{1}{|\mathcal{D}_k|}\sum_{i\in\mathcal{D}_k}
 \frac{\sum_{j\in\mathcal{A}_{ik}}w_{ij}u_{ij}}{\sum_{j\in\mathcal{A}_{ik}}w_{ij}}.
 \label{eq:oar_group_quality}
\end{equation}
The quantities $Q$, $P$, and $R_k$ lie in the unit interval. For presentation, quality and group scores are converted to the 0--100 scale, and pass rates are expressed as percentages. Each applicable case contributes equally after within-case weighting. For criterion-level tables, each criterion is averaged over its own applicable scored cases. Different applicable case sets must be considered when comparing means. 
\endgroup

\subsection{Patent Translation}
Let $q_{di}\in[0,1]$ be the score of instance $i$ in direction $d$, where $d=1$ denotes Chinese-to-English translation and $d=2$ denotes English-to-Chinese translation. Let $b_{di}$, $m_{di}$, $r_{di}$, and $z_{di}$ denote its BLEU-1, METEOR, ROUGE-1, and SacreBLEU scores, respectively, each normalized to $[0,1]$. Ordinary character-, sentence-, and paragraph-level translation uses
\begin{equation}
    q_{di}=\begin{cases}
        \dfrac{b_{di}+m_{di}}{2},
            & d=1,\\[5pt]
        \dfrac{r_{di}+m_{di}}{2},
            & d=2.
    \end{cases}
    \label{eq:translation_general}
\end{equation}
Document-level translation uses
\begin{equation}
    q_{di}=z_{di}.
    \label{eq:translation_document}
\end{equation}
Category-specific checks replace these overlap scores where the instance tests a particular patent requirement. Let $y_i$ be the translated output and $T_i$ its required target terms. Terminology accuracy is
\begin{equation}
    q_{di}^{\mathrm{term}}=
    \prod_{t\in T_i}\mathbf{1}[t\text{ occurs in }y_i].
\end{equation}
Terminology consistency additionally checks the required occurrence count $c_{it}$ for each annotated term:
\begin{equation}
    q_{di}^{\mathrm{cons}}=
    \prod_{t\in T_i}\mathbf{1}[\mathrm{count}(t,y_i)=c_{it}].
\end{equation}
For the required special-character set $H_i$, preservation is
\begin{equation}
    q_{di}^{\mathrm{char}}=
    \prod_{h\in H_i}\mathbf{1}[h\text{ occurs in }y_i].
\end{equation}
The English-title convention check assigns
\begin{equation}
    q_{di}^{\mathrm{title}}=
    \mathbf{1}[\mathrm{first}(y_i)\notin\{\text{a},\text{an},\text{the}\}],
\end{equation}
with the initial word checked case-insensitively. For a required set of patent-writing expressions $W_i$, the score is
\begin{equation}
    q_{di}^{\mathrm{norm}}=
    \frac{\sum_{w\in W_i}\mathbf{1}[w\text{ occurs in }y_i]}{|W_i|}.
\end{equation}
Each instance uses the rule corresponding to its annotated category. Hallucination and omission checks are auxiliary diagnostics and do not subtract additional points from $q_{di}$.

Let $D$ be the number of translation directions, with $D=2$, and let $N_d$ be the number of instances assigned to direction $d$. The directional score $S_d$ and the final translation score $S_{\tau}$ are
\begin{align}
    S_d&=\frac{c}{N_d}\sum_{i=1}^{N_d}q_{di},\\
    S_{\tau}&=\frac{1}{D}\sum_{d=1}^{D}S_d.
    \label{eq:translation_bidirectional}
\end{align}
Execution failures remain in the directional denominators with zero contribution. The two directional scores are averaged only after their respective instance-level aggregation.

\clearpage
\onecolumn

\clearpage
\section{Task Completion Rates}
\label{app:execution_success}
\begingroup\small
We measure task completion of whether an agent delivers the requested result under the evaluation setup. A run is successful if it produces the requested task output, regardless of its quality score Success rate is the proportion of recorded tasks completed after the allowed retries. Table~\ref{tab:execution_success} reports this rate for each task, model, and harness.
\begin{center}\captionsetup{hypcap=false}
\captionof{table}{Task completion rates for each model--harness configuration.}
\label{tab:execution_success}
\small\setlength{\tabcolsep}{3pt}\renewcommand{\arraystretch}{0.92}
\resizebox{\linewidth}{!}{\begin{tabular}{@{}lcccccccc@{}}
\toprule
Model & \makecell[c]{Novelty\\search} & \makecell[c]{FTO} & \makecell[c]{Design\\FTO} & \makecell[c]{Invalidity\\search} & \makecell[c]{Claim\\charting} & \makecell[c]{Drafting} & \makecell[c]{OAR} & \makecell[c]{Translation} \\
\midrule
\multicolumn{9}{l}{\textit{Codex}} \\
\midrule
GPT-5.6-Sol & $100.0$ & $100.0$ & $100.0$ & $100.0$ & $100.0$ & $100.0$ & $100.0$ & $100.0$ \\
Qwen3.8-Max & $100.0$ & $100.0$ & $82.5$ & $98.8$ & $96.7$ & $40.0$ & $45.0$ & $100.0$ \\
Kimi K3 & $100.0$ & $100.0$ & $98.8$ & $100.0$ & $90.0$ & $83.3$ & $17.5$ & $100.0$ \\
DeepSeek-V4-Pro & $26.3$ & $53.3$ & $7.5$ & $57.5$ & $16.7$ & $73.3$ & $30.0$ & $100.0$ \\
MiniMax-M3 & $65.0$ & $96.7$ & $12.5$ & $3.8$ & $66.7$ & $25.0$ & $0.0$ & $100.0$ \\
GLM-5.3 & $100.0$ & $100.0$ & $7.5$ & $100.0$ & $43.3$ & $93.3$ & $75.0$ & $100.0$ \\
\midrule
\multicolumn{9}{l}{\textit{Claude Code}} \\
\midrule
GPT-5.6-Sol & $100.0$ & $100.0$ & $100.0$ & $100.0$ & $96.7$ & $100.0$ & $100.0$ & $100.0$ \\
Claude Opus 5 & $100.0$ & $100.0$ & $100.0$ & $100.0$ & $93.3$ & $100.0$ & $97.5$ & $100.0$ \\
Qwen3.8-Max & $100.0$ & $100.0$ & $100.0$ & $100.0$ & $100.0$ & $100.0$ & $97.5$ & $100.0$ \\
Kimi K3 & $100.0$ & $100.0$ & $100.0$ & $100.0$ & $46.7$ & $100.0$ & $100.0$ & $100.0$ \\
DeepSeek-V4-Pro & $100.0$ & $100.0$ & $0.0$ & $100.0$ & $83.3$ & $55.0$ & $100.0$ & $100.0$ \\
MiniMax-M3 & $100.0$ & $100.0$ & $100.0$ & $100.0$ & $50.0$ & $100.0$ & $82.5$ & $100.0$ \\
GLM-5.3 & $100.0$ & $100.0$ & $0.0$ & $100.0$ & $6.7$ & $100.0$ & $0.0$ & $100.0$ \\
\midrule
\multicolumn{9}{l}{\textit{Hermes}} \\
\midrule
GPT-5.6-Sol & $100.0$ & $100.0$ & $97.5$ & $100.0$ & $16.7$ & $66.7$ & $100.0$ & $100.0$ \\
Claude Opus 5 & $100.0$ & $100.0$ & $97.5$ & $100.0$ & $96.7$ & $100.0$ & $90.0$ & $99.9$ \\
Qwen3.8-Max & $42.5$ & $96.7$ & $61.3$ & $48.8$ & $0.0$ & $51.7$ & $5.0$ & $100.0$ \\
Kimi K3 & $100.0$ & $100.0$ & $100.0$ & $100.0$ & $6.7$ & $90.0$ & $85.0$ & $100.0$ \\
DeepSeek-V4-Pro & $68.8$ & $76.7$ & $95.0$ & $100.0$ & $13.3$ & $86.7$ & $20.0$ & $100.0$ \\
MiniMax-M3 & $91.3$ & $96.7$ & $90.0$ & $91.3$ & $20.0$ & $81.7$ & $67.5$ & $100.0$ \\
GLM-5.3 & $100.0$ & $100.0$ & $0.0$ & $98.8$ & $0.0$ & $100.0$ & $62.5$ & $100.0$ \\
\midrule
\multicolumn{9}{l}{\textit{OpenClaw}} \\
\midrule
GPT-5.6-Sol & $100.0$ & $98.3$ & $100.0$ & $100.0$ & $100.0$ & $96.7$ & $80.0$ & $100.0$ \\
Claude Opus 5 & $22.5$ & $93.3$ & $100.0$ & $31.3$ & $23.3$ & $88.3$ & $5.0$ & $100.0$ \\
Qwen3.8-Max & $100.0$ & $100.0$ & $98.8$ & $93.8$ & $6.7$ & $96.7$ & $12.5$ & $100.0$ \\
Kimi K3 & $100.0$ & $100.0$ & $100.0$ & $93.8$ & $6.7$ & $71.7$ & $45.0$ & $100.0$ \\
DeepSeek-V4-Pro & $100.0$ & $100.0$ & $76.3$ & $97.5$ & $0.0$ & $66.7$ & $87.5$ & $100.0$ \\
MiniMax-M3 & $82.5$ & $100.0$ & $11.3$ & $85.0$ & $0.0$ & $86.7$ & $62.5$ & $100.0$ \\
GLM-5.3 & $97.5$ & $96.7$ & $42.5$ & $83.8$ & $0.0$ & $1.7$ & $5.0$ & $99.9$ \\
\bottomrule\end{tabular}}
\end{center}
Claim charting exposes a failure to reserve enough output budget for the final deliverable. In our runs, some model--harness combinations continue reasoning until they exhaust the maximum output length, leaving no complete claim chart. Repeated attempts do not consistently resolve this behavior. For example, DeepSeek--OpenClaw produces no evaluable chart across 30 cases, and every stored response reports truncation at the model's output-token limit. The failure occurs before the substantive quality of the chart can be assessed: the agent spends its budget on reasoning but does not deliver the requested result.

Office-action response exposes a different failure in document handling. Scanned examination documents require agents to extract their contents before drafting. OCR alone does not prevent failure if later tools return image inputs that the model endpoint cannot accept. The case in Appendix~\ref{app:execution_failure_cases} shows an OAR run ending when image crops reach a text-only endpoint. Reliable completion therefore requires both control over intermediate reasoning and compatibility between tool outputs and model inputs.
\endgroup
\clearpage

\section{Detailed Task Results}
\label{app:detailed_results}
Scores use a 0--100 scale with one decimal place. The untested Codex--Opus configuration is omitted.

\subsection{Novelty Search}

\begingroup
\small
\setlength{\tabcolsep}{5pt}
\renewcommand{\arraystretch}{1.08}
\centering
\captionsetup{hypcap=false}
\captionof{table}{Novelty-search results over 80 cases. Hit measures whether a case retrieves at least one annotated patent family; recall measures the fraction of annotated families recovered, averaged over cases.}
\label{tab:app_novelty}
\sbox0{\begin{tabular}{@{}lcccc@{}}
\toprule
Model & Hit@10 & Recall@10 & Hit@100 & Recall@100 \\
\midrule
\multicolumn{5}{l}{\textit{Codex}} \\
\midrule
GPT-5.6-Sol & 27.5 & 9.4 & 27.5 & 9.4 \\
Qwen3.8-Max & 35.0 & 10.0 & 51.3 & 21.6 \\
Kimi K3 & 12.5 & 5.1 & 13.8 & 5.7 \\
DeepSeek-V4-Pro & 7.5 & 2.7 & 10.0 & 3.4 \\
MiniMax-M3 & 15.0 & 5.2 & 15.0 & 5.6 \\
GLM-5.3 & 76.3 & 33.6 & 86.3 & 42.3 \\
\midrule
\multicolumn{5}{l}{\textit{Claude Code}} \\
\midrule
GPT-5.6-Sol & 38.8 & 11.6 & 38.8 & 11.8 \\
Claude Opus 5 & 37.5 & 11.3 & 52.5 & 19.4 \\
Qwen3.8-Max & 17.5 & 4.5 & 32.5 & 9.6 \\
Kimi K3 & 31.3 & 8.9 & 33.8 & 9.5 \\
DeepSeek-V4-Pro & 35.0 & 10.0 & 37.5 & 10.6 \\
MiniMax-M3 & 23.8 & 6.1 & 27.5 & 7.7 \\
GLM-5.3 & 41.3 & 13.6 & 50.0 & 19.6 \\
\midrule
\multicolumn{5}{l}{\textit{Hermes}} \\
\midrule
GPT-5.6-Sol & 57.5 & 26.1 & 62.5 & 31.0 \\
Claude Opus 5 & 45.0 & 14.4 & 52.5 & 21.6 \\
Qwen3.8-Max & 12.5 & 3.4 & 17.5 & 5.2 \\
Kimi K3 & 36.3 & 9.6 & 36.3 & 9.6 \\
DeepSeek-V4-Pro & 20.0 & 5.4 & 22.5 & 6.2 \\
MiniMax-M3 & 8.8 & 2.3 & 11.3 & 2.7 \\
GLM-5.3 & 36.3 & 8.7 & 43.8 & 12.1 \\
\midrule
\multicolumn{5}{l}{\textit{OpenClaw}} \\
\midrule
GPT-5.6-Sol & 46.3 & 13.5 & 46.3 & 13.5 \\
Claude Opus 5 & 5.0 & 1.1 & 5.0 & 1.1 \\
Qwen3.8-Max & 33.8 & 9.7 & 36.3 & 10.7 \\
Kimi K3 & 37.5 & 10.9 & 40.0 & 11.6 \\
DeepSeek-V4-Pro & 22.5 & 6.0 & 23.8 & 6.2 \\
MiniMax-M3 & 23.8 & 6.0 & 28.8 & 7.1 \\
GLM-5.3 & 52.5 & 18.0 & 57.5 & 21.9 \\
\bottomrule
\end{tabular}}
\ifdim\wd0>\linewidth\resizebox{\linewidth}{!}{\usebox0}\else\usebox0\fi
\par\endgroup
\par\vspace{8pt}\noindent

\noindent Novelty search recovers at least one relevant patent family far more often than it recovers the complete reference set. GLM-5.3 with Codex leads every reported metric, but its 86.3 Hit@100 is accompanied by only 42.3 Recall@100. Expanding the cutoff from ten to one hundred raises its recall from 33.6 to 42.3, so some relevant families occur deeper in the results, while most remain unrecovered on average. The preferred harness also depends on the model: GLM scores higher with Codex than with Hermes, whereas GPT-5.6-Sol shows the reverse ordering. Strong case-level hit rates therefore coexist with incomplete evidence coverage, even for the leading configuration.

\clearpage\subsection{Patent FTO}

\begingroup
\small
\setlength{\tabcolsep}{5pt}
\renewcommand{\arraystretch}{1.08}
\centering
\captionsetup{hypcap=false}
\captionof{table}{Patent FTO results over 60 cases. High-risk recall retains candidates labeled high risk. References are matched at the application level. Macro recall weights cases equally; micro recall pools reference applications.}
\label{tab:app_fto}
\sbox0{\begin{tabular}{@{}lccccc@{}}
\toprule
Model & Hit & \makecell{Recall\\macro} & \makecell{Recall\\micro} & \makecell{High-risk\\macro} & \makecell{High-risk\\micro} \\
\midrule
\multicolumn{6}{l}{\textit{Codex}} \\
\midrule
GPT-5.6-Sol & 60.0 & 54.2 & 51.1 & 45.8 & 39.8 \\
Qwen3.8-Max & 56.7 & 55.5 & 48.9 & 55.2 & 47.7 \\
Kimi K3 & 6.7 & 5.7 & 5.7 & 5.3 & 4.6 \\
DeepSeek-V4-Pro & 45.0 & 43.2 & 44.3 & 39.5 & 40.9 \\
MiniMax-M3 & 8.3 & 8.3 & 6.8 & 8.3 & 6.8 \\
GLM-5.3 & 76.7 & 74.0 & 62.5 & 69.8 & 59.1 \\
\midrule
\multicolumn{6}{l}{\textit{Claude Code}} \\
\midrule
GPT-5.6-Sol & 86.7 & 79.4 & 67.1 & 74.2 & 61.4 \\
Claude Opus 5 & 90.0 & 86.5 & 77.3 & 85.3 & 73.9 \\
Qwen3.8-Max & 88.3 & 86.2 & 76.1 & 83.1 & 70.5 \\
Kimi K3 & 86.7 & 83.2 & 75.0 & 81.6 & 70.5 \\
DeepSeek-V4-Pro & 80.0 & 78.2 & 70.5 & 76.8 & 63.6 \\
MiniMax-M3 & 75.0 & 72.5 & 65.9 & 68.4 & 62.5 \\
GLM-5.3 & 86.7 & 83.9 & 72.7 & 82.6 & 69.3 \\
\midrule
\multicolumn{6}{l}{\textit{Hermes}} \\
\midrule
GPT-5.6-Sol & 65.0 & 62.7 & 52.3 & 56.7 & 45.5 \\
Claude Opus 5 & 90.0 & 86.4 & 77.3 & 85.2 & 73.9 \\
Qwen3.8-Max & 83.3 & 80.5 & 75.0 & 77.1 & 68.2 \\
Kimi K3 & 86.7 & 82.7 & 75.0 & 79.0 & 67.1 \\
DeepSeek-V4-Pro & 63.3 & 61.2 & 56.8 & 57.5 & 52.3 \\
MiniMax-M3 & 40.0 & 40.0 & 33.0 & 36.7 & 30.7 \\
GLM-5.3 & 90.0 & 86.2 & 76.1 & 81.4 & 65.9 \\
\midrule
\multicolumn{6}{l}{\textit{OpenClaw}} \\
\midrule
GPT-5.6-Sol & 86.7 & 82.2 & 72.7 & 55.3 & 50.0 \\
Claude Opus 5 & 80.0 & 76.8 & 65.9 & 75.1 & 64.8 \\
Qwen3.8-Max & 86.7 & 84.2 & 73.9 & 83.2 & 71.6 \\
Kimi K3 & 86.7 & 83.0 & 70.5 & 81.9 & 68.2 \\
DeepSeek-V4-Pro & 73.3 & 69.9 & 56.8 & 69.9 & 56.8 \\
MiniMax-M3 & 80.0 & 77.5 & 69.3 & 74.8 & 65.9 \\
GLM-5.3 & 86.7 & 82.5 & 72.7 & 81.5 & 70.5 \\
\bottomrule
\end{tabular}}
\ifdim\wd0>\linewidth\resizebox{\linewidth}{!}{\usebox0}\else\usebox0\fi
\par\endgroup
\par\vspace{8pt}\noindent

\noindent Risk assignment determines whether retrieved FTO references remain in the high-risk shortlist. With GPT-5.6-Sol fixed, OpenClaw has higher macro recall than Claude Code, at 82.2 versus 79.4, but lower high-risk macro recall, at 55.3 versus 74.2. The retrieval advantage reverses after risk filtering because more annotated references are excluded from OpenClaw's high-risk subset. Opus with Claude Code preserves substantially more of its retrieved evidence: macro recall changes only from 86.5 to 85.3 after filtering. These results establish two separate requirements for FTO: retrieving the relevant applications and assigning them a risk level that keeps them in the actionable shortlist.

\clearpage\subsection{Design FTO}

\begingroup
\small
\setlength{\tabcolsep}{5pt}
\renewcommand{\arraystretch}{1.08}
\centering
\captionsetup{hypcap=false}
\captionof{table}{Design FTO results over 80 cases. PRES@200 is the source evaluator's rank-sensitive retrieval score at cutoff 200; all three metrics use the assigned case set.}
\label{tab:app_design_fto}
\sbox0{\begin{tabular}{@{}lccc@{}}
\toprule
Model & Hit@200 & Recall@200 & PRES@200 \\
\midrule
\multicolumn{4}{l}{\textit{Codex}} \\
\midrule
GPT-5.6-Sol & 1.3 & 0.4 & 0.4 \\
Qwen3.8-Max & 0.0 & 0.0 & 0.0 \\
Kimi K3 & 0.0 & 0.0 & 0.0 \\
DeepSeek-V4-Pro & 0.0 & 0.0 & 0.0 \\
MiniMax-M3 & 0.0 & 0.0 & 0.0 \\
GLM-5.3 & 0.0 & 0.0 & 0.0 \\
\midrule
\multicolumn{4}{l}{\textit{Claude Code}} \\
\midrule
GPT-5.6-Sol & 0.0 & 0.0 & 0.0 \\
Claude Opus 5 & 5.0 & 2.7 & 2.5 \\
Qwen3.8-Max & 3.8 & 3.8 & 3.7 \\
Kimi K3 & 0.0 & 0.0 & 0.0 \\
DeepSeek-V4-Pro & 0.0 & 0.0 & 0.0 \\
MiniMax-M3 & 0.0 & 0.0 & 0.0 \\
GLM-5.3 & 0.0 & 0.0 & 0.0 \\
\midrule
\multicolumn{4}{l}{\textit{Hermes}} \\
\midrule
GPT-5.6-Sol & 0.0 & 0.0 & 0.0 \\
Claude Opus 5 & 2.5 & 1.9 & 1.8 \\
Qwen3.8-Max & 0.0 & 0.0 & 0.0 \\
Kimi K3 & 0.0 & 0.0 & 0.0 \\
DeepSeek-V4-Pro & 0.0 & 0.0 & 0.0 \\
MiniMax-M3 & 0.0 & 0.0 & 0.0 \\
GLM-5.3 & 0.0 & 0.0 & 0.0 \\
\midrule
\multicolumn{4}{l}{\textit{OpenClaw}} \\
\midrule
GPT-5.6-Sol & 0.0 & 0.0 & 0.0 \\
Claude Opus 5 & 1.3 & 1.3 & 1.2 \\
Qwen3.8-Max & 1.3 & 0.4 & 0.4 \\
Kimi K3 & 1.3 & 0.4 & 0.4 \\
DeepSeek-V4-Pro & 0.0 & 0.0 & 0.0 \\
MiniMax-M3 & 0.0 & 0.0 & 0.0 \\
GLM-5.3 & 0.0 & 0.0 & 0.0 \\
\bottomrule
\end{tabular}}
\ifdim\wd0>\linewidth\resizebox{\linewidth}{!}{\usebox0}\else\usebox0\fi
\par\endgroup
\par\vspace{8pt}\noindent

\noindent Reference coverage is uniformly low in design FTO, including for the best model--harness pairing. Qwen3.8-Max with Claude Code leads Recall@200 and PRES@200 at 3.8 and 3.7, while the highest Hit@200 is only 5.0, achieved by Opus with Claude Code. Thus, even the highest hit rate leaves nearly all cases without an annotated design in the returned results. Qwen's recall falls to 0.0 with Codex and Hermes and to 0.4 with OpenClaw, so changing the harness does not remove this shared shortfall. The central problem is recovering the reference designs; rearranging the retrieved candidates cannot compensate for references absent from the list.

\clearpage\subsection{Invalidity Search}

\begingroup
\small
\setlength{\tabcolsep}{5pt}
\renewcommand{\arraystretch}{1.08}
\centering
\captionsetup{hypcap=false}
\captionof{table}{Invalidity-search results over the assigned 80 cases. Core denotes the source evaluator's designated core references; All denotes the full annotated reference set. The core and full reference sets can have different eligible case counts, so their scores are reported separately rather than subtracted.}
\label{tab:app_invalidity}
\sbox0{\begin{tabular}{@{}lcccccccc@{}}
\toprule
Model & \makecell{Core hit\\@5} & \makecell{Core recall\\@5} & \makecell{All hit\\@5} & \makecell{All recall\\@5} & \makecell{Core hit\\@All} & \makecell{Core recall\\@All} & \makecell{All hit\\@All} & \makecell{All recall\\@All} \\
\midrule
\multicolumn{9}{l}{\textit{Codex}} \\
\midrule
GPT-5.6-Sol & 26.9 & 23.5 & 32.5 & 23.3 & 29.5 & 26.7 & 33.8 & 26.4 \\
Qwen3.8-Max & 30.8 & 23.7 & 42.5 & 22.6 & 41.0 & 32.5 & 53.8 & 31.2 \\
Kimi K3 & 5.1 & 3.0 & 7.5 & 3.0 & 6.4 & 4.3 & 8.8 & 4.6 \\
DeepSeek-V4-Pro & 14.1 & 10.0 & 18.8 & 9.5 & 18.0 & 12.4 & 21.3 & 10.3 \\
MiniMax-M3 & 1.3 & 1.3 & 1.3 & 0.8 & 1.3 & 1.3 & 1.3 & 0.8 \\
GLM-5.3 & 29.5 & 22.8 & 43.8 & 26.2 & 39.7 & 31.1 & 51.3 & 33.8 \\
\midrule
\multicolumn{9}{l}{\textit{Claude Code}} \\
\midrule
GPT-5.6-Sol & 30.8 & 23.3 & 36.3 & 21.4 & 39.7 & 31.0 & 48.8 & 31.8 \\
Claude Opus 5 & 28.2 & 22.2 & 38.8 & 20.6 & 34.6 & 27.1 & 48.8 & 28.7 \\
Qwen3.8-Max & 14.1 & 11.3 & 28.8 & 15.2 & 21.8 & 18.2 & 35.0 & 21.0 \\
Kimi K3 & 24.4 & 19.0 & 37.5 & 19.1 & 30.8 & 25.4 & 45.0 & 25.8 \\
DeepSeek-V4-Pro & 19.2 & 13.7 & 31.3 & 16.7 & 26.9 & 21.8 & 42.5 & 24.9 \\
MiniMax-M3 & 10.3 & 6.8 & 17.5 & 8.3 & 14.1 & 10.5 & 22.5 & 12.4 \\
GLM-5.3 & 38.5 & 30.3 & 52.5 & 28.3 & 42.3 & 33.6 & 57.5 & 32.4 \\
\midrule
\multicolumn{9}{l}{\textit{Hermes}} \\
\midrule
GPT-5.6-Sol & 9.0 & 9.0 & 11.3 & 8.4 & 10.3 & 9.6 & 13.8 & 10.3 \\
Claude Opus 5 & 56.4 & 53.9 & 61.3 & 42.2 & 60.3 & 58.7 & 67.5 & 55.0 \\
Qwen3.8-Max & 20.5 & 20.5 & 25.0 & 17.0 & 20.5 & 20.5 & 27.5 & 20.5 \\
Kimi K3 & 42.3 & 36.8 & 50.0 & 28.6 & 46.2 & 40.6 & 55.0 & 34.5 \\
DeepSeek-V4-Pro & 33.3 & 27.1 & 38.8 & 23.7 & 39.7 & 33.3 & 45.0 & 29.9 \\
MiniMax-M3 & 6.4 & 5.8 & 8.8 & 4.8 & 6.4 & 5.8 & 11.3 & 6.0 \\
GLM-5.3 & 41.0 & 34.6 & 51.3 & 32.6 & 43.6 & 37.6 & 57.5 & 40.1 \\
\midrule
\multicolumn{9}{l}{\textit{OpenClaw}} \\
\midrule
GPT-5.6-Sol & 24.4 & 18.2 & 35.0 & 18.1 & 30.8 & 24.4 & 41.3 & 24.8 \\
Claude Opus 5 & 9.0 & 7.7 & 10.0 & 6.2 & 10.3 & 9.0 & 12.5 & 7.7 \\
Qwen3.8-Max & 20.5 & 15.2 & 28.8 & 14.7 & 23.1 & 17.7 & 31.3 & 17.3 \\
Kimi K3 & 19.2 & 15.0 & 25.0 & 14.3 & 23.1 & 18.2 & 31.3 & 17.7 \\
DeepSeek-V4-Pro & 7.7 & 6.4 & 8.8 & 6.5 & 7.7 & 6.4 & 8.8 & 6.5 \\
MiniMax-M3 & 9.0 & 7.7 & 12.5 & 7.6 & 9.0 & 9.0 & 16.3 & 10.5 \\
GLM-5.3 & 21.8 & 17.1 & 33.8 & 17.0 & 24.4 & 20.0 & 38.8 & 21.3 \\
\bottomrule
\end{tabular}}
\ifdim\wd0>\linewidth\resizebox{\linewidth}{!}{\usebox0}\else\usebox0\fi
\par\endgroup
\par\vspace{8pt}\noindent

\noindent Opus with Hermes has a consistent invalidity-search advantage across both reference groups and both cutoffs, but its reference coverage remains incomplete. It leads all eight metrics, with full-reference recall increasing from 42.2 in the top five results to 55.0 over the complete list. The additional results improve coverage, yet 45.0\% of annotated references remain unrecovered on average. This harness advantage is specific to the model: Opus reaches only 28.7 full-list recall with Claude Code, whereas GPT-5.6-Sol performs better with Claude Code than with Hermes, at 31.8 versus 10.3. The table therefore establishes a strong Opus--Hermes pairing, rather than a general retrieval advantage for Hermes.

\clearpage\subsection{Claim Charting}

\begingroup
\small
\setlength{\tabcolsep}{5pt}
\renewcommand{\arraystretch}{1.08}
\centering
\captionsetup{hypcap=false}
\captionof{table}{Claim-charting results over all 30 assigned cases. SP-F1 measures feature extraction, CP-F1 measures feature-level evidence comparison, and PB-AC is the source evaluator's patentability-judgment accuracy. All columns use full-set scores rather than scores conditional on completed outputs.}
\label{tab:app_charting}
\sbox0{\begin{tabular}{@{}lccc@{}}
\toprule
Model & SP-F1 & CP-F1 & PB-AC \\
\midrule
\multicolumn{4}{l}{\textit{Codex}} \\
\midrule
GPT-5.6-Sol & 73.7 & 53.3 & 72.9 \\
Qwen3.8-Max & 67.6 & 54.9 & 73.1 \\
Kimi K3 & 62.3 & 48.2 & 69.5 \\
DeepSeek-V4-Pro & 12.4 & 9.3 & 13.5 \\
MiniMax-M3 & 40.1 & 30.3 & 48.3 \\
GLM-5.3 & 30.0 & 24.8 & 35.9 \\
\midrule
\multicolumn{4}{l}{\textit{Claude Code}} \\
\midrule
GPT-5.6-Sol & 70.6 & 54.1 & 72.5 \\
Claude Opus 5 & 71.3 & 54.0 & 72.6 \\
Qwen3.8-Max & 51.7 & 50.7 & 72.7 \\
Kimi K3 & 39.5 & 31.8 & 37.7 \\
DeepSeek-V4-Pro & 54.0 & 40.8 & 63.0 \\
MiniMax-M3 & 25.7 & 21.1 & 34.9 \\
GLM-5.3 & 5.0 & 5.5 & 6.0 \\
\midrule
\multicolumn{4}{l}{\textit{Hermes}} \\
\midrule
GPT-5.6-Sol & 14.8 & 12.7 & 15.3 \\
Claude Opus 5 & 73.9 & 55.1 & 74.4 \\
Qwen3.8-Max & 0.0 & 0.0 & 0.0 \\
Kimi K3 & 4.7 & 3.2 & 5.9 \\
DeepSeek-V4-Pro & 9.5 & 8.3 & 11.2 \\
MiniMax-M3 & 13.9 & 8.4 & 12.9 \\
GLM-5.3 & 0.0 & 0.0 & 0.0 \\
\midrule
\multicolumn{4}{l}{\textit{OpenClaw}} \\
\midrule
GPT-5.6-Sol & 72.9 & 54.7 & 74.1 \\
Claude Opus 5 & 20.0 & 16.3 & 20.1 \\
Qwen3.8-Max & 5.1 & 3.7 & 5.9 \\
Kimi K3 & 5.0 & 5.8 & 6.0 \\
DeepSeek-V4-Pro & 0.0 & 0.0 & 0.0 \\
MiniMax-M3 & 0.0 & 0.0 & 0.0 \\
GLM-5.3 & 0.0 & 0.0 & 0.0 \\
\bottomrule
\end{tabular}}
\ifdim\wd0>\linewidth\resizebox{\linewidth}{!}{\usebox0}\else\usebox0\fi
\par\endgroup
\par\vspace{8pt}\noindent

\noindent Accurate feature extraction does not ensure accurate feature-to-evidence alignment. Opus with Hermes leads all three claim-charting metrics, but its CP-F1 reaches only 55.1 compared with an SP-F1 of 73.9. The distinction also changes model rankings within Codex: GPT-5.6-Sol extracts features more accurately than Qwen3.8-Max, at 73.7 versus 67.6, while Qwen achieves higher CP-F1, at 54.9 versus 53.3. The model with the stronger extracted feature set is therefore not the model with the stronger evidence comparison. Evaluation must retain both stages, because an extraction score alone does not establish whether the resulting chart correctly maps those features to the supplied documents.

\clearpage\subsection{Patent Drafting}

\begingroup
\small
\setlength{\tabcolsep}{5pt}
\renewcommand{\arraystretch}{1.08}
\centering
\captionsetup{hypcap=false}
\captionof{table}{Drafting quality and pass rate. The expert-defined passing threshold is 80/100. Quality is the mean over successfully delivered drafts. Pass rate is $100\,|\{i:q_i\geq80\}|/60$; an unscored case does not pass. The expert-defined threshold identifies practically valuable, usable drafts. We do not report the full-set task total.}
\label{tab:app_drafting}
\sbox0{\begin{tabular}{@{}lcc@{}}
\toprule
Model & Quality & Pass rate (\%) \\
\midrule
\multicolumn{3}{l}{\textit{Codex}} \\
\midrule
GPT-5.6-Sol & 84.7 & 76.7 \\
Qwen3.8-Max & 79.2 & 25.0 \\
Kimi K3 & 81.2 & 46.7 \\
DeepSeek-V4-Pro & 79.4 & 35.0 \\
MiniMax-M3 & 71.6 & 5.0 \\
GLM-5.3 & 79.9 & 51.7 \\
\midrule
\multicolumn{3}{l}{\textit{Claude Code}} \\
\midrule
GPT-5.6-Sol & 84.0 & 65.0 \\
Claude Opus 5 & 82.8 & 66.7 \\
Qwen3.8-Max & 80.8 & 63.3 \\
Kimi K3 & 82.9 & 65.0 \\
DeepSeek-V4-Pro & 78.1 & 25.0 \\
MiniMax-M3 & 72.4 & 28.3 \\
GLM-5.3 & 82.3 & 66.7 \\
\midrule
\multicolumn{3}{l}{\textit{Hermes}} \\
\midrule
GPT-5.6-Sol & 85.0 & 53.3 \\
Claude Opus 5 & 81.0 & 53.3 \\
Qwen3.8-Max & 82.7 & 30.0 \\
Kimi K3 & 81.3 & 50.0 \\
DeepSeek-V4-Pro & 77.6 & 35.0 \\
MiniMax-M3 & 71.2 & 16.7 \\
GLM-5.3 & 77.4 & 40.0 \\
\midrule
\multicolumn{3}{l}{\textit{OpenClaw}} \\
\midrule
GPT-5.6-Sol & 83.3 & 66.7 \\
Claude Opus 5 & 81.2 & 55.0 \\
Qwen3.8-Max & 81.6 & 53.3 \\
Kimi K3 & 82.1 & 41.7 \\
DeepSeek-V4-Pro & 79.9 & 35.0 \\
MiniMax-M3 & 72.8 & 23.3 \\
GLM-5.3 & 83.6 & 1.7 \\
\bottomrule
\end{tabular}}
\ifdim\wd0>\linewidth\resizebox{\linewidth}{!}{\usebox0}\else\usebox0\fi
\par\endgroup
\par\vspace{8pt}\noindent

\noindent GPT-5.6-Sol with Codex produces the highest proportion of drafts meeting the expert-defined usability standard, despite GPT with Hermes having the highest mean quality. Experts set the passing threshold at 80/100 to identify drafts with practical value that meet the benchmark's requirements for a usable deliverable. GPT reaches a 76.7\% pass rate with Codex, compared with 53.3\% with Hermes, although their quality means are close at 84.7 and 85.0. The mean describes scored documents, while pass rate measures usable deliverables over all assigned cases. For this task, selecting by mean quality alone chooses a different configuration from selecting by the frequency of delivering a usable draft.

\clearpage\subsection{Office-Action Response}

\begingroup\small
\setlength{\tabcolsep}{5pt}\renewcommand{\arraystretch}{1.08}
\centering\captionsetup{hypcap=false}
\captionof{table}{OAR quality and pass rate: quality averages only successfully scored responses, whereas pass rate uses all 40 assigned cases as its denominator, with unscored cases counted as not passing. The expert-defined threshold of 80/100 identifies practically valuable, usable responses.}
\label{tab:app_oar_quality}
\sbox0{\begin{tabular}{@{}lcc@{}}
\toprule
Model & Quality & Pass rate (\%) \\
\midrule
\multicolumn{3}{l}{\textit{Codex}} \\
\midrule
GPT-5.6-Sol & 34.8 & 0.0 \\
Qwen3.8-Max & 27.6 & 0.0 \\
Kimi K3 & 36.9 & 0.0 \\
DeepSeek-V4-Pro & 21.2 & 0.0 \\
MiniMax-M3 & 0.0 & 0.0 \\
GLM-5.3 & 28.0 & 0.0 \\
\midrule
\multicolumn{3}{l}{\textit{Claude Code}} \\
\midrule
GPT-5.6-Sol & 38.0 & 2.5 \\
Claude Opus 5 & 25.7 & 0.0 \\
Qwen3.8-Max & 17.3 & 0.0 \\
Kimi K3 & 12.2 & 0.0 \\
DeepSeek-V4-Pro & 22.3 & 0.0 \\
MiniMax-M3 & 0.6 & 0.0 \\
GLM-5.3 & 0.0 & 0.0 \\
\midrule
\multicolumn{3}{l}{\textit{Hermes}} \\
\midrule
GPT-5.6-Sol & 54.8 & 22.5 \\
Claude Opus 5 & 22.7 & 0.0 \\
Qwen3.8-Max & 7.3 & 0.0 \\
Kimi K3 & 24.0 & 2.5 \\
DeepSeek-V4-Pro & 23.4 & 0.0 \\
MiniMax-M3 & 15.1 & 0.0 \\
GLM-5.3 & 27.2 & 0.0 \\
\midrule
\multicolumn{3}{l}{\textit{OpenClaw}} \\
\midrule
GPT-5.6-Sol & 53.3 & 15.0 \\
Claude Opus 5 & 22.2 & 0.0 \\
Qwen3.8-Max & 23.3 & 0.0 \\
Kimi K3 & 30.3 & 2.5 \\
DeepSeek-V4-Pro & 28.1 & 0.0 \\
MiniMax-M3 & 24.5 & 5.0 \\
GLM-5.3 & 29.5 & 0.0 \\
\bottomrule\end{tabular}}
\ifdim\wd0>\linewidth\resizebox{\linewidth}{!}{\usebox0}\else\usebox0\fi
\par\endgroup\par\vspace{8pt}
\noindent Usable OAR responses remain uncommon across configurations. GPT-5.6-Sol with Hermes leads mean quality at 54.8 and passes nine of forty cases, giving a pass rate of 22.5\%. The same model with OpenClaw passes six cases, while Claude Code passes one. MiniMax with OpenClaw passes two cases, and Kimi passes one case each with Hermes and OpenClaw. Every other configuration has zero passing cases. Even the leading configuration therefore fails to meet the expert-defined usability standard on most assigned cases; high-quality responses occur in individual cases without establishing consistent delivery.

\clearpage

\begingroup\small
\setlength{\tabcolsep}{5pt}\renewcommand{\arraystretch}{1.08}
\centering\captionsetup{hypcap=false}
\captionof{table}{OAR basic-quality criteria. G1: case understanding; G2: response strategy; G7: logical consistency; G8: response completeness; G9: deliverable completeness and text quality; G10: jurisdictional compliance; G11: factual and citation accuracy. Each criterion is averaged over its applicable scored cases on a 0--100 scale. Missing outputs and failed grading are excluded.}
\label{tab:app_oar_foundation}
\sbox0{\begin{tabular}{@{}lccccccc@{}}
\toprule
Model & G1 & G2 & G7 & G8 & G9 & G10 & G11 \\
\midrule
\multicolumn{8}{l}{\textit{Codex}} \\
\midrule
GPT-5.6-Sol & 23.8 & 17.5 & 91.3 & 97.5 & 71.3 & 87.5 & 15.0 \\
Qwen3.8-Max & 0.0 & 33.3 & 19.4 & 97.2 & 8.3 & 16.7 & 0.0 \\
Kimi K3 & 21.4 & 50.0 & 42.9 & 78.6 & 14.3 & 50.0 & 14.3 \\
DeepSeek-V4-Pro & 12.5 & 16.7 & 54.2 & 70.8 & 29.2 & 54.2 & 0.0 \\
MiniMax-M3 & 0.0 & 0.0 & 0.0 & 0.0 & 0.0 & 0.0 & 0.0 \\
GLM-5.3 & 16.7 & 20.0 & 55.0 & 91.7 & 58.3 & 65.0 & 13.3 \\
\midrule
\multicolumn{8}{l}{\textit{Claude Code}} \\
\midrule
GPT-5.6-Sol & 28.8 & 30.0 & 86.3 & 93.8 & 57.5 & 90.0 & 8.8 \\
Claude Opus 5 & 9.0 & 14.1 & 41.0 & 96.2 & 29.5 & 30.8 & 0.0 \\
Qwen3.8-Max & 6.6 & 14.5 & 26.3 & 72.4 & 2.6 & 35.5 & 0.0 \\
Kimi K3 & 6.3 & 11.3 & 26.3 & 43.8 & 7.5 & 18.8 & 0.0 \\
DeepSeek-V4-Pro & 7.5 & 20.0 & 40.0 & 92.5 & 20.0 & 53.8 & 8.8 \\
MiniMax-M3 & 0.0 & 0.0 & 15.2 & 0.0 & 0.0 & 4.5 & 0.0 \\
GLM-5.3 & 0.0 & 0.0 & 0.0 & 0.0 & 0.0 & 0.0 & 0.0 \\
\midrule
\multicolumn{8}{l}{\textit{Hermes}} \\
\midrule
GPT-5.6-Sol & 47.5 & 56.3 & 88.8 & 100.0 & 66.3 & 93.8 & 18.8 \\
Claude Opus 5 & 4.2 & 12.5 & 30.6 & 98.6 & 30.6 & 26.4 & 0.0 \\
Qwen3.8-Max & 0.0 & 0.0 & 50.0 & 100.0 & 0.0 & 25.0 & 0.0 \\
Kimi K3 & 10.3 & 23.5 & 38.2 & 95.6 & 36.8 & 47.1 & 5.9 \\
DeepSeek-V4-Pro & 6.3 & 25.0 & 75.0 & 100.0 & 37.5 & 81.3 & 25.0 \\
MiniMax-M3 & 5.6 & 11.1 & 24.1 & 88.9 & 14.8 & 18.5 & 7.4 \\
GLM-5.3 & 8.0 & 28.0 & 38.0 & 94.0 & 12.0 & 32.0 & 8.0 \\
\midrule
\multicolumn{8}{l}{\textit{OpenClaw}} \\
\midrule
GPT-5.6-Sol & 39.1 & 46.9 & 70.3 & 95.3 & 32.8 & 87.5 & 18.8 \\
Claude Opus 5 & 25.0 & 0.0 & 25.0 & 100.0 & 25.0 & 50.0 & 0.0 \\
Qwen3.8-Max & 0.0 & 60.0 & 0.0 & 100.0 & 0.0 & 30.0 & 10.0 \\
Kimi K3 & 16.7 & 38.9 & 25.0 & 97.2 & 19.4 & 58.3 & 8.3 \\
DeepSeek-V4-Pro & 18.6 & 31.4 & 57.1 & 95.7 & 27.1 & 65.7 & 5.7 \\
MiniMax-M3 & 10.0 & 20.0 & 18.0 & 92.0 & 20.0 & 38.0 & 0.0 \\
GLM-5.3 & 25.0 & 100.0 & 100.0 & 100.0 & 0.0 & 50.0 & 0.0 \\
\bottomrule\end{tabular}}
\ifdim\wd0>\linewidth\resizebox{\linewidth}{!}{\usebox0}\else\usebox0\fi
\par\endgroup\par\vspace{8pt}
\noindent Completeness does not establish a reliable factual or strategic basis. Under Claude Code, GPT-5.6-Sol scores 93.8 on response completeness but only 28.8 on case understanding, 30.0 on response strategy, and 8.8 on factual and citation accuracy. Opus exhibits the same separation, with completeness at 96.2 and factual accuracy at 0.0. Hermes improves GPT's understanding and strategy scores to 47.5 and 56.3, yet factual accuracy remains 18.8 despite complete objection coverage. These results distinguish responding to the listed objections from interpreting the supplied materials accurately and building a defensible response.

\clearpage

\begingroup\small
\setlength{\tabcolsep}{5pt}\renewcommand{\arraystretch}{1.08}
\centering\captionsetup{hypcap=false}
\captionof{table}{OAR amendment criteria. G3: amendments addressing the objections; G4: proportionality and preservation of scope; G5: amendment legality and form. Required but omitted amendments retain zero scores. Each criterion is averaged over its applicable scored cases on a 0--100 scale. Missing outputs and failed grading are excluded. }
\label{tab:app_oar_amendment}
\sbox0{\begin{tabular}{@{}lccc@{}}
\toprule
Model & G3 & G4 & G5 \\
\midrule
\multicolumn{4}{l}{\textit{Codex}} \\
\midrule
GPT-5.6-Sol & 0.0 & 3.8 & 3.8 \\
Qwen3.8-Max & 17.9 & 60.7 & 21.4 \\
Kimi K3 & 0.0 & 50.0 & 33.3 \\
DeepSeek-V4-Pro & 0.0 & 20.0 & 20.0 \\
MiniMax-M3 & 0.0 & 0.0 & 0.0 \\
GLM-5.3 & 8.7 & 26.1 & 17.4 \\
\midrule
\multicolumn{4}{l}{\textit{Claude Code}} \\
\midrule
GPT-5.6-Sol & 12.0 & 24.0 & 20.0 \\
Claude Opus 5 & 50.0 & 50.0 & 50.0 \\
Qwen3.8-Max & 6.9 & 22.4 & 6.9 \\
Kimi K3 & 6.1 & 6.1 & 6.1 \\
DeepSeek-V4-Pro & 4.3 & 39.1 & 13.0 \\
MiniMax-M3 & 0.0 & 1.6 & 0.0 \\
GLM-5.3 & 0.0 & 0.0 & 0.0 \\
\midrule
\multicolumn{4}{l}{\textit{Hermes}} \\
\midrule
GPT-5.6-Sol & 43.3 & 50.0 & 40.0 \\
Claude Opus 5 & 5.6 & 16.7 & 11.1 \\
Qwen3.8-Max & 0.0 & 0.0 & 0.0 \\
Kimi K3 & 8.7 & 30.4 & 10.9 \\
DeepSeek-V4-Pro & 0.0 & 20.0 & 0.0 \\
MiniMax-M3 & 5.6 & 13.9 & 5.6 \\
GLM-5.3 & 11.9 & 45.2 & 19.0 \\
\midrule
\multicolumn{4}{l}{\textit{OpenClaw}} \\
\midrule
GPT-5.6-Sol & 34.1 & 59.1 & 43.2 \\
Claude Opus 5 & 0.0 & 0.0 & 0.0 \\
Qwen3.8-Max & 0.0 & 0.0 & 0.0 \\
Kimi K3 & 28.6 & 50.0 & 21.4 \\
DeepSeek-V4-Pro & 11.4 & 38.6 & 22.7 \\
MiniMax-M3 & 10.5 & 36.8 & 18.4 \\
GLM-5.3 & 0.0 & 0.0 & 0.0 \\
\bottomrule\end{tabular}}
\ifdim\wd0>\linewidth\resizebox{\linewidth}{!}{\usebox0}\else\usebox0\fi
\par\endgroup\par\vspace{8pt}
\noindent Amendment quality is limited even when the need for amendment is recognized. Under Claude Code, GPT-5.6-Sol scores 12.0 for matching amendments to the objections, 24.0 for proportionality, and 20.0 for legality and form across 25 applicable cases. Under Hermes, the corresponding means are 43.3, 50.0, and 40.0 across 30 applicable cases. Opus reaches 50.0 on all three criteria under Claude Code, but only four cases are applicable. These results identify a concrete failure in executing suitable amendments, while the differing applicable case sets prevent treating the raw means as a controlled ranking of amendment ability.

\clearpage

\begingroup\small
\setlength{\tabcolsep}{5pt}\renewcommand{\arraystretch}{1.08}
\centering\captionsetup{hypcap=false}
\captionof{table}{OAR inventive-step argumentation. G6.1.1: closest prior art and distinguishing features; G6.1.2: technical effects and objective technical problem; G6.1.3: other references and common knowledge; G6.1.4: motivation to combine. Each criterion is averaged over its applicable scored cases on a 0--100 scale. Missing outputs and failed grading are excluded.}
\label{tab:app_oar_inventive}
\sbox0{\begin{tabular}{@{}lcccc@{}}
\toprule
Model & G6.1.1 & G6.1.2 & G6.1.3 & G6.1.4 \\
\midrule
\multicolumn{5}{l}{\textit{Codex}} \\
\midrule
GPT-5.6-Sol & 46.3 & 25.9 & 45.8 & 46.3 \\
Qwen3.8-Max & 29.2 & 30.8 & 40.0 & 25.0 \\
Kimi K3 & 66.7 & 33.3 & 0.0 & 50.0 \\
DeepSeek-V4-Pro & 40.0 & 10.0 & 20.0 & 15.0 \\
MiniMax-M3 & 0.0 & 0.0 & 0.0 & 0.0 \\
GLM-5.3 & 52.5 & 27.5 & 31.6 & 23.7 \\
\midrule
\multicolumn{5}{l}{\textit{Claude Code}} \\
\midrule
GPT-5.6-Sol & 40.7 & 27.8 & 50.0 & 51.9 \\
Claude Opus 5 & 42.3 & 20.4 & 33.3 & 30.8 \\
Qwen3.8-Max & 24.0 & 14.0 & 21.7 & 8.0 \\
Kimi K3 & 19.2 & 7.7 & 20.8 & 11.5 \\
DeepSeek-V4-Pro & 33.9 & 8.9 & 25.0 & 11.1 \\
MiniMax-M3 & 0.0 & 0.0 & 0.0 & 0.0 \\
GLM-5.3 & 0.0 & 0.0 & 0.0 & 0.0 \\
\midrule
\multicolumn{5}{l}{\textit{Hermes}} \\
\midrule
GPT-5.6-Sol & 57.4 & 51.9 & 52.1 & 53.7 \\
Claude Opus 5 & 27.1 & 14.6 & 40.0 & 20.5 \\
Qwen3.8-Max & 0.0 & 0.0 & 0.0 & 0.0 \\
Kimi K3 & 25.0 & 25.0 & 21.1 & 16.7 \\
DeepSeek-V4-Pro & 62.5 & 0.0 & 25.0 & 0.0 \\
MiniMax-M3 & 26.5 & 8.8 & 10.7 & 9.4 \\
GLM-5.3 & 38.5 & 23.1 & 13.6 & 26.9 \\
\midrule
\multicolumn{5}{l}{\textit{OpenClaw}} \\
\midrule
GPT-5.6-Sol & 52.5 & 45.0 & 55.9 & 50.0 \\
Claude Opus 5 & 0.0 & 0.0 & 0.0 & 0.0 \\
Qwen3.8-Max & 0.0 & 0.0 & 50.0 & 0.0 \\
Kimi K3 & 43.8 & 12.5 & 42.9 & 12.5 \\
DeepSeek-V4-Pro & 40.9 & 20.5 & 21.1 & 14.3 \\
MiniMax-M3 & 40.0 & 10.0 & 23.1 & 14.3 \\
GLM-5.3 & 50.0 & 50.0 & 50.0 & 0.0 \\
\bottomrule\end{tabular}}
\ifdim\wd0>\linewidth\resizebox{\linewidth}{!}{\usebox0}\else\usebox0\fi
\par\endgroup\par\vspace{8pt}
\noindent Connecting technical differences to their effects and the objective technical problem is a specific weakness in the Claude Code results. GPT-5.6-Sol scores 40.7 on prior-art and feature analysis but 27.8 on effects and problem formulation; Opus scores 42.3 and 20.4, respectively. For both models, the latter is the lowest of the four inventive-step criteria. GPT with Hermes scores between 51.9 and 57.4 across these criteria, showing more balanced argumentation. Identifying a distinguishing feature alone therefore does not establish the reasoning needed to defend inventive step.

\clearpage

\begingroup\small
\setlength{\tabcolsep}{5pt}\renewcommand{\arraystretch}{1.08}
\centering\captionsetup{hypcap=false}
\captionof{table}{OAR responses to other examination issues. G6.2: novelty; G6.3: eligibility; G6.4: disclosure, enablement, and written description; G6.5: clarity, support, and claim form; G6.6: essential features. Each criterion is averaged over its applicable scored cases on a 0--100 scale. Missing outputs and failed grading are excluded. }
\label{tab:app_oar_objections}
\sbox0{\begin{tabular}{@{}lccccc@{}}
\toprule
Model & G6.2 & G6.3 & G6.4 & G6.5 & G6.6 \\
\midrule
\multicolumn{6}{l}{\textit{Codex}} \\
\midrule
GPT-5.6-Sol & 36.7 & 50.0 & 16.7 & 20.4 & 50.0 \\
Qwen3.8-Max & 38.9 & 100.0 & 100.0 & 26.9 & 50.0 \\
Kimi K3 & 75.0 & 0.0 & 50.0 & 25.0 & 25.0 \\
DeepSeek-V4-Pro & 0.0 & 0.0 & 0.0 & 6.3 & 0.0 \\
MiniMax-M3 & 0.0 & 0.0 & 0.0 & 0.0 & 0.0 \\
GLM-5.3 & 31.8 & 25.0 & 0.0 & 28.9 & 0.0 \\
\midrule
\multicolumn{6}{l}{\textit{Claude Code}} \\
\midrule
GPT-5.6-Sol & 36.7 & 50.0 & 50.0 & 31.5 & 25.0 \\
Claude Opus 5 & 16.7 & 25.0 & 0.0 & 22.2 & 0.0 \\
Qwen3.8-Max & 7.7 & 0.0 & 0.0 & 16.7 & 0.0 \\
Kimi K3 & 10.0 & 0.0 & 0.0 & 9.3 & 25.0 \\
DeepSeek-V4-Pro & 18.8 & 50.0 & 50.0 & 18.5 & 50.0 \\
MiniMax-M3 & 0.0 & 0.0 & 0.0 & 0.0 & 0.0 \\
GLM-5.3 & 0.0 & 0.0 & 0.0 & 0.0 & 0.0 \\
\midrule
\multicolumn{6}{l}{\textit{Hermes}} \\
\midrule
GPT-5.6-Sol & 56.7 & 75.0 & 37.5 & 51.9 & 50.0 \\
Claude Opus 5 & 23.3 & 75.0 & 37.5 & 25.0 & 25.0 \\
Qwen3.8-Max & 0.0 & 0.0 & 0.0 & 0.0 & 0.0 \\
Kimi K3 & 30.8 & 50.0 & 100.0 & 23.9 & 0.0 \\
DeepSeek-V4-Pro & 0.0 & 0.0 & 100.0 & 8.3 & 0.0 \\
MiniMax-M3 & 15.0 & 50.0 & 50.0 & 19.4 & 50.0 \\
GLM-5.3 & 30.0 & 100.0 & 25.0 & 31.0 & 25.0 \\
\midrule
\multicolumn{6}{l}{\textit{OpenClaw}} \\
\midrule
GPT-5.6-Sol & 50.0 & 50.0 & 75.0 & 60.0 & 75.0 \\
Claude Opus 5 & 0.0 & 0.0 & 0.0 & 25.0 & 0.0 \\
Qwen3.8-Max & 50.0 & 0.0 & 0.0 & 12.5 & 50.0 \\
Kimi K3 & 8.3 & 0.0 & 100.0 & 27.3 & 50.0 \\
DeepSeek-V4-Pro & 23.1 & 0.0 & 0.0 & 23.9 & 50.0 \\
MiniMax-M3 & 14.3 & 50.0 & 50.0 & 25.0 & 100.0 \\
GLM-5.3 & 50.0 & 0.0 & 0.0 & 0.0 & 0.0 \\
\bottomrule\end{tabular}}
\ifdim\wd0>\linewidth\resizebox{\linewidth}{!}{\usebox0}\else\usebox0\fi
\par\endgroup\par\vspace{8pt}
\noindent Novelty and clarity objections remain only partially resolved even for GPT-5.6-Sol. Under Claude Code, its novelty score is 36.7 across 15 applicable cases and its clarity/support score is 31.5 across 27; with Hermes, these rise to 56.7 and 51.9 over the same numbers of applicable cases. The results show a harness-dependent change in the quality of substantive responses for these two objection types. They do not establish a difficulty ranking across objection types: eligibility and essential-feature scores for GPT with Hermes each cover only two cases, while clarity/support covers 27.

\clearpage\subsection{Patent Translation: ZH--EN}

\begingroup
\small
\setlength{\tabcolsep}{5pt}
\renewcommand{\arraystretch}{1.08}
\centering
\captionsetup{hypcap=false}
\captionof{table}{ZH--EN translation: key quality measures. The suite size counts 80 document-level cases, with 881 items. The metric-specific evaluation sets contain 421 ordinary-text items, 24 document-accuracy items for SacreBLEU, and 205 and 124 terminology items for accuracy and consistency, respectively.}
\label{tab:app_zh_en_key}
\sbox0{\begin{tabular}{@{}lcccc@{}}
\toprule
Model & BLEU-1 & \makecell{Document\\SacreBLEU} & \makecell{Term\\accuracy} & \makecell{Term\\consistency} \\
\midrule
\multicolumn{5}{l}{\textit{Codex}} \\
\midrule
GPT-5.6-Sol & 67.5 & 63.0 & 53.7 & 46.8 \\
Qwen3.8-Max & 69.1 & 65.4 & 54.6 & 50.8 \\
Kimi K3 & 70.5 & 67.2 & 58.1 & 47.6 \\
DeepSeek-V4-Pro & 70.6 & 64.6 & 56.6 & 53.2 \\
MiniMax-M3 & 69.7 & 64.4 & 52.7 & 46.0 \\
GLM-5.3 & 70.2 & 64.2 & 57.6 & 47.6 \\
\midrule
\multicolumn{5}{l}{\textit{Claude Code}} \\
\midrule
GPT-5.6-Sol & 67.1 & 60.8 & 54.2 & 46.0 \\
Claude Opus 5 & 70.0 & 67.3 & 55.6 & 50.0 \\
Qwen3.8-Max & 69.8 & 64.8 & 55.6 & 51.6 \\
Kimi K3 & 70.5 & 66.8 & 58.5 & 49.2 \\
DeepSeek-V4-Pro & 70.8 & 67.5 & 58.5 & 49.2 \\
MiniMax-M3 & 68.6 & 65.0 & 53.2 & 47.6 \\
GLM-5.3 & 70.8 & 64.3 & 59.0 & 50.0 \\
\midrule
\multicolumn{5}{l}{\textit{Hermes}} \\
\midrule
GPT-5.6-Sol & 67.6 & 63.5 & 53.2 & 46.0 \\
Claude Opus 5 & 69.8 & 66.5 & 57.1 & 49.2 \\
Qwen3.8-Max & 70.3 & 59.2 & 54.6 & 49.2 \\
Kimi K3 & 69.1 & 67.2 & 56.6 & 48.4 \\
DeepSeek-V4-Pro & 70.7 & 65.4 & 56.6 & 51.6 \\
MiniMax-M3 & 68.6 & 64.7 & 56.6 & 43.6 \\
GLM-5.3 & 69.8 & 54.6 & 55.1 & 49.2 \\
\midrule
\multicolumn{5}{l}{\textit{OpenClaw}} \\
\midrule
GPT-5.6-Sol & 68.1 & 63.5 & 54.6 & 45.2 \\
Claude Opus 5 & 70.4 & 66.7 & 57.6 & 48.4 \\
Qwen3.8-Max & 69.6 & 54.0 & 55.6 & 48.4 \\
Kimi K3 & 70.1 & 50.6 & 57.1 & 50.0 \\
DeepSeek-V4-Pro & 71.1 & 61.8 & 59.5 & 53.2 \\
MiniMax-M3 & 68.9 & 51.2 & 52.7 & 47.6 \\
GLM-5.3 & 72.0 & 68.5 & 57.1 & 50.0 \\
\bottomrule
\end{tabular}}
\ifdim\wd0>\linewidth\resizebox{\linewidth}{!}{\usebox0}\else\usebox0\fi
\par\endgroup
\par\vspace{8pt}\noindent

\noindent The strongest Chinese-to-English text-overlap scores do not identify the strongest terminology performance. GLM-5.3 with OpenClaw leads BLEU-1 and document SacreBLEU at 72.0 and 68.5, but DeepSeek-V4-Pro with OpenClaw leads terminology accuracy at 59.5 and shares the highest consistency score of 53.2. Terminology remains incomplete even at these maxima. Harness differences also depend on text length: Qwen3.8-Max has similar BLEU-1 scores with Codex and OpenClaw, at 69.1 and 69.6, but document SacreBLEU changes from 65.4 to 54.0. Ordinary-text overlap therefore misses both document-level differences and the separate requirement to select and reuse patent terminology correctly.

\clearpage\subsection{Patent Translation: EN--ZH}

\begingroup
\small
\setlength{\tabcolsep}{5pt}
\renewcommand{\arraystretch}{1.08}
\centering
\captionsetup{hypcap=false}
\captionof{table}{EN--ZH translation: key quality measures. The suite size counts 80 document-level cases with 881 items. The metric-specific evaluation sets contain 421 ordinary-text items, 24 document-accuracy items for SacreBLEU, and 205 and 124 terminology items for accuracy and consistency, respectively. }
\label{tab:app_en_zh_key}
\sbox0{\begin{tabular}{@{}lcccc@{}}
\toprule
Model & ROUGE-1 & \makecell{Document\\SacreBLEU} & \makecell{Term\\accuracy} & \makecell{Term\\consistency} \\
\midrule
\multicolumn{5}{l}{\textit{Codex}} \\
\midrule
GPT-5.6-Sol & 76.1 & 65.8 & 66.3 & 75.8 \\
Qwen3.8-Max & 75.3 & 68.2 & 64.4 & 73.4 \\
Kimi K3 & 77.4 & 67.5 & 65.4 & 75.8 \\
DeepSeek-V4-Pro & 77.0 & 70.0 & 66.8 & 75.0 \\
MiniMax-M3 & 75.1 & 67.0 & 63.9 & 67.7 \\
GLM-5.3 & 76.4 & 66.7 & 67.3 & 73.4 \\
\midrule
\multicolumn{5}{l}{\textit{Claude Code}} \\
\midrule
GPT-5.6-Sol & 75.3 & 65.7 & 66.3 & 78.2 \\
Claude Opus 5 & 76.8 & 65.4 & 67.3 & 75.8 \\
Qwen3.8-Max & 77.1 & 65.4 & 63.9 & 75.8 \\
Kimi K3 & 77.2 & 68.2 & 68.8 & 78.2 \\
DeepSeek-V4-Pro & 76.4 & 70.6 & 68.8 & 75.8 \\
MiniMax-M3 & 74.9 & 66.4 & 62.9 & 64.5 \\
GLM-5.3 & 77.9 & 63.9 & 69.3 & 77.4 \\
\midrule
\multicolumn{5}{l}{\textit{Hermes}} \\
\midrule
GPT-5.6-Sol & 75.7 & 65.8 & 65.9 & 79.0 \\
Claude Opus 5 & 77.9 & 66.3 & 65.4 & 74.2 \\
Qwen3.8-Max & 76.6 & 57.9 & 63.4 & 75.0 \\
Kimi K3 & 76.3 & 69.4 & 65.9 & 72.6 \\
DeepSeek-V4-Pro & 76.7 & 69.5 & 67.3 & 74.2 \\
MiniMax-M3 & 75.2 & 63.6 & 64.4 & 70.2 \\
GLM-5.3 & 77.0 & 55.9 & 66.3 & 75.8 \\
\midrule
\multicolumn{5}{l}{\textit{OpenClaw}} \\
\midrule
GPT-5.6-Sol & 74.0 & 66.7 & 64.4 & 79.0 \\
Claude Opus 5 & 77.8 & 9.1 & 68.8 & 77.4 \\
Qwen3.8-Max & 76.0 & 49.7 & 62.9 & 73.4 \\
Kimi K3 & 77.4 & 59.1 & 65.9 & 77.4 \\
DeepSeek-V4-Pro & 77.6 & 65.1 & 66.3 & 72.6 \\
MiniMax-M3 & 75.0 & 53.7 & 62.0 & 67.7 \\
GLM-5.3 & 78.1 & 55.3 & 68.3 & 79.8 \\
\bottomrule
\end{tabular}}
\ifdim\wd0>\linewidth\resizebox{\linewidth}{!}{\usebox0}\else\usebox0\fi
\par\endgroup
\par\vspace{8pt}\noindent

\noindent Strong ordinary-text overlap does not ensure strong document translation in English-to-Chinese tasks. Opus with Hermes and OpenClaw has nearly identical ROUGE-1 scores of 77.9 and 77.8, but document SacreBLEU reaches 66.3 with Hermes and only 9.1 with OpenClaw. This contrast establishes a document-level difference that the ordinary-text score does not capture. Terminology accuracy and consistency also separate: GLM-5.3 with OpenClaw scores 79.8 on consistency but 68.3 on accuracy. Reusing a term consistently does not establish that the term was selected correctly. Document translation and terminology thus remain necessary evaluation dimensions alongside ordinary-text overlap.

\clearpage\section{Qualitative Case Comparisons}\label{app:case_comparisons}

Each scenario compares models under a fixed harness and harnesses with a fixed model. Panels retain the curated records and original score precision. These selected runs illustrate local differences; full-set statistics are reported separately.

\subsection{Novelty Search}

\begin{center}\captionsetup{hypcap=false}

\begin{minipage}{0.78\linewidth}\centering

\includegraphics[width=\linewidth]{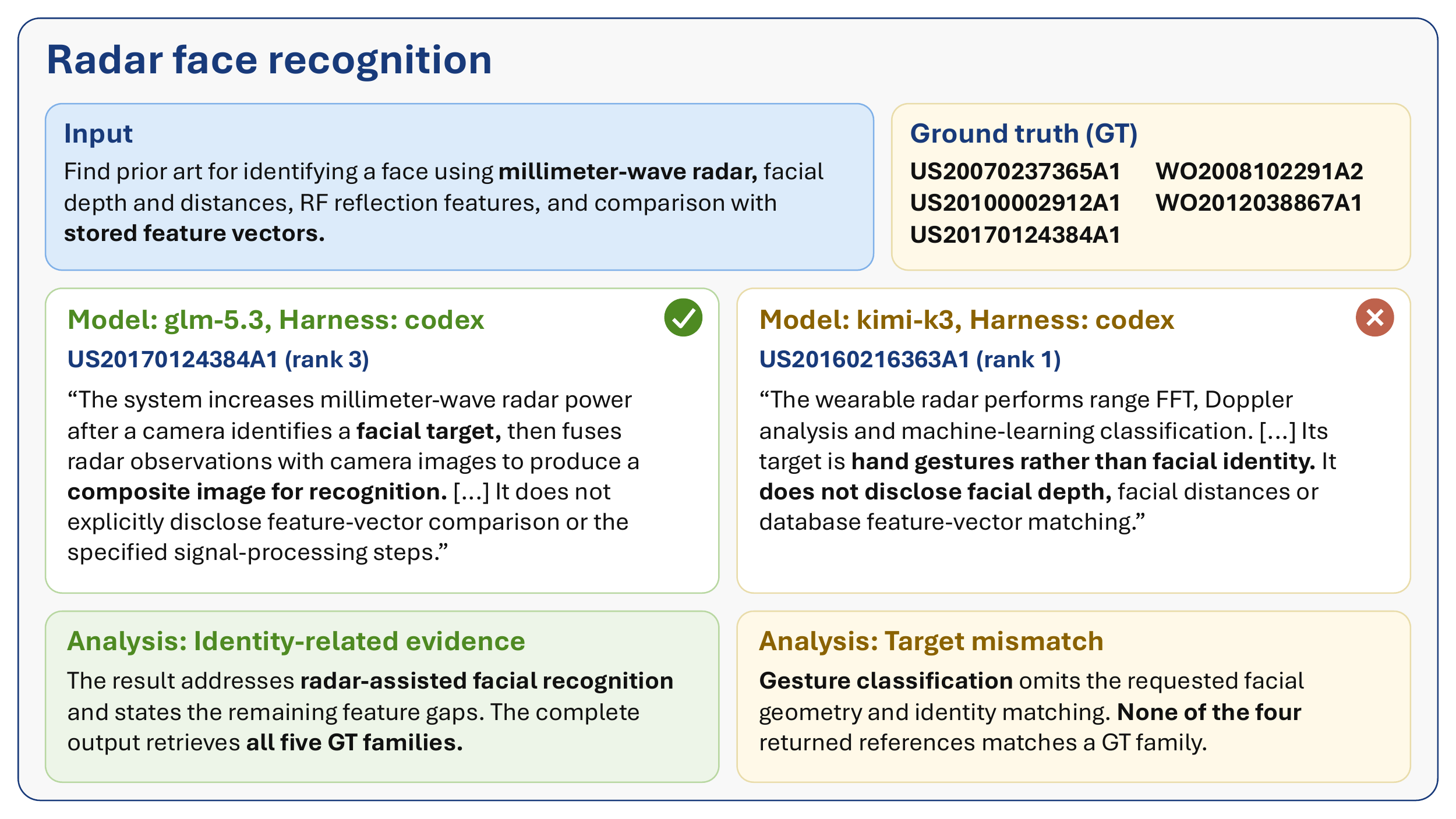}

\par\small (a) Codex: GLM-5.3 versus Kimi K3.

\end{minipage}\par\vspace{4pt}

\begin{minipage}{0.78\linewidth}\centering

\includegraphics[width=\linewidth]{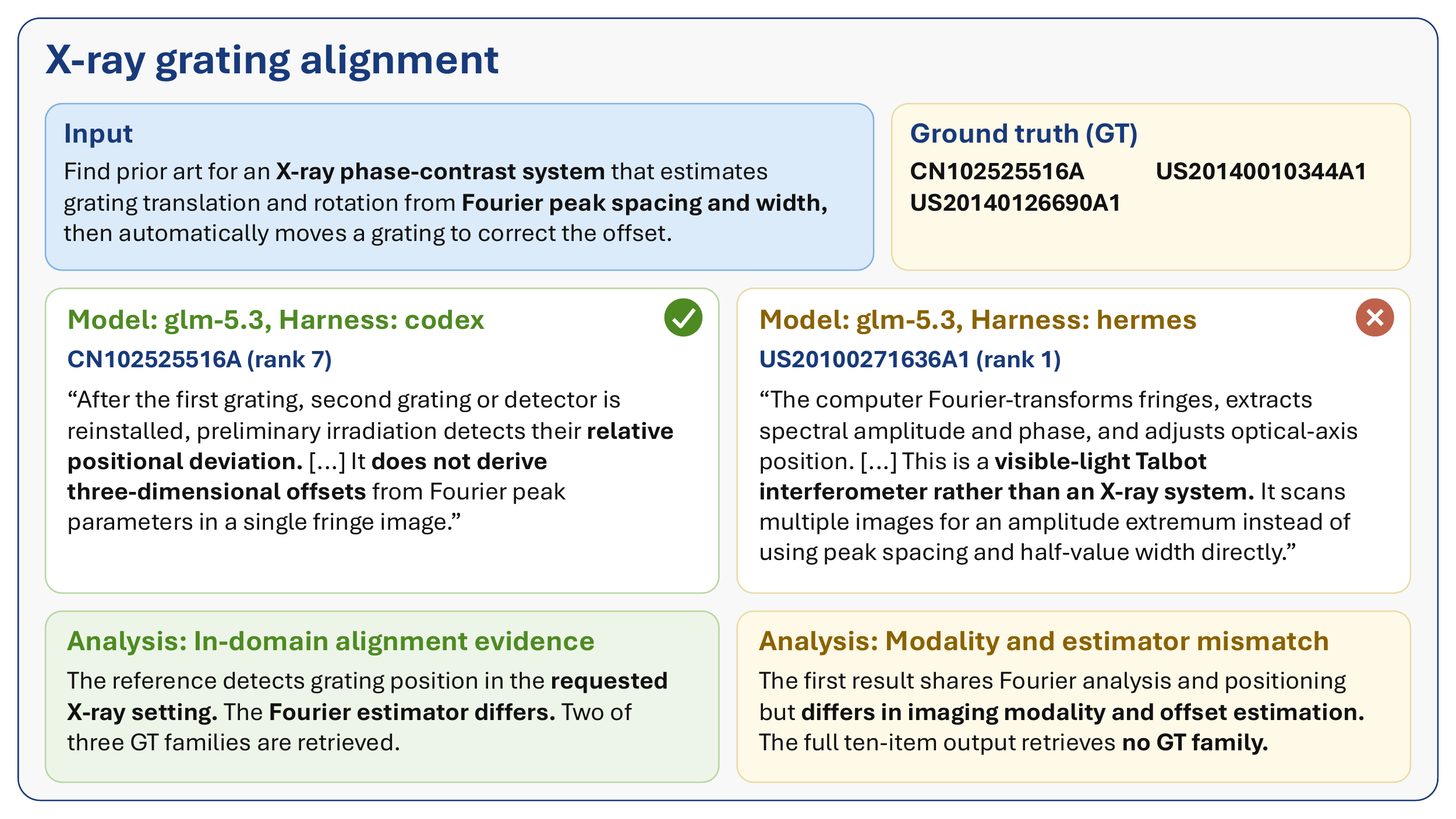}

\par\small (b) GLM-5.3: Codex versus Hermes.

\end{minipage}\par\vspace{4pt}

\captionof{figure}{Novelty Search: fixed-harness comparison above and fixed-model comparison below.}

\label{fig:case_novelty_search}

\end{center}

\noindent Target mismatches cause missed reference families in both comparisons. In the radar-face case, GLM with Codex retrieves all five annotated families, whereas Kimi retrieves none and prioritizes a hand-gesture patent. In the X-ray alignment case, Codex retrieves two of three annotated families, whereas Hermes retrieves none and prioritizes a visible-light Talbot reference. Both contrasts expose a specific mismatch: shared radar or Fourier terminology does not preserve the target object, imaging modality, and measurement procedure.

\clearpage

\subsection{Patent FTO}

\begin{center}\captionsetup{hypcap=false}

\begin{minipage}{0.88\linewidth}\centering

\includegraphics[width=\linewidth]{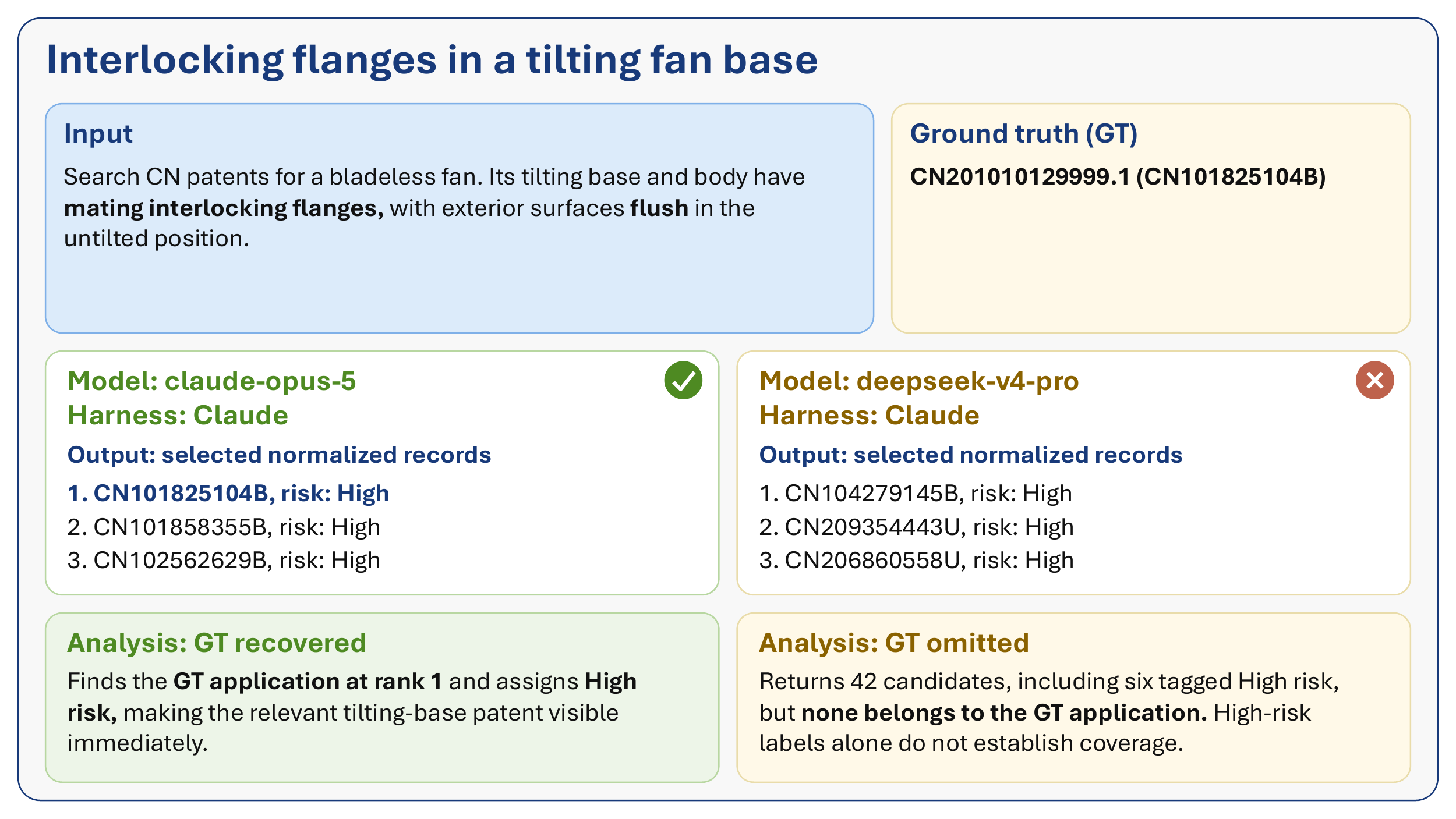}

\par\small (a) Claude Code: Opus versus DeepSeek-V4-Pro.

\end{minipage}\par\vspace{4pt}

\begin{minipage}{0.88\linewidth}\centering

\includegraphics[width=\linewidth]{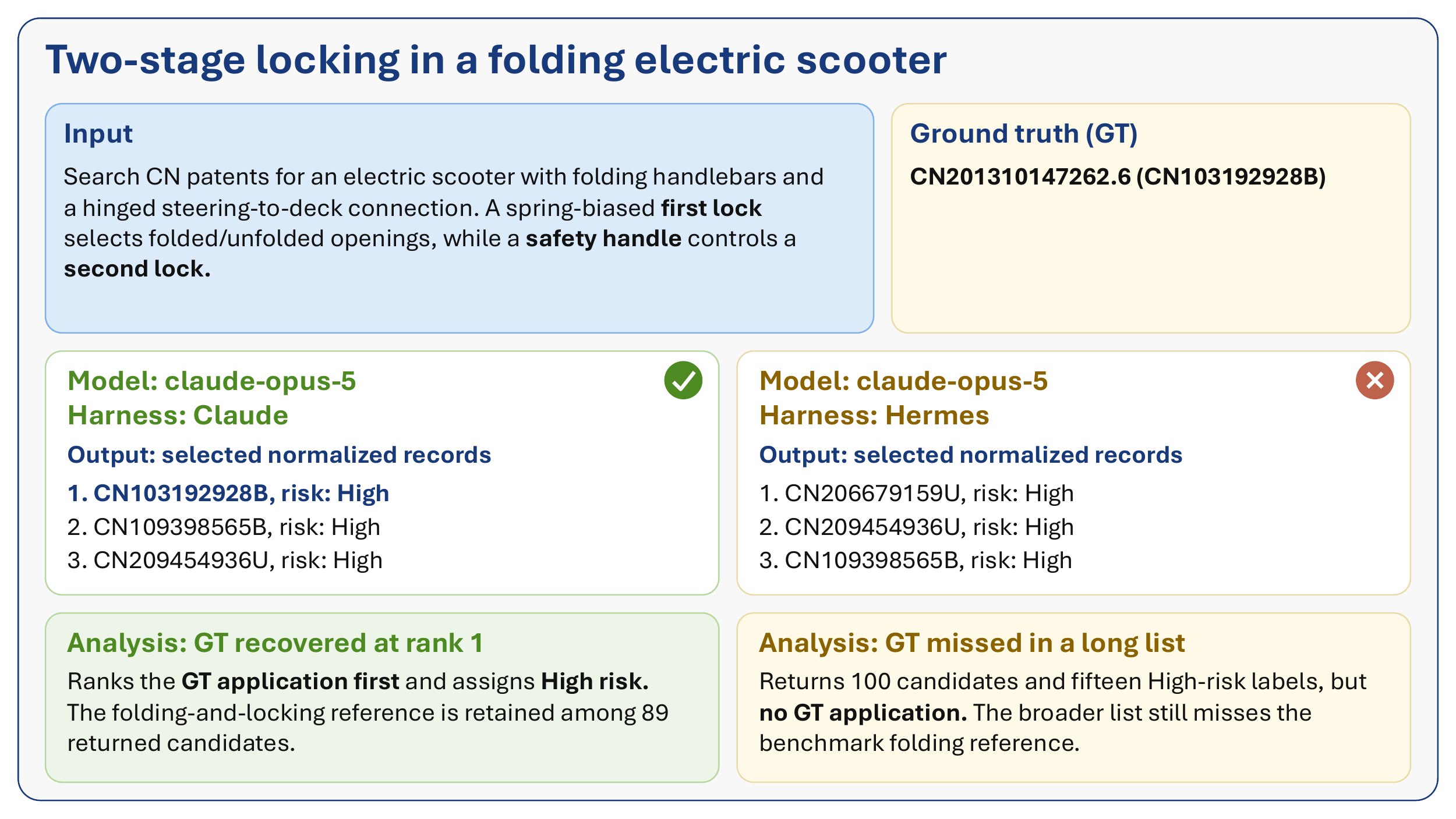}

\par\small (b) Opus: Claude Code versus Hermes.

\end{minipage}\par\vspace{4pt}

\captionof{figure}{Patent FTO: fixed-harness comparison above and fixed-model comparison below.}

\label{fig:case_fto}

\end{center}

\noindent More candidates and high-risk labels do not ensure reference coverage. For the tilting fan, Opus with Claude Code ranks the annotated application first and labels it high risk. DeepSeek returns 42 candidates, including six high-risk candidates, but misses that application. For the folding scooter, Claude Code ranks the annotated two-stage locking reference first among 89 candidates. Hermes returns 100 candidates, including 15 high-risk candidates, but misses it. In both comparisons, a longer or explicitly risk-labeled list fails to recover the mechanism-specific reference.

\clearpage

\subsection{Design FTO}

\begin{center}\captionsetup{hypcap=false}

\begin{minipage}{0.88\linewidth}\centering

\includegraphics[width=\linewidth]{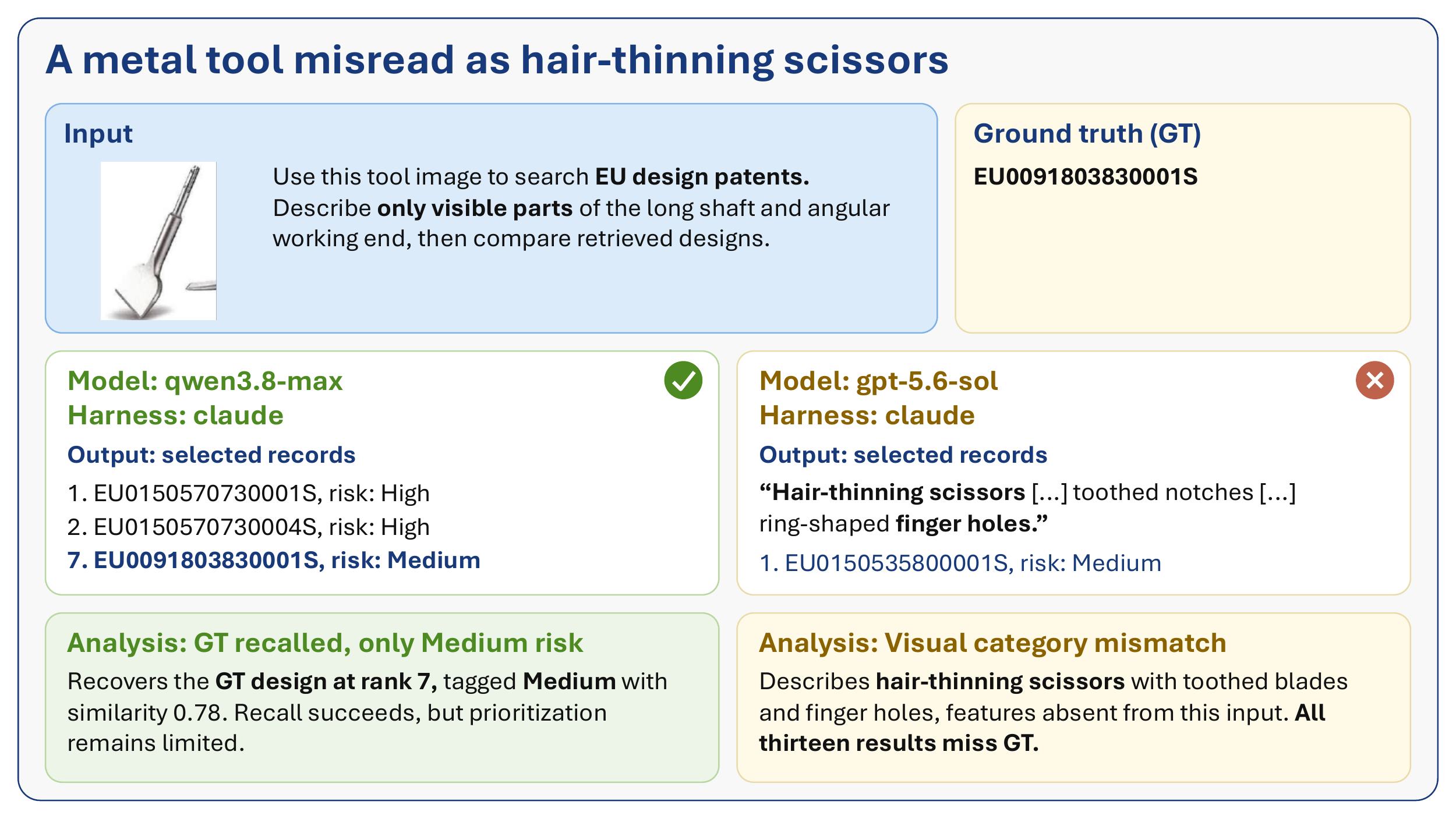}

\par\small (a) Claude Code: Qwen3.8-Max versus GPT-5.6-Sol.

\end{minipage}\par\vspace{4pt}

\begin{minipage}{0.88\linewidth}\centering

\includegraphics[width=\linewidth]{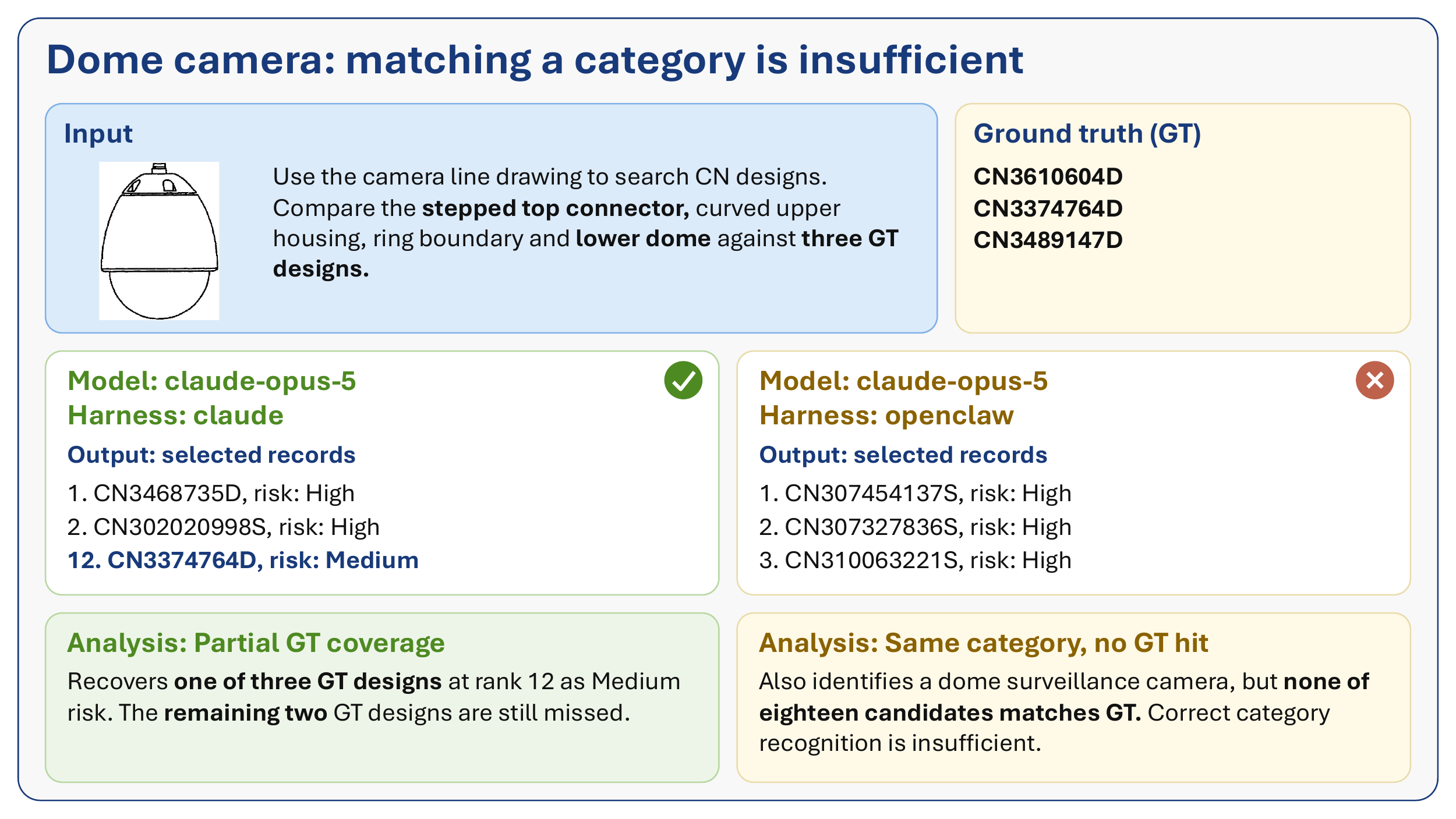}

\par\small (b) Opus: Claude Code versus OpenClaw.

\end{minipage}\par\vspace{4pt}

\captionof{figure}{Design FTO: fixed-harness comparison above and fixed-model comparison below.}

\label{fig:case_design_fto}

\end{center}

\noindent Visual misclassification prevents retrieval, while correct broad classification remains insufficient. For the metal tool, Qwen retrieves the annotated design at rank seven. GPT instead describes hair-thinning scissors with finger holes absent from the input and misses the reference. For the dome camera, OpenClaw identifies the broad product category but retrieves none of the three annotated designs. Claude Code retrieves one at rank twelve with a medium-risk label. The first contrast identifies a category error; the second establishes that category recognition alone does not recover the relevant appearance details. The latter result is partial coverage, not complete retrieval.

\clearpage

\subsection{Invalidity Search}

\begin{center}\captionsetup{hypcap=false}

\begin{minipage}{0.88\linewidth}\centering

\includegraphics[width=\linewidth]{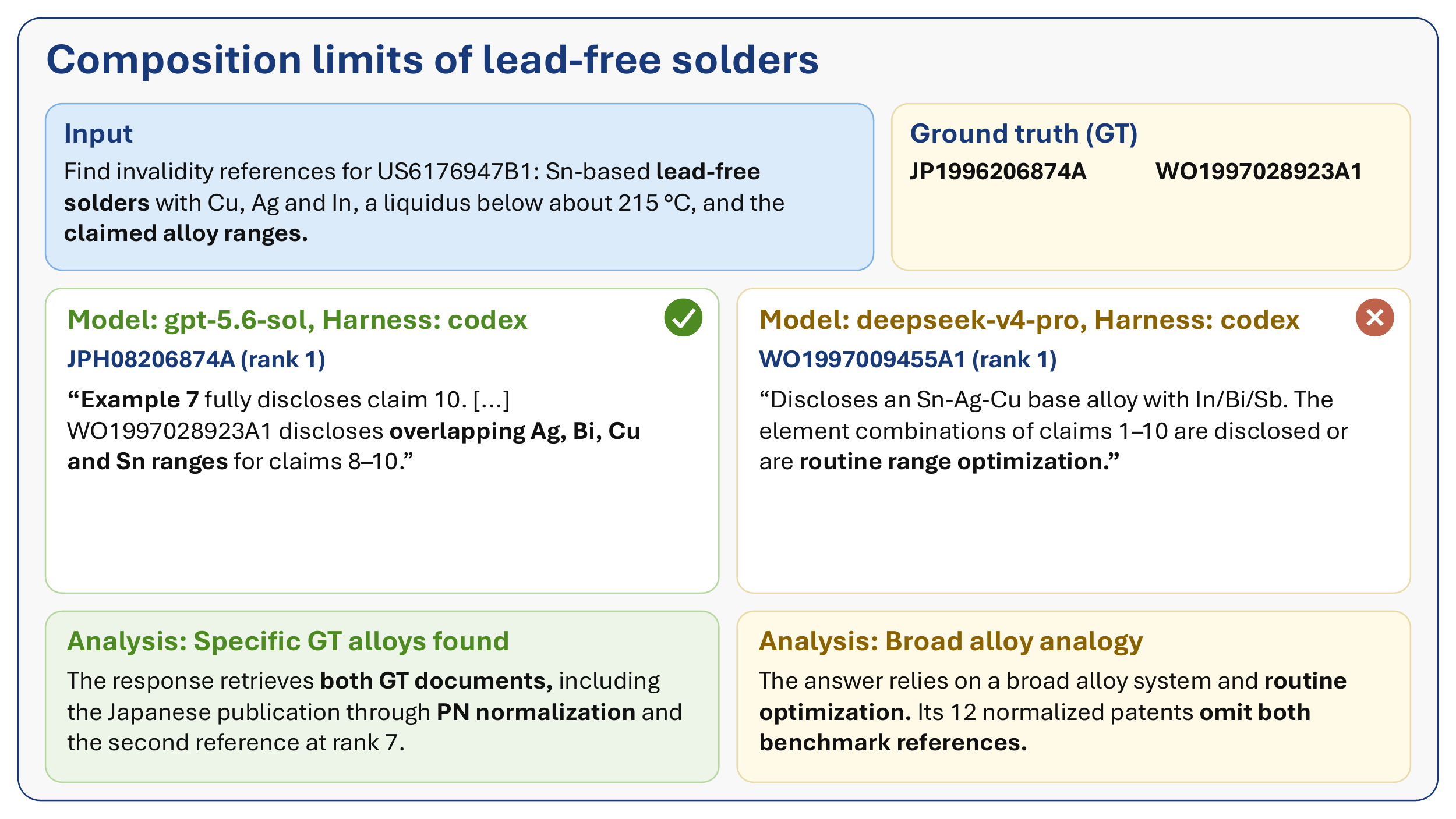}

\par\small (a) Codex: GPT-5.6-Sol versus DeepSeek-V4-Pro.

\end{minipage}\par\vspace{4pt}

\begin{minipage}{0.88\linewidth}\centering

\includegraphics[width=\linewidth]{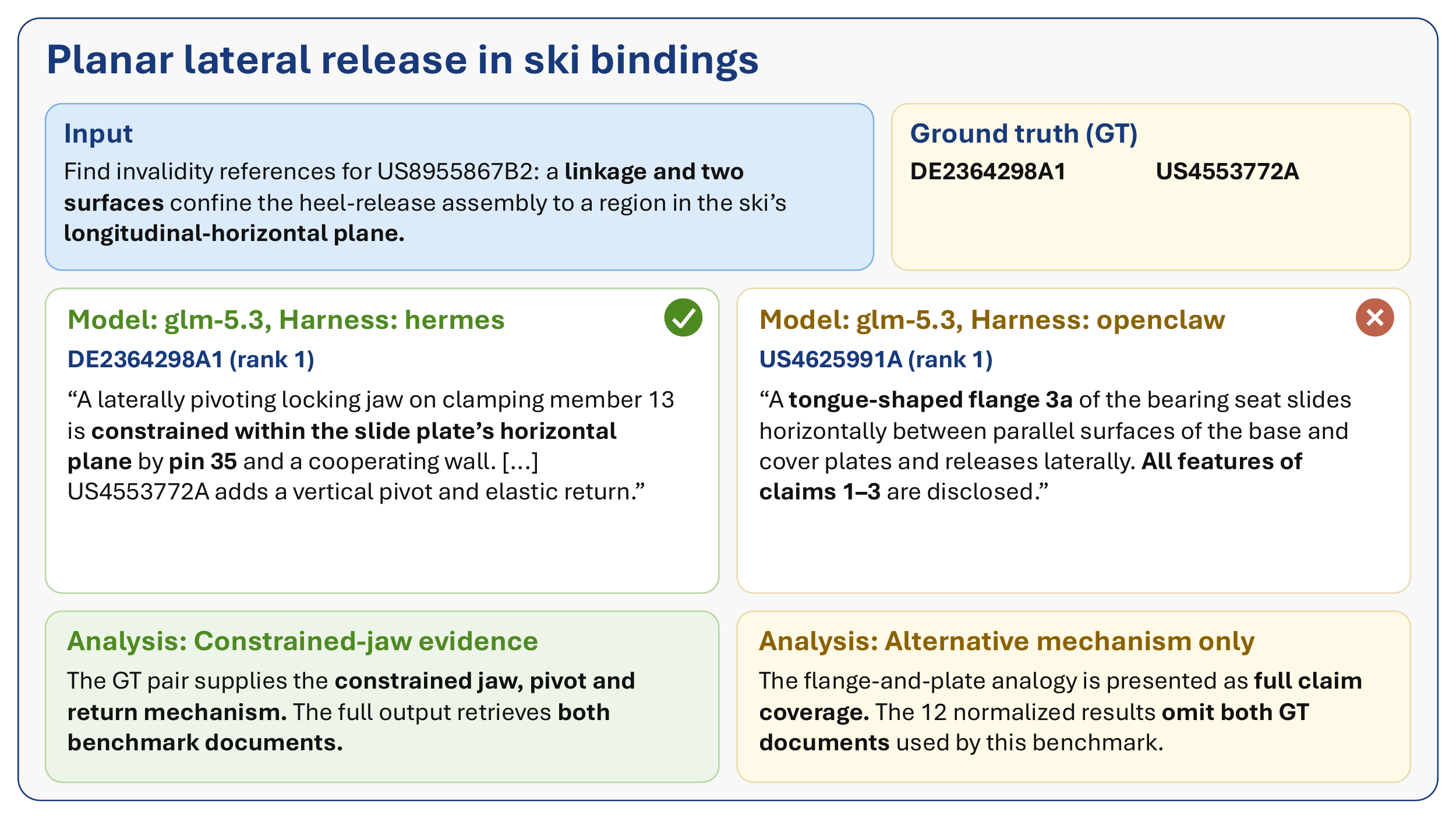}

\par\small (b) GLM-5.3: Hermes versus OpenClaw.

\end{minipage}\par\vspace{4pt}

\captionof{figure}{Invalidity Search: fixed-harness comparison above and fixed-model comparison below.}

\label{fig:case_invalid_search}

\end{center}

\noindent Broad technical similarity does not replace mechanism-specific prior art. For the lead-free solder, GPT with Codex retrieves both annotated references, while DeepSeek misses both and relies on broader alloy ranges and routine-optimization arguments. For the ski binding, Hermes retrieves both annotated references and identifies the jaw, wall, and return arrangement that constrains release. OpenClaw misses both and asserts coverage using a different flange-and-plate mechanism. These cases distinguish retrieval of the designated technical evidence from a plausible general argument; neither the argument nor a reference hit by itself establishes a legal invalidity conclusion.

\clearpage

\subsection{Claim Charting}

\begin{center}\captionsetup{hypcap=false}

\begin{minipage}{0.88\linewidth}\centering

\includegraphics[width=\linewidth]{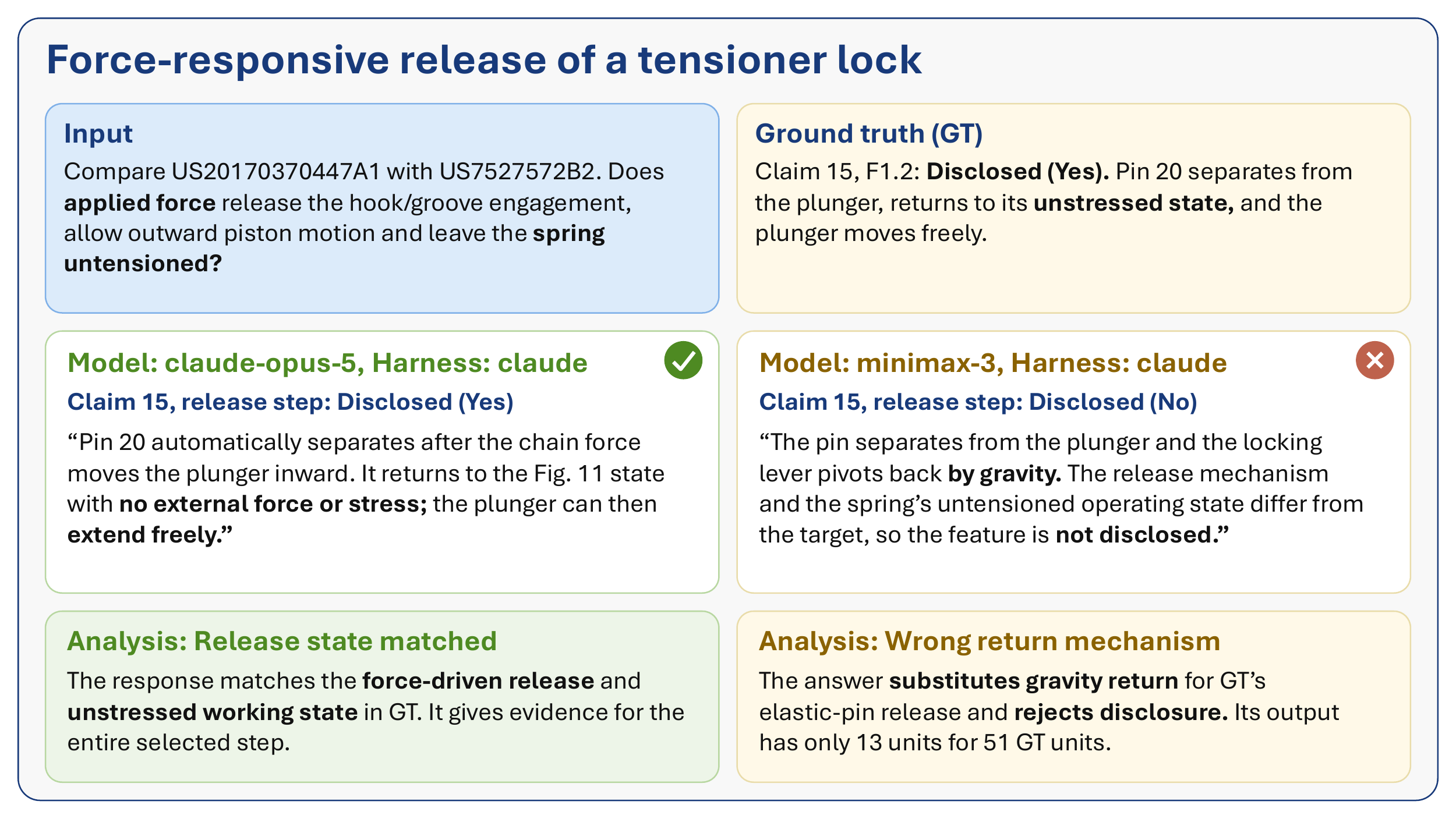}

\par\small (a) Claude Code: Opus versus MiniMax-M3.

\end{minipage}\par\vspace{4pt}

\begin{minipage}{0.88\linewidth}\centering

\includegraphics[width=\linewidth]{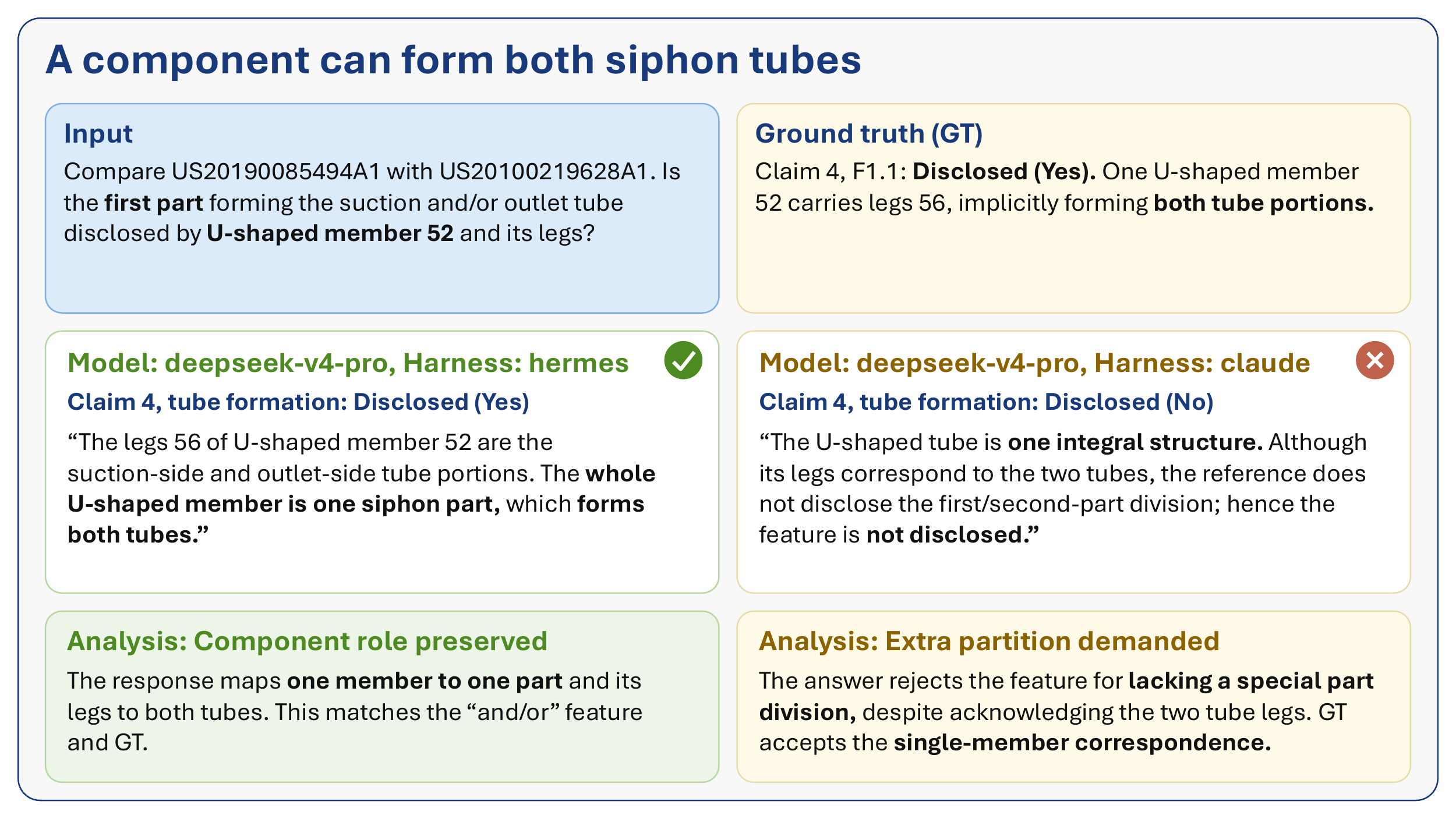}

\par\small (b) DeepSeek-V4-Pro: Hermes versus Claude Code.

\end{minipage}\par\vspace{4pt}

\captionof{figure}{Claim Charting: fixed-harness comparison above and fixed-model comparison below.}

\label{fig:case_ai_cc}

\end{center}

\noindent Incorrect mechanism reconstruction reverses feature-level disclosure judgments. For the tensioner, Opus correctly maps release of the pin to the untensioned spring and free plunger. MiniMax substitutes a gravity-return mechanism and incorrectly marks the feature as absent. For the siphon, Hermes maps the two legs of one U-shaped component to the suction and outlet portions. Claude Code recognizes the legs but adds a separate-partition requirement and marks the feature as absent. In both cases, the negative judgment arises from changing the mechanism or adding a condition beyond the feature being compared.

\clearpage

\subsection{Patent Drafting}

\begin{center}\captionsetup{hypcap=false}

\begin{minipage}{0.88\linewidth}\centering

\includegraphics[width=\linewidth]{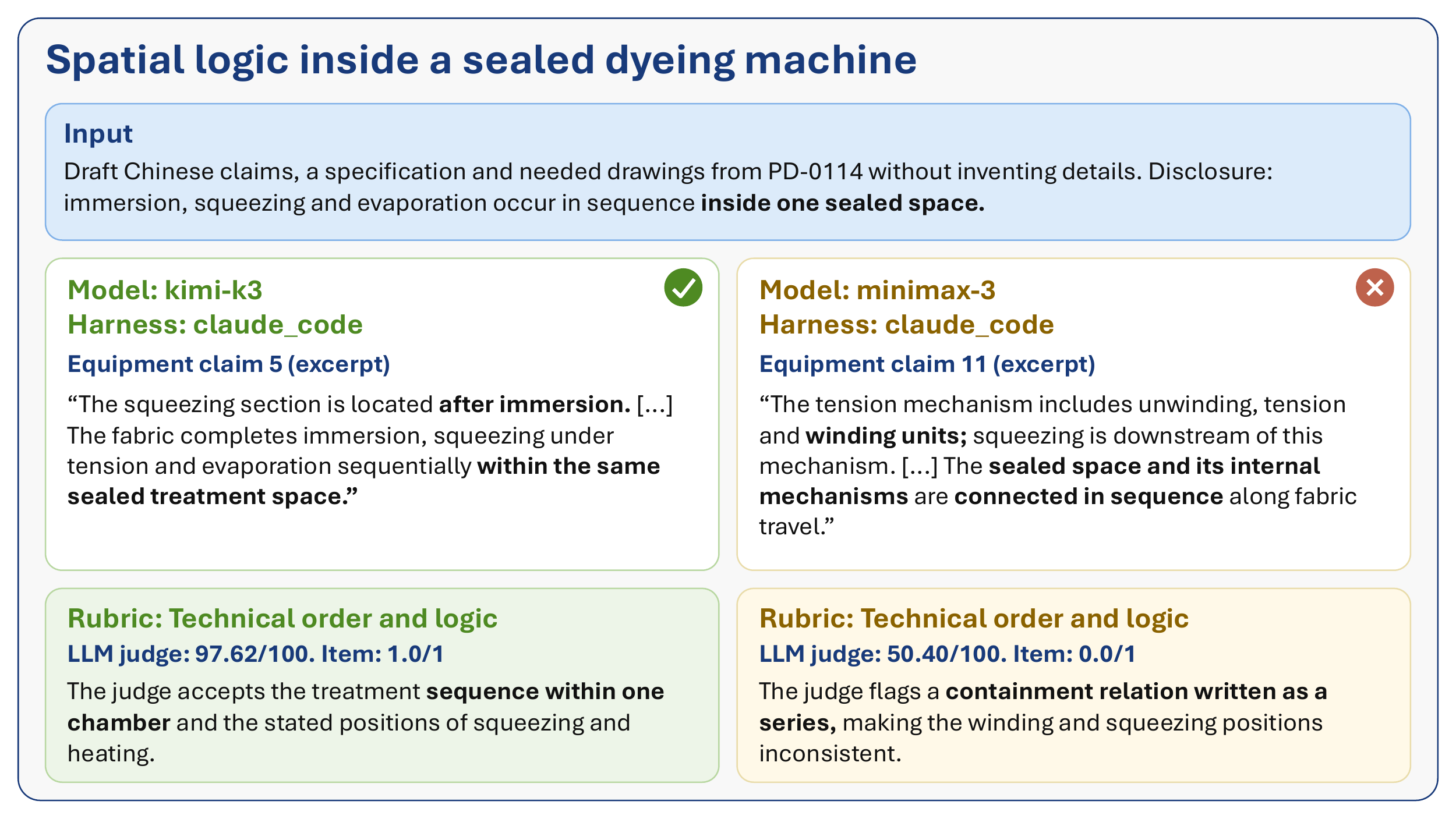}

\par\small (a) Claude Code: Kimi K3 versus MiniMax-M3.

\end{minipage}\par\vspace{4pt}

\begin{minipage}{0.88\linewidth}\centering

\includegraphics[width=\linewidth]{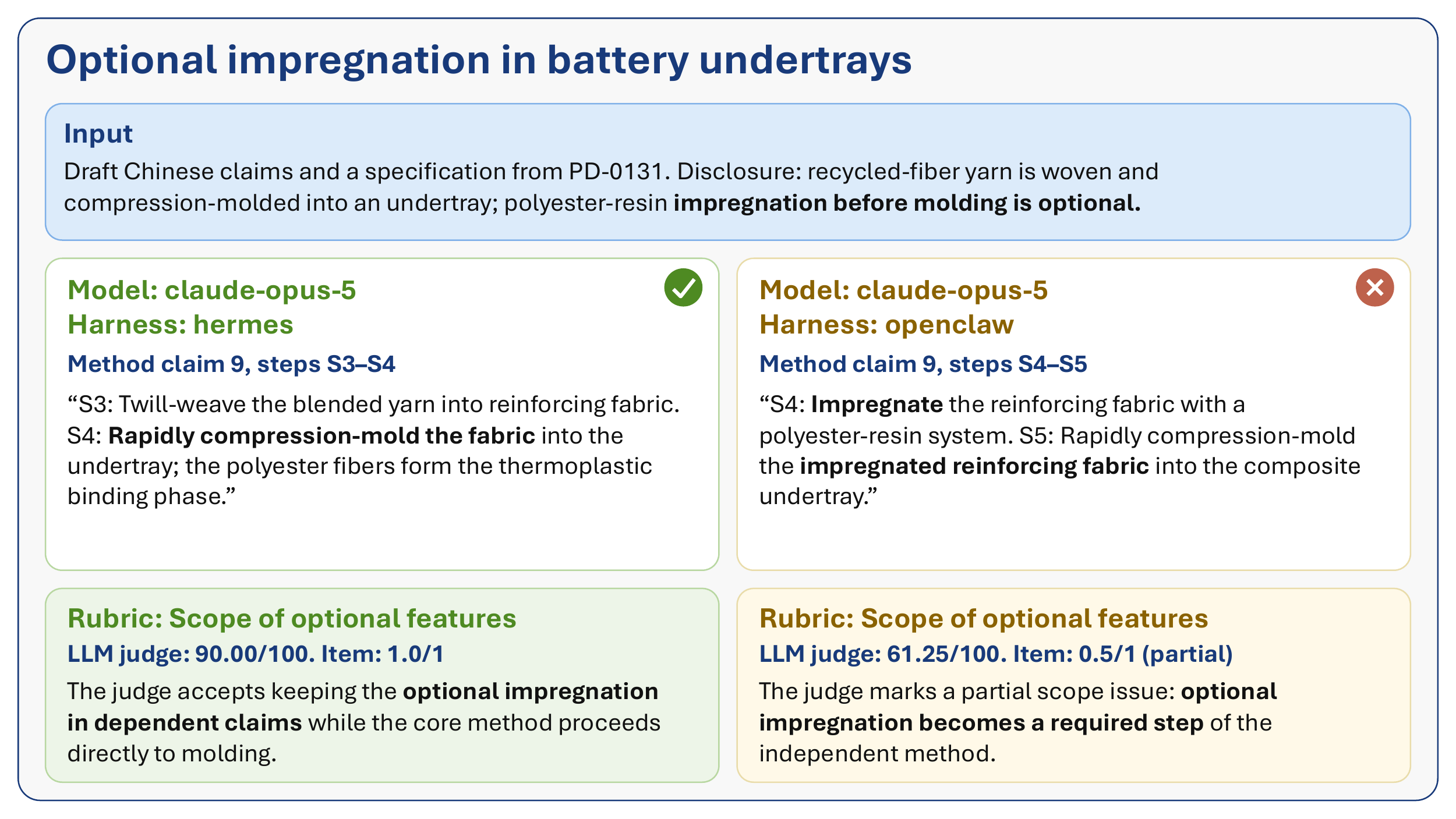}

\par\small (b) Opus: Hermes versus OpenClaw.

\end{minipage}\par\vspace{4pt}

\captionof{figure}{Patent Drafting: fixed-harness comparison above and fixed-model comparison below.}

\label{fig:case_drafting}

\end{center}

\noindent Claim quality depends on preserving disclosed relationships and optionality. For the dyeing machine, Kimi preserves the immersion, squeezing, and evaporation sequence within one chamber; MiniMax confuses containment with processing order and misplaces components. For the battery undertray, Hermes keeps optional resin impregnation in dependent claims, while OpenClaw makes it mandatory in the independent method claim. The first error changes the disclosed apparatus relationship; the second narrows the claimed method by requiring an optional step. These identified rubric errors explain local quality differences without attributing the entire document-score gap to a single criterion.

\clearpage

\subsection{Office-Action Response}
\begin{center}\captionsetup{hypcap=false}
\begin{minipage}{0.88\linewidth}\centering
\includegraphics[width=\linewidth]{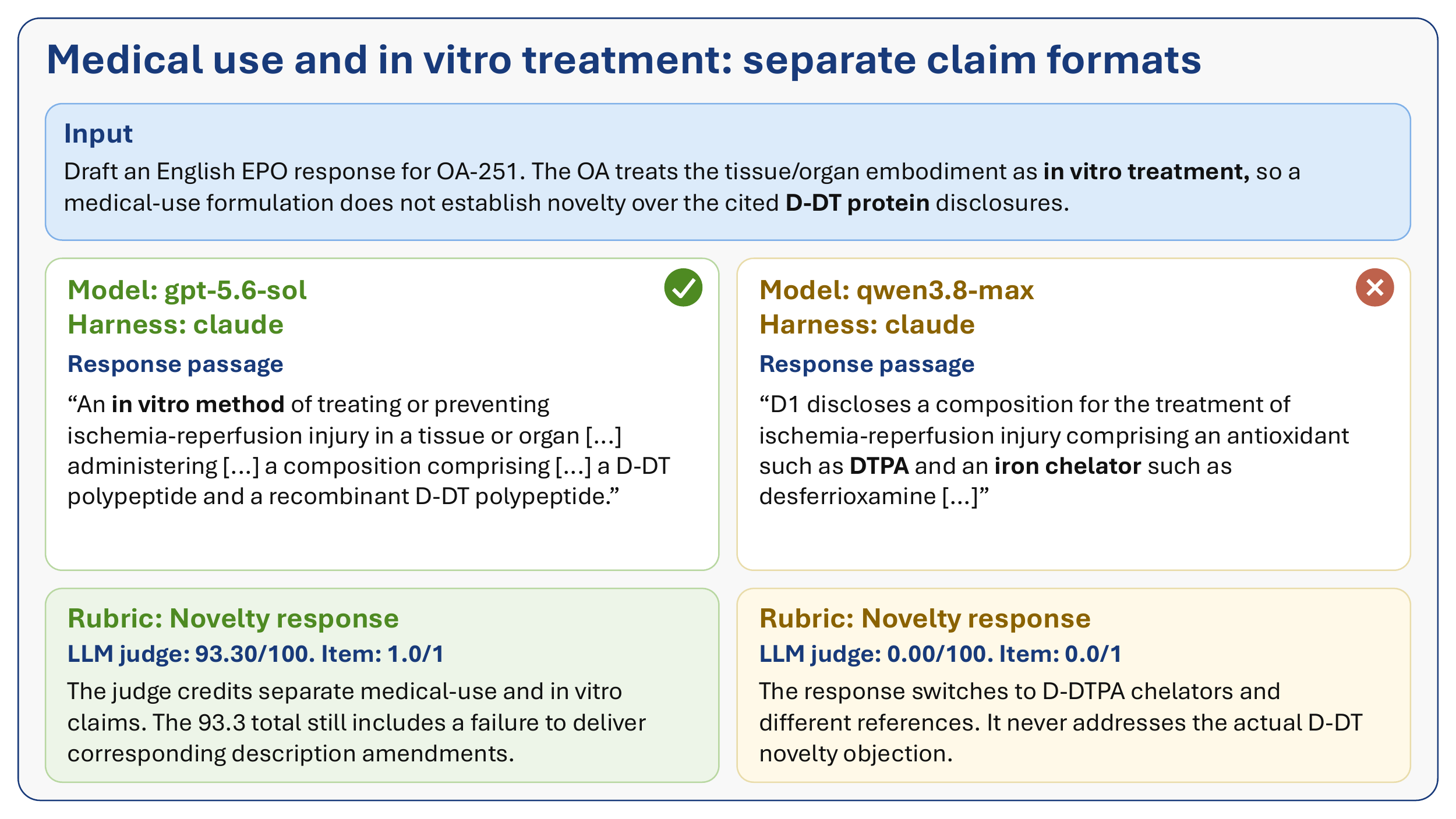}
\par\small(a) Claude Code: GPT-5.6-Sol versus Qwen3.8-Max.
\end{minipage}\par\vspace{4pt}
\begin{minipage}{0.88\linewidth}\centering
\includegraphics[width=\linewidth]{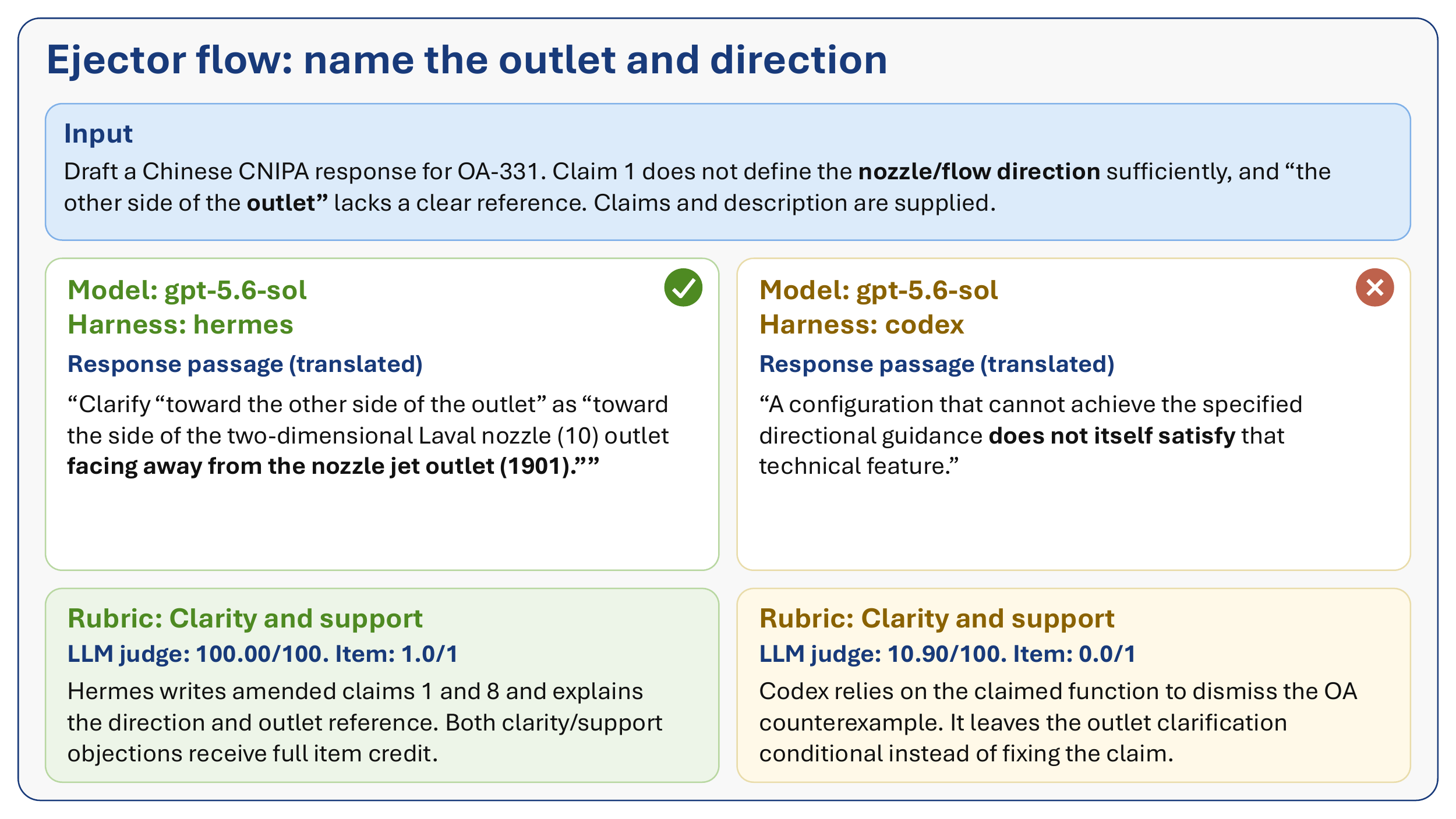}
\par\small(b) GPT-5.6-Sol: Hermes versus Codex.
\end{minipage}\par\vspace{4pt}
\captionof{figure}{Office-action response: fixed-harness comparison above and fixed-model comparison below. The excerpts show how case interpretation and amendment execution affect the response to a specific objection.}
\label{fig:case_oar}
\end{center}

\noindent An effective response must address the actual technical issue and implement the required correction. In the medical-use case, the EPO distinguishes treatment of a patient from in vitro treatment of tissue or an organ. Under Claude Code, GPT-5.6-Sol separates the claim formats and receives full credit on the novelty-response criterion. Qwen instead discusses DTPA chelators and different references, leaving the actual D-DT novelty objection unanswered. In the ejector case, holding GPT fixed isolates a difference in amendment execution: Hermes specifies the outlet reference and flow direction in amended claims, whereas Codex relies on the claimed function to dismiss the objection and leaves the clarification conditional. The clarity/support criterion assigns 1.0 to Hermes and 0.0 to Codex. Both comparisons show that a substantive response requires preserving the case facts and translating the objection into a concrete argument or amendment. GPT's 93.3 total in the first case still includes a failure to supply corresponding description amendments, so full credit on the illustrated criterion does not imply that every response requirement has been met.

\clearpage
\section{Execution Failure Cases}
\label{app:execution_failure_cases}
\begingroup
The traces illustrate failures to complete a task within the available time and to proceed from extracted document text to a final response.

\begin{center}\captionsetup{hypcap=false}
\includegraphics[width=0.86\linewidth]{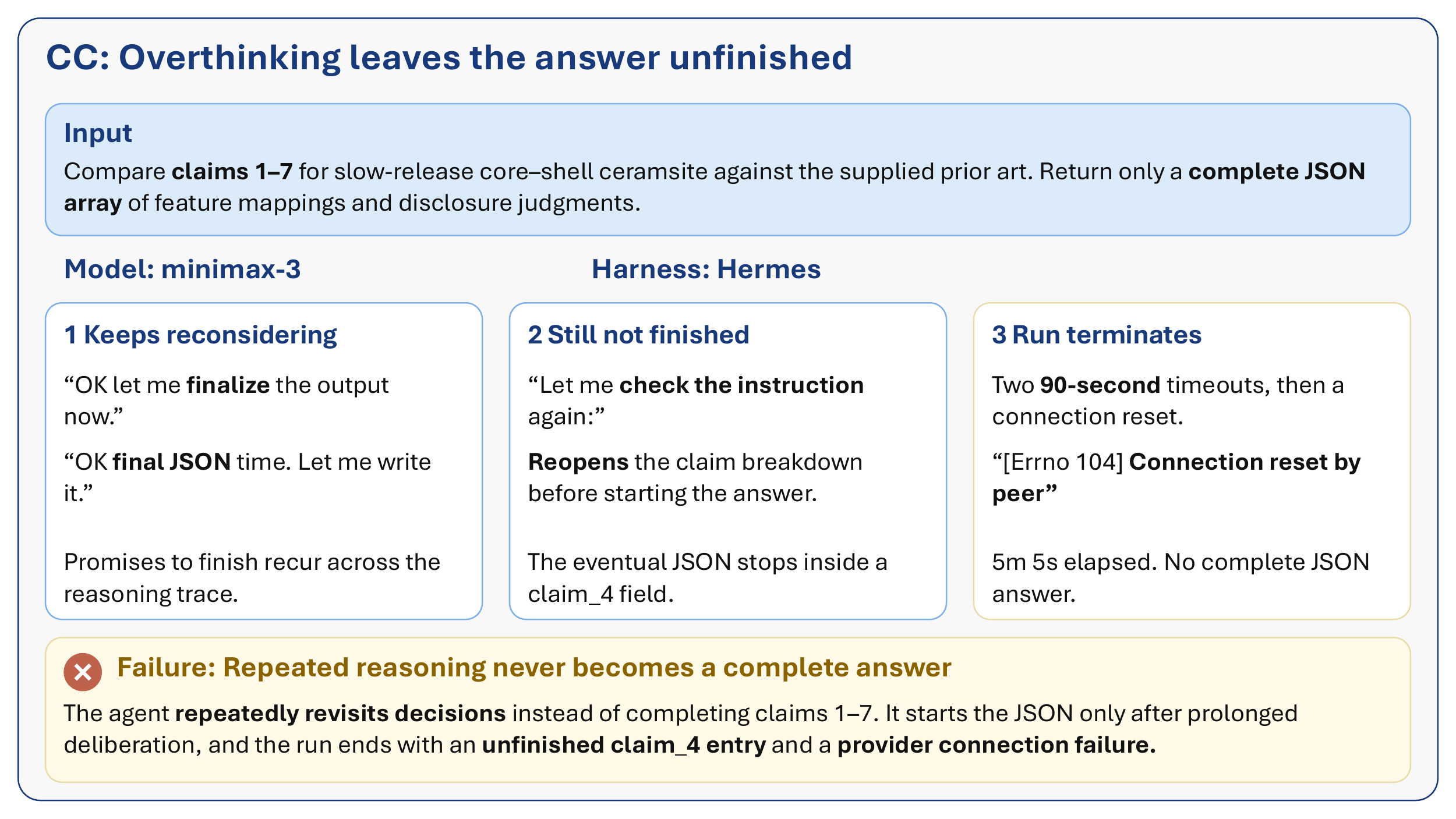}
\captionof{figure}{Claim-charting failure: MiniMax-M3 repeatedly revisits its reasoning, begins the answer late, and leaves the requested JSON unfinished.}
\label{fig:cc_execution_failure}
\end{center}
\noindent MiniMax is asked to compare seven claims and return a complete JSON array. It repeatedly announces that it will finalize the answer, then reopens the analysis. When the run terminates after timeouts and a connection reset, the output remains unfinished within claim 4. The model does not complete the task within the allotted time. Its extended reasoning fails to produce the complete deliverable required for evaluation.

\begin{center}\captionsetup{hypcap=false}
\includegraphics[width=0.86\linewidth]{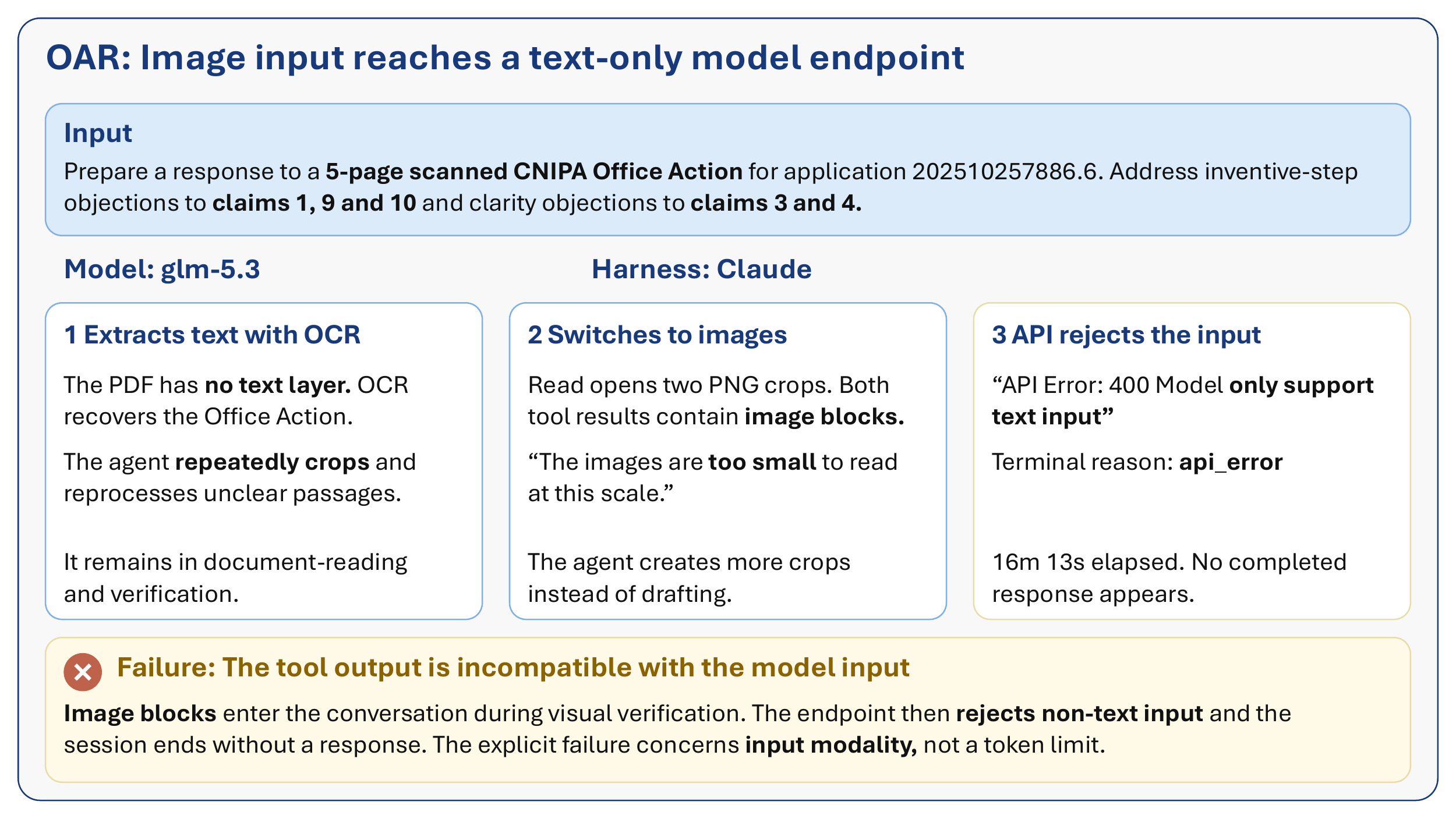}
\captionof{figure}{OAR failure: GLM-5.3 with Claude Code receives image blocks from a tool, and its text-only endpoint rejects the request before the response is completed.}
\label{fig:oar_execution_failure}
\end{center}
\noindent For a five-page scanned CNIPA office action, Claude Code performs OCR and makes the extracted text available. GLM then requests PNG crops to check unclear passages, but its text-only endpoint rejects the returned images. No response is delivered. Although the harness has extracted the document text, the model fails to choose a viable next step and carry the task through to drafting. This case exposes a failure to use available intermediate results to complete the task.
\endgroup
\clearpage

\clearpage\section{Task Prompts}\label{app:task_prompts}
The following panels reproduce the supplied English display versions of the task prompts. They summarize the task instructions, input placeholders, constraints, and output contracts, rather than reproduce the executed templates verbatim. Requirements for Chinese output are retained. Both translation directions are shown separately.
\begin{center}\captionsetup{hypcap=false}
\begin{minipage}{1.0\linewidth}\centering
\includegraphics[page=1,width=\linewidth]{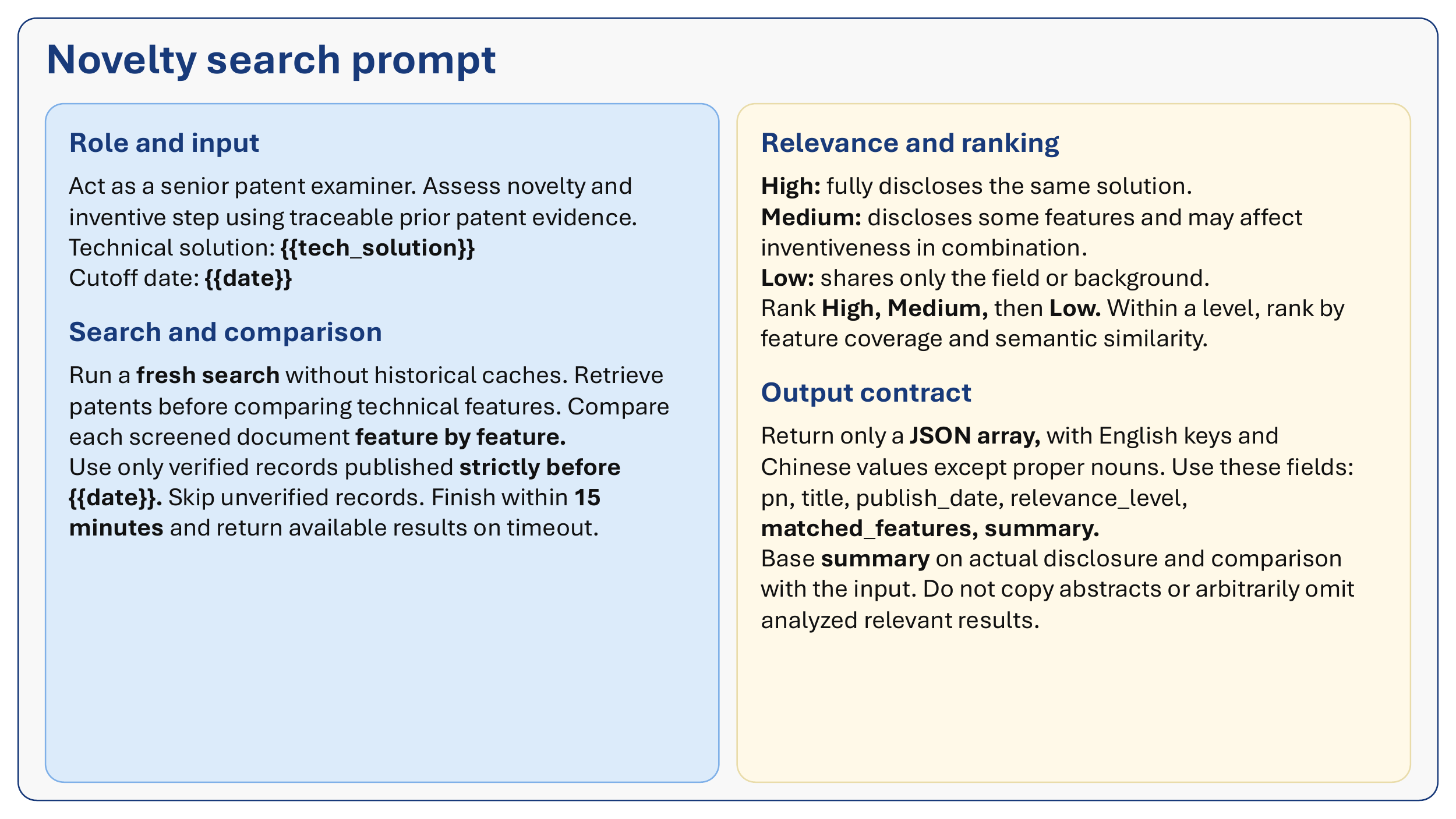}
\par\small (a) Novelty Search
\end{minipage}\par\vspace{6pt}
\begin{minipage}{1.0\linewidth}\centering
\includegraphics[page=2,width=\linewidth]{images/fig13_task_prompts.pdf}
\par\small (b) Patent FTO
\end{minipage}\par\vspace{6pt}
\captionof{figure}{English prompt displays for Novelty Search and Patent FTO.}
\label{fig:prompts_1}
\end{center}

\clearpage\begin{center}\captionsetup{hypcap=false}
\begin{minipage}{1.0\linewidth}\centering
\includegraphics[page=3,width=\linewidth]{images/fig13_task_prompts.pdf}
\par\small (a) Design FTO
\end{minipage}\par\vspace{6pt}
\begin{minipage}{1.0\linewidth}\centering
\includegraphics[page=8,width=\linewidth]{images/fig13_task_prompts.pdf}
\par\small (b) Invalidity Search
\end{minipage}\par\vspace{6pt}
\captionof{figure}{English prompt displays for Design FTO and Invalidity Search.}
\label{fig:prompts_2}
\end{center}

\clearpage\begin{center}\captionsetup{hypcap=false}
\begin{minipage}{1.0\linewidth}\centering
\includegraphics[page=5,width=\linewidth]{images/fig13_task_prompts.pdf}
\par\small (a) Claim Charting
\end{minipage}\par\vspace{6pt}
\begin{minipage}{1.0\linewidth}\centering
\includegraphics[page=4,width=\linewidth]{images/fig13_task_prompts.pdf}
\par\small (b) Office-Action Response
\end{minipage}\par\vspace{6pt}
\captionof{figure}{English prompt displays for Claim Charting and Office-Action Response.}
\label{fig:prompts_3}
\end{center}

\clearpage\begin{center}\captionsetup{hypcap=false}
\begin{minipage}{1.0\linewidth}\centering
\includegraphics[page=6,width=\linewidth]{images/fig13_task_prompts.pdf}
\par\small (a) Chinese-to-English Translation
\end{minipage}\par\vspace{6pt}
\begin{minipage}{1.0\linewidth}\centering
\includegraphics[page=7,width=\linewidth]{images/fig13_task_prompts.pdf}
\par\small (b) English-to-Chinese Translation
\end{minipage}\par\vspace{6pt}
\captionof{figure}{English prompt displays for Chinese-to-English Translation and English-to-Chinese Translation.}
\label{fig:prompts_4}
\end{center}

\end{document}